\pdfoutput=1
\documentclass[11pt]{article}

\usepackage{fullpage}
\usepackage{amsmath,amssymb,amsthm,graphicx} 
\usepackage{epsfig}
\usepackage[numbers]{natbib} 
\usepackage{psfrag}
\usepackage{enumerate} 
\usepackage{enumitem} 
\usepackage{setspace}
\usepackage{float} 
\usepackage{color}
\usepackage{array}
\usepackage{pgf,tikz}
\usepackage{mathrsfs}
\usepackage{hyperref}
\usepackage{subcaption}
\usepackage{wrapfig}
\usepackage[ruled,vlined,linesnumbered]{algorithm2e}
\usepackage{tabularx}
\usepackage{makecell}

\usepackage[T1]{fontenc}
\usepackage{lmodern}
\usepackage{caption}
\usepackage{adjustbox}
\usepackage{geometry}

\usepackage[utf8]{inputenc} 
\usepackage[T1]{fontenc}    
\usepackage{hyperref}       
\usepackage{url}            
\usepackage{booktabs}       
\usepackage{amsfonts}       
\usepackage{nicefrac}       
\usepackage{microtype}      

\usetikzlibrary{arrows}
\definecolor{ffqqqq}{rgb}{1.,0.,0.}
\definecolor{xfqqff}{rgb}{0.4980392156862745,0.,1.}

\definecolor{mypink}{rgb}{0.858, 0.188, 0.478}
\definecolor{myred}{rgb}{1, 0, 0}

\makeatletter
\long\def\@makecaption#1#2{
        \vskip 0.8ex
        \setbox\@tempboxa\hbox{\small {\bf #1:} #2}
        \dimen0=\hsize
        \advance\dimen0 by 0cm
        \ifdim \wd\@tempboxa >\dimen0
                \hbox to \hsize{
                        \parindent 0em
                        \hfil 
                        \parbox{\dimen0}{\def\baselinestretch{0.96}\small
                                {\bf #1.} #2
                                } 
                        \hfil}
        \else \hbox to \hsize{\hfil \box\@tempboxa \hfil}
        \fi
        \vspace{0.4cm}
        }
\makeatother

\usepackage{macros}

\usepackage{amsmath,amsfonts,bm}

\def\eqref#1{(\ref{#1})}

\def\1{\bm{1}}

\def\eps{{\epsilon}}

\DeclareMathAlphabet{\mathsfit}{\encodingdefault}{\sfdefault}{m}{sl}
\SetMathAlphabet{\mathsfit}{bold}{\encodingdefault}{\sfdefault}{bx}{n}

\newcommand{\R}{\mathbb{R}}

\DeclareMathOperator*{\argmin}{arg\,min}

\usepackage{etoolbox}
\makeatletter
\patchcmd{\@algocf@start}
  {-1.5em}
  {0pt}
  {}{}
\makeatother

\makeatletter
\renewcommand{\paragraph}{%
  \@startsection{paragraph}{4}%
  {\z@}{0.25ex \@plus 1ex \@minus .2ex}{-1em}%
  {\normalfont\normalsize\bfseries}%
}
\makeatother

\newif\ifarxiv
\arxivtrue

\renewcommand\citet{\citep}

\makeatletter
\appto\@floatboxreset{%
  \ifx\@captype\andy@table
    \sffamily
  \fi
}
\def\andy@table{table}
\makeatother

\usepackage{caption} 
\newif\ifshownotes
\shownotestrue

\title{Margin-based sampling in high dimensions:\\ When being active is less efficient than staying passive}

\author{Alexandru Țifrea\thanks{Equal contribution.}, \enskip
  Jacob Clarysse\footnotemark[1], \enskip Fanny Yang\\
	Department of Computer Science, ETH Zurich\\
	\tt\small tifreaa@inf.ethz.ch \tt\small jacob.clarysse@inf.ethz.ch \tt\small fan.yang@inf.ethz.ch
}

\begin{document}
\hypersetup{pageanchor=false}

\maketitle

\begin{abstract}

It is widely believed that given the same labeling budget, active learning (AL)
algorithms like margin-based active learning achieve better predictive
performance than passive learning (PL), albeit at a higher computational cost.
Recent empirical evidence suggests that this added cost might be in vain, as
margin-based AL can sometimes perform even worse than PL. While existing works
offer different explanations in the low-dimensional regime, this paper shows
that the underlying mechanism is entirely different in high dimensions: we prove
for logistic regression that PL outperforms margin-based AL even for noiseless
data and when using the Bayes optimal decision boundary for sampling. Insights
from our proof indicate that this high-dimensional phenomenon is exacerbated
when the separation between the classes is small. We corroborate this intuition
with experiments on 20 high-dimensional datasets spanning a diverse range of
applications, from finance and histology to chemistry and computer vision.
\looseness=-1

\end{abstract}

\vspace{-0.3cm}
\section{Introduction}
\label{sec:intro}

\vspace{-0.1cm}
In numerous machine learning applications, it is often prohibitively expensive
to acquire labeled data, even when unlabeled data is readily available.  For
instance, consider the task of inferring the sleep quality of a patient from
data collected during usual health checks (e.g. EEG, EKG, blood tests etc). To
get a high-precision label for this task, patients need to spend a night in a
sleep lab, which is expensive and time-consuming. Therefore, the labeled dataset
that we can collect cannot be too large. However, a large unlabeled set of
medical records of similar patients is available and can potentially be
leveraged for the task.
Active learning algorithms aim to reduce labeling costs, by collecting a small
labeled set that still results in a model with good predictive performance. 
\looseness=-1

\vspace{-0.1cm}
A popular family of active learning algorithms is margin-based active
learning (M-AL) \citep{Scheffer01, Balcan07,
ducoffe18}.
This paradigm proposes to alternate between (i) training a prediction model
(e.g.\ logistic regression, deep neural network) on the currently available
labeled set; and (ii) augmenting the labeled set by acquiring labels for the
unlabeled points that lie close to the decision boundary of the model.
M-AL is closely related to strategies like uncertainty sampling \citep{Lewis94},
entropy sampling \citep{settles09}, or softmax sampling for neural networks.
\looseness=-1



\begin{figure*}[t]
  \centering
  \begin{subfigure}[t]{0.67\textwidth}
    \centering
    \includegraphics[width=\textwidth]{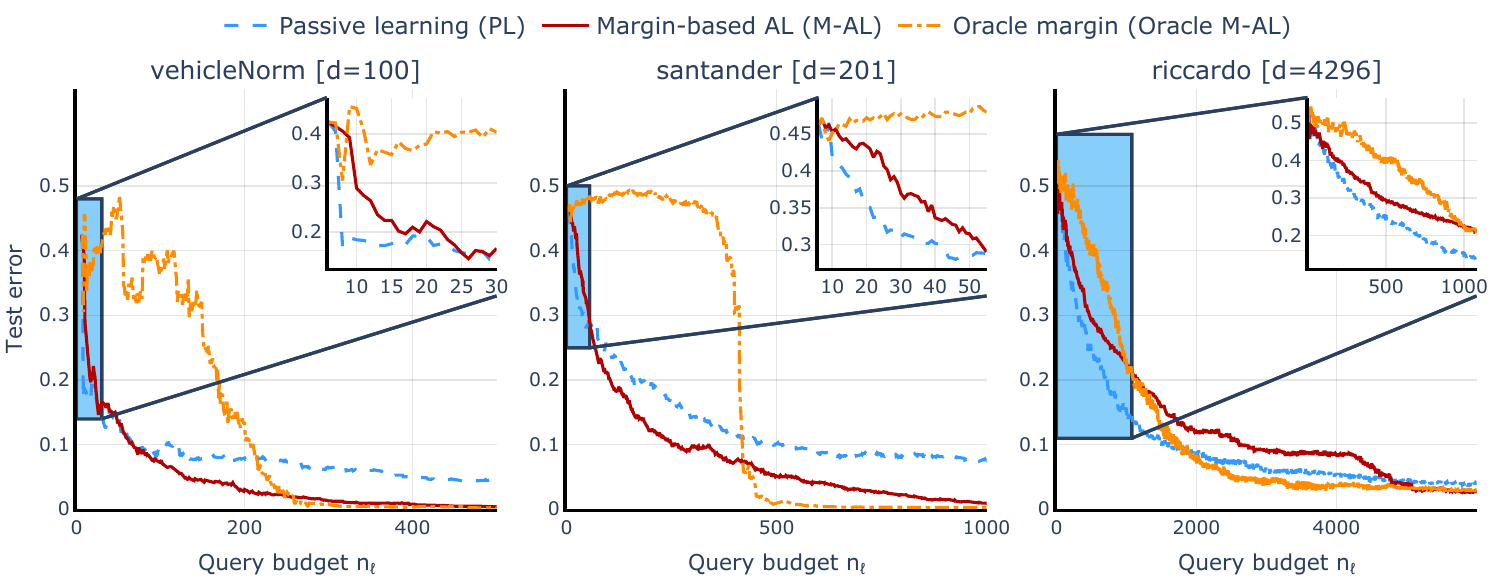}
  \end{subfigure}
  \begin{subfigure}[t]{0.29\textwidth}
    \centering
    \includegraphics[width=\textwidth]{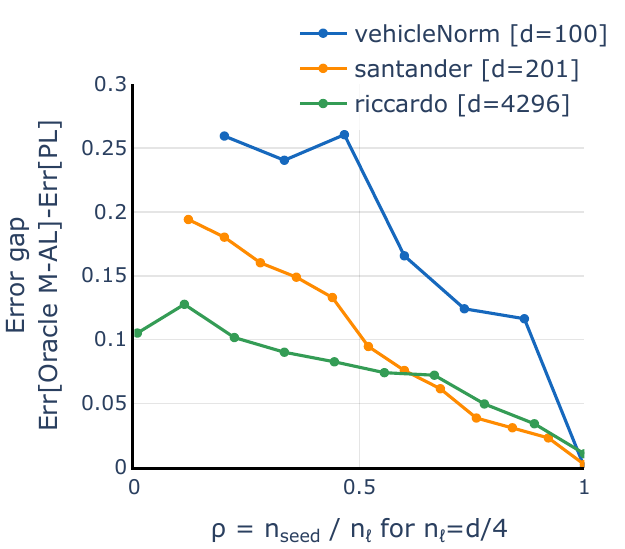}
  \end{subfigure}


  \caption{\small{\textbf{Left:} In the low-dimensional regime ($\nlab \gg d$) M-AL achieves
      low test error with fewer labeled samples than PL. However, in the
      high-dimensional regime ($\nlab \ll d$, see zoomed-in insets), M-AL
      fails. Oracle M-AL exhibits the same failure in high dimensions, despite
      performing well for large query budgets, thus ruling out the cold start
      problem as a cause for this phenomenon.  
      \textbf{Right:} Increasing the seed set size $\nseed$ reduces the gap
      between Oracle M-AL and PL.
      See Appendix~\ref{sec:appendix_error_vs_budget} for more datasets.}}
  \label{fig:teaser}
  \vspace{-0.5cm}

\end{figure*}

\vspace{-0.1cm}
Numerous prior works have documented the success of M-AL in low dimensions
\citep{tong01, schein07, yang18, schohn20}. As it is evident in
Figure~\ref{fig:teaser}, in the regime where the query budget is large (i.e.\
$\nlab \gg d$), M-AL achieves low test error with a lot less labeled data than
passive learning (PL, i.e.\ \emph{uniform sampling}). This is in line with the
intuition about M-AL developed in prior works. At the same time, the figure
reveals that in the low-sample regime (i.e.\ $\nlab \ll d$) M-AL ``fails'' --
that is, it leads to worse predictive error than PL. This regime is much less
studied in the AL literature.\footnote{The work of \citet{Zhang18} also focuses
  on high-dimensional data, but with the purpose of improved computational
efficiency.}

\vspace{-0.1cm}
In this paper, we characterize theoretically and empirically the settings that
lead to the failure of M-AL for high-dimensional logistic regression.  We rule
out two likely causes for this phenomenon.  First, it is known that, in low
dimensions, M-AL does not improve upon the sample efficiency of PL when the
Bayes optimal model has high error \citep{mussmann18}.
Second, several works \citep{huang14, sener18, Hacohen2022} argue that M-AL can
fail due to the \emph{cold start problem}: using only a small labeled set, one
cannot obtain a meaningful decision boundary to be used for sampling. 
\looseness=-1

\vspace{-0.1cm}
Perhaps surprisingly,
for high-dimensional problems with a low labeling budget, M-AL underperforms PL
\emph{even} when (i) the Bayes error is zero; 
and (ii) one uses the distance to the Bayes optimal decision boundary for
sampling
(referred to as \emph{oracle margin-based active learning} or \emph{Oracle
M-AL}).
Our experiments reveal the failure of M-AL for logistic regression on a wide
variety of datasets, a subset of which are presented in Figure~\ref{fig:teaser}
(a few works make a similar observation for neural networks \citep{sener18,
Hacohen2022, sorscher2022}). This failure of M-AL occurs only in the low-budget
regime (see zoomed-in insets in Figure~\ref{fig:teaser}), which coincides with
the scenario in which AL is often employed in practice (i.e.\ high labeling costs).
%
%
Since prior explanations do not apply to the setting that we consider
(high-dimensional, noiseless), to date, there exists no result that sheds light
on this failure case.
Our contributions in this paper are as follows: \looseness=-1



\vspace{-0.4cm}
\begin{enumerate}[leftmargin=*]
  \compresslist

  \item We observe that M-AL performs worse than PL for logistic regression on
    numerous high-dimensional datasets from different application domains (e.g.\
    finance, chemistry, histology).
    \looseness=-1

  \item We prove non-asymptotic error bounds for logistic regression that
    directly imply worse performance of M-AL compared to PL -- even when using
    Oracle M-AL (Section~\ref{sec:theory}) and for data distributions
    with noiseless labels (truncated Gaussian mixture and Gaussian marginal).
    \looseness=-1
    
  \item Distinct from the low-dimensional intuition \citep{mussmann18}, our
    proof suggests that in high dimensions M-AL benefits from a
    large separation margin between the classes.
    We confirm this intuition experimentally for logistic regression on
    $\numdatasets$ real-world datasets (Section~\ref{sec:experiments}). 
\looseness=-1

\end{enumerate}
\vspace{-0.3cm}

Our results reveal that for high-dimensional data, margin-based AL is not
only more computationally costly compared to passive learning, but often
provably less effective as well.
Our paper hence suggests an important avenue for future work: identify active
learning algorithms that are provably
consistent and substantially outperform passive learning in high-dimensional and
low-budget settings. 
\looseness=-1

\vspace{-0.1cm}
\section{Active learning for classification}
\label{sec:setting}

We now introduce the active learning framework that we consider throughout 
this paper.
Our high-level goal is to train a binary classifier that predicts a label $y \in
\{-1, 1\}$ from covariates $x \in \R^d$, where $(x, y) \sim \PP_{XY}$.
More specifically, we seek parameters $\theta \in \Theta$ such that
the classifier $x \rightarrow \sgn\left( \fx{x}\right)$ achieves a low
population error $\testerr{\f}= \EE_{(x, y) \sim \PP_{XY}} \indicator[y\neq
\sgn(\fx{x})]$.
%
In practice 
the population error is not available,
and hence, one can instead minimize the empirical risk defined by a loss
function $\loss$ on a collection of labeled training points $\thetahat =
\argmin_\theta \frac{1}{|\Dlab|} \sum_{(x, y) \in \Dlab} \loss(\fx{x}, y)$.
The goal of active learning is to find a good set $\Dlab$ which induces a
$\thetahat$ that generalizes well.  \looseness=-1


\paragraph{Collecting the training set via margin-based AL.}
We consider standard pool-based active learning
like in Algorithm~\ref{algo:us} and assume access to a large unlabeled dataset
$\Dunl$ of size $\nunl$.  At first the labeled set $\Dlab$ consists of a small
seed set $\Dseed$ 
containing 
$\nseed$ i.i.d.\ samples drawn from the training distribution.
At each querying step $\step$, we first sample and label the unlabeled point
that is closest to the decision boundary of the trained classifier
according to distance function $\distance{x}{\theta} \in [0,\infty)$\footnote{In
  Section~\ref{sec:appendix_eps_greedy} we discuss the implications of our
results to strategies that combine a diversity and a margin-based score (e.g.\
\citet{brinker03}).} and add it to the labeled set.
Then we train a classifier on the resulting labeled set.\footnote{For the
theoretical analysis of M-AL we use the same modification of \citet{chaudhuri15,
mussmann18} to slightly change this procedure (see
Section~\ref{sec:main_theory}).}
These querying steps are repeated until we exhaust the labeling budget, denoted
by $\nlab$ (we use labeling or query budget interchangeably). Moreover, we
define the seed set proportion $\passiveratio := \frac{\nseed}{\nlab}$, which
effectively captures the fraction of labeled points sampled via uniform
sampling. Note that $\passiveratio=1$ corresponds to passive learning.

\paragraph{Oracle vs empirical M-AL.} In practice, new queries are selected
using a classifier trained on the currently available labeled data, as described
in the paragraph above. We refer to this strategy as \emph{empirical M-AL}. We
also consider a setting that could potentially be more
beneficial for M-AL, namely using the Bayes optimal classifier for sampling at
every querying step (i.e.\ using $\thetastar$ instead of $\thetahat$ in the
first step of the \emph{for} loop in Algorithm~\ref{algo:us}). We call this
strategy \emph{oracle M-AL} and elaborate on how it compares to empirical M-AL in Section~\ref{sec:theory}.

\paragraph{Intuition behind margin-based AL.} Intuitively,
in low dimensions (i.e.\ $d < \nlab$), M-AL behaves like binary search
\citep{Cohn94}, and hence, needs significantly fewer samples to find the optimal
decision boundary. Intuitively, in low dimensions, sampling based on the margin
of the Bayes optimal classifier (i.e.\ oracle M-AL) is expected to further
improve the sample complexity of M-AL for noiseless data (see
Section~\ref{sec:low_dim}).  Finally, note that M-AL is often equivalent to
uncertainty sampling \citep{Lewis94}. For instance, for binary linear predictors
under the logistic noise model, the uncertainty is proportional to the distance
between $x$ and the decision boundary determined by $\theta$ \citep{Platt99,
mussmann18}.
\looseness=-1

\setlength{\textfloatsep}{0.2cm}
        \setlength{\algomargin}{0.5cm}
        \begin{algorithm}[t]
        \DontPrintSemicolon
        \small
 
     \SetKwInOut{KwInput}{Input}

     \KwInput{Seed set $\Dseed$, unlabeled set $\Dunl$, budget $\nlab$,
     distance function $\distfnc$, loss function $\loss$}

     \KwResult{Prediction model $\fhat$}

     $\Dlab \gets \Dseed$

     $\thetahat \gets \argmin_{\theta} \frac{1}{|\Dseed|} \sum_{(x, y) \in
     \Dseed} \loss(\fx{x}, y)$

     \For{$\step \in \{|\Dseed|+1, ..., \nlab \}$} {
       $\xquery \gets \argmin_{x \in \Dunl} \distance{x}{\thetahat}$ \;\label{alg:sample}
       $\yquery \gets \textit{AcquireLabel}(\xquery)$ \;
       $\Dlab \gets \Dlab \cup \{(\xquery, \yquery)\}; \Dunl \gets \Dunl
     \setminus \{\xquery\}$ \;
       $\thetahat \gets \argmin_{\theta} \frac{1}{|\Dlab|} \sum_{(x, y) \in
       \Dlab} \loss(\fx{x}, y)$ \;
     }

     \KwRet $\fhat$ 

     \caption{Margin-based active learning}
     \label{algo:us}
    \end{algorithm}

\begin{wrapfigure}{h}{0.5\linewidth}
\vspace{-0.3cm}
  \centering
  \includegraphics[width=0.4\columnwidth]{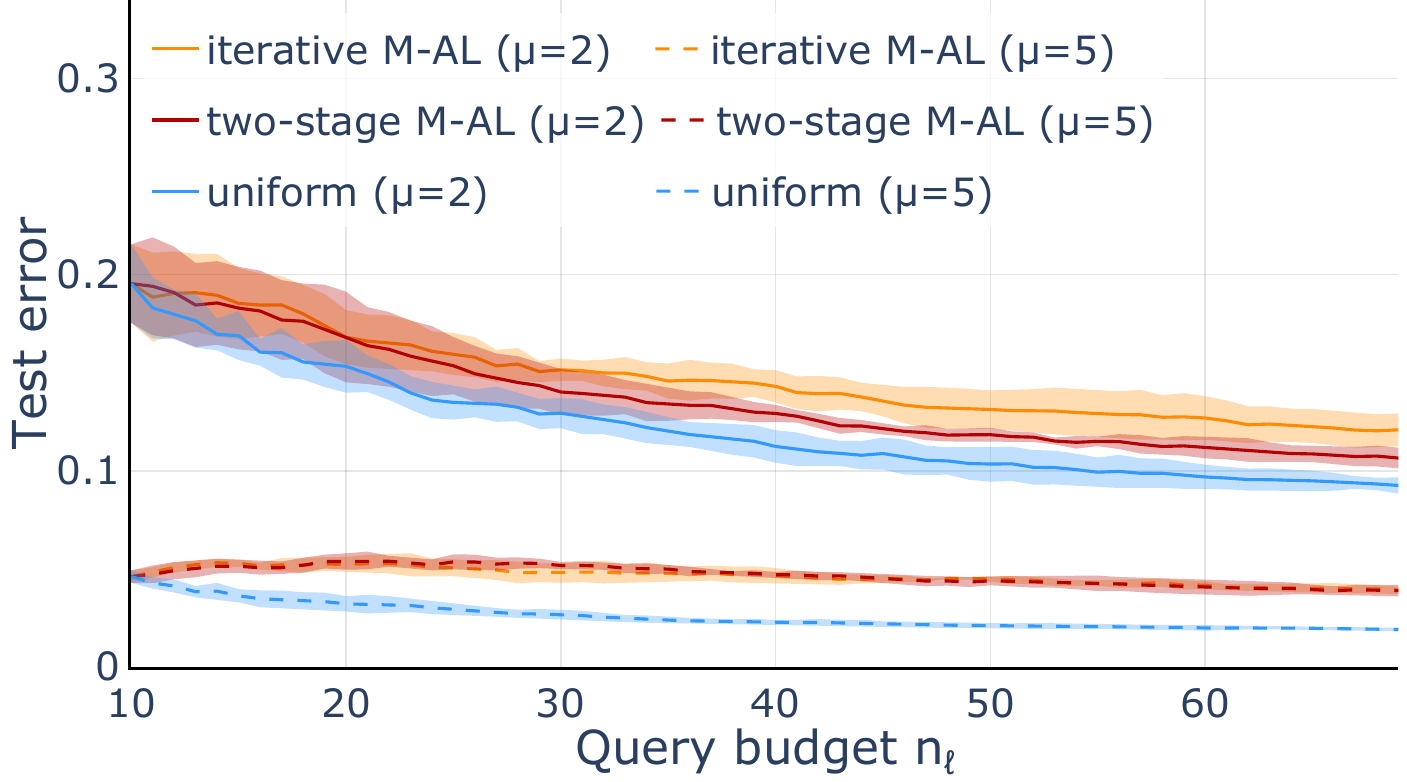}
  \caption{\small{Two-stage M-AL (which we analyze in
      Section~\ref{sec:main_theory}) is
  on-par or better than iterative M-AL (Algorithm~\ref{algo:us}). Data is drawn
  from the truncated mixture distribution from Section~\ref{sec:datadistr}, with
$\mu \in \{2, 5\}, \sigma=3$ and $d=1000$. Shaded areas
indicate standard deviations over 5 runs.}}
  \label{fig:twostage_synth}
\vspace{-0.5cm}
\end{wrapfigure}

\paragraph{Two-stage margin-based AL.} Similar to other theoretical analyses of
active learning \citep{mussmann18, chaudhuri15}
we modify Algorithm~\ref{algo:us} slightly, and develop the theory instead for a
two-stage procedure:
1) we obtain $\thetaseed$ using the initial small seed set; and 2) we use
$\thetaseed$ to select a batch of $(1-\passiveratio)\nlab$ samples to query from
the unlabeled set. This two-stage process avoids the dependence of the
classifier at stage $n$ on the unlabeled dataset. \citet{chaudhuri15} argue that
two-stage strategies are asymptotically not worse than iterative strategies for
MLE estimators. Moreover, \citet{mussmann18} show that theoretically analyzing
this two-stage strategy can reveal insights about iterative AL
strategies that are confirmed by experiments on real-world data. In our setting
as well, experiments with the two-stage strategy in
Figure~\ref{fig:twostage_synth} follow closely the same trends as iterative
M-AL.

\section{Theoretical analysis of margin-based AL in high dimensions}
\label{sec:theory}

In this section we give rigorous intuition for the failure of M-AL for
high-dimensional logistic regression.
In particular, we prove that two-stage M-AL is less sample efficient than PL.
Since two-stage M-AL is empirically on-par or better than iterative M-AL in the
setting that we consider (Figure~\ref{fig:twostage_synth}), our theory gives
insights into the failure of the strategy in Algorithm~\ref{algo:us}. 
To rule out the \emph{cold start problem} as a potential cause of this
phenomenon, we prove that M-AL also fails when using the Bayes optimal decision
boundary for sampling (oracle M-AL), a strategy that is more favorable than M-AL
in low-dimensions.
%


\vspace{-0.1cm}
\subsection{Logistic regression and the max-$\ell_2$-margin solution}
\label{sec:uncert_linear}

We consider linear models of the form $\fx{x} = \langle\theta, x\rangle$ with
$\theta$ in a fixed-norm ball and minimize the logistic loss, i.e.\ $\loss(z,
y)=\log(1+e^{-zy})$. We note that for linearly separable data, minimizing the
logistic loss with gradient descent recovers the max-$\ell_2$-margin
(interpolating) solution \citep{soudry18,Ji19}.  The generalization behavior of
this interpolating estimator has been analyzed extensively in recent years in
different contexts \citep{Bartlett20, Javanmard20b,  muthukumar21, donhauser21}.  
In what follows, we
refer to the max-$\ell_2$-margin classifier trained on a labeled dataset
acquired (i) via uniform sampling ($\nseed=\nlab$) as $\thetaunif$ and  (ii) via
oracle and empirical M-AL respectively as $\thetaor$ and
$\thetaunc$.

\vspace{-0.1cm}
\subsection{Data distribution}
\label{sec:datadistr}

We now introduce the family of joint data distributions $\PP$
for our theoretical analysis that includes distributions where the covariates
follow a Gaussian or mixture of truncated Gaussians distribution. These
distributions can adequately approximate data generated in many practical
applications \citep{bouguila19} and have often been considered in theoretical
analyses of machine learning algorithms \citep{tsipras2018, mahdi20, frei22}.

\paragraph{Noiseless and balanced observations.} Recall that for
linear classifiers, the intuitive reason for the effectiveness of M-AL sampling
is that after only a few queries, it selects points in the neighborhood of the
optimal decision boundary. Notice that this region is where the label noise is
concentrated, for a Gaussian mixture model. Therefore, M-AL is prone to query
numerous noisy samples. We wish to show the failure of M-AL even in the most
benign setting.
Hence, we assume a noiseless binary classification problem where the Bayes error
vanishes, i.e.\ $\testerr{\thetastar}=0$ for some $\thetastar$ unique up to
scaling and in this section we set $\|\thetastar\|_2=1$.\footnote{Note that
similar results can also be derived for noisy data.}   
More precisely, we assume that the joint distribution $\PP$ is such that the
labels $y = \sgn(\langle \thetastar, x \rangle) \in \{-1, 1\}$ with
$\thetastar\in \RR^d$. For ease of exposition, we assume without loss of
generality that $\thetastar = e_1 = [1, 0, ..., 0]$; if $\thetastar\neq e_1$ we
can rotate and translate the data to get $\thetastar=e_1$ (see
Appendix~\ref{sec:appendix_rotation} for more details).  We can then rewrite the
covariates as $x = [\xsig, \xnoise]$ to distinguish between a signal $\xsig \in
\RR$ and non-signal component $\xnoise \in \RR^{d-1}$. Further, to disentangle
from phenomena stemming from imbalanced data,
we consider a distribution with equal class proportions in expectation (see
Appendix~\ref{sec:appendix_balanced} for a discussion).



We obtain a family of joint distributions satisfying these conditions by
sampling $y \in \{+1, -1\}$ each  with probability one half, and then sampling
from the class-conditional distribution defined by $\PP_{\xsig | y} =
\gausstr(y\mean, \stdsig^2, y)$ and $\PP_{\xnoise | y} = \gauss(0,I_{d-1})$,
where $\gausstr(y\mu, \sigma^2, y)$ denotes the truncated Gaussian distribution
with support $(-\infty, 0)$
if $y=-1$ and, respectively, support $(0,\infty)$ if $y=1$. The parameters $\mu,
\stdsig  \ge 0$ denote the mean and standard deviation of the non-truncated
Gaussian. 
\paragraph{Gaussian marginals.} Further note that by setting $\mean=0$, we
recover the marginal Gaussian covariate distribution (also known as a
discriminative model) -- a popular distribution to prove benefits for active
learning \citep{beygelzimer10,Hanneke13}, as both supervised and semi-supervised
learning require large amounts of labeled data to achieve low prediction error,
even given infinite unlabeled samples \citep{scholkopf2012}.

\begin{figure*}[t]
  \centering
  \begin{subfigure}{0.31\textwidth}
    \centering
    \includegraphics[width=\textwidth]{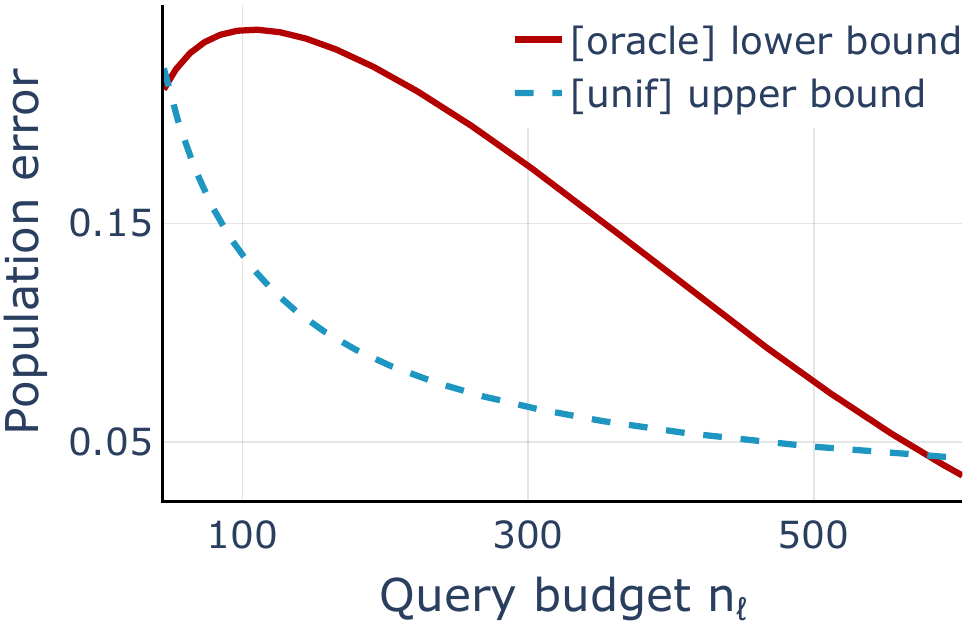}
    \caption{Error vs $\nlab$ (fixed $\nseed$).}
    \label{fig:oracleUStheory}
  \end{subfigure}
  \begin{subfigure}{0.31\textwidth}
    \centering
    \includegraphics[width=\textwidth]{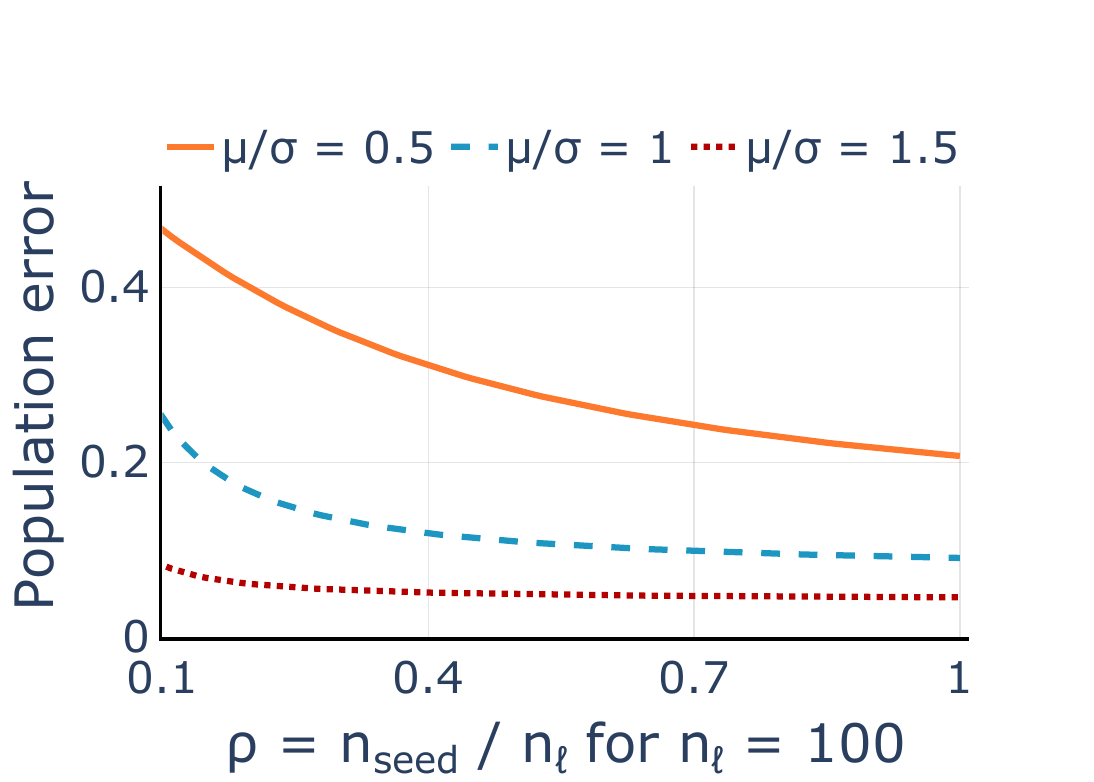}
    \caption{Error vs $\passiveratio$ (fixed $\nlab$).}
    \label{fig:seedset}
  \end{subfigure}
   \begin{subfigure}{0.33\textwidth}
    \centering
    \includegraphics[width=\textwidth]{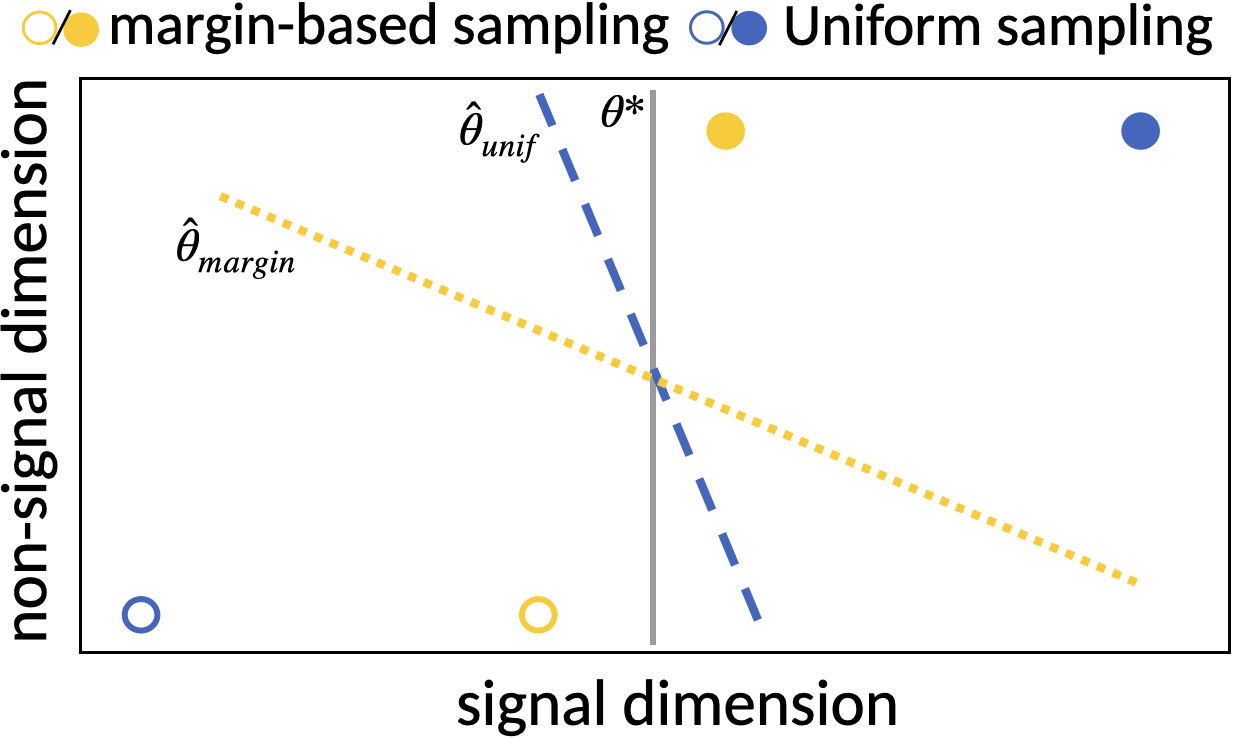}
    \caption{Proof intuition in 2D.}
    \label{fig:synth_sketch}
  \end{subfigure}

  \caption{\small{ 
        Theoretical population error lower bounds for oracle M-AL and upper bounds for PL
        in Theorem~\ref{thm:mixturesoracle} on the truncated Gaussian mixture model.
      (a)~For large $d/\nlab$ the lower bound of Theorem~\ref{thm:mixturesoracle}
      on the error of oracle M-AL is much larger than the upper bound on the
      error of PL for fixed $\nseed =10$ and $\mean/\stdsig = 1$. 
      (b)~The lower bound on the error is smaller when the seed set
    proportion $\passiveratio$ and the ratio $\mean/\stdsig$ are increasing, for fixed $d =
    1000$ and $\nlab = 100$.
      (c)~Intuition for the failure of M-AL in high
      dimensions. The classifier assigns higher weight to the non-signal
      dimension when trained on the yellow points close to the optimal
      decision boundary.
  }} \label{fig:synthetic}
\vspace{-0.3cm}
\end{figure*}

\paragraph{Mixture of truncated Gaussians.} For $\mu > 0$, the marginal
covariate distribution is a mixture of two truncated Gaussians.
Each truncated component has standard deviation $\stdsigtr \leq \sigma$ and mean

\vspace{-0.5cm}
\begin{equation}
\label{eq:meantr_main}
\meantr := \EE[ y \xsig ] = \mean + \stdsig
\frac{\phi(-\mean/\stdsig)}{1-\Phi(-\mean/\stdsig)},
\end{equation}
\vspace{-0.4cm}

\noindent where $\mean$ and $\stdsig$ determine the \emph{non-truncated}
Gaussian, and $\phi$ and $\Phi$ denote the pdf and the CDF of the standard
normal distribution, respectively.

\vspace{-0.2cm}
\subsection{Warm-up: M-AL versus PL in low dimensions}
\label{sec:low_dim}

\vspace{-0.2cm}
Before we introduce our main results, we discuss why we expect M-AL to
outperform PL for this family of prediction problems. In low-dimensions,
\citet{mussmann18} show that M-AL requires significantly fewer samples than PL
to achieve the same test performance for distributions with vanishing Bayes
error -- this corresponds to noiseless data. Moreover, in low dimensions and for
noiseless data, oracle M-AL \emph{further improves} the sample complexity, as
implied, for instance, by the results in \citet{chaudhuri15} (see
Appendix~\ref{sec:appendix_oracle_mal} for further discussion and an
illustrative 1D example). Indeed, experiments on real-world tabular
(Figure~\ref{fig:teaser}) or image data \citep{sener18} also confirm that oracle
M-AL outperforms M-AL for large labeling budgets. However, we show in the
following sections that these intuitions do not transfer to the low-sample
regime. Oracle M-AL has not been studied in this setting, prior to our work.
\looseness=-1

\vspace{-0.25cm}
\subsection{Main result for high-dimensional M-AL}
\label{sec:main_theory}

\vspace{-0.2cm} In this section we present theorems that rigorously prove how
logistic regression with margin-based AL leads to worse classifiers in the
high-dimensional setting (i.e.\ $\nlab \ll d$) where most samples are acquired
with M-AL (i.e.\ $\passiveratio \ll 1$) --- a phenomenon observed on real-world
data in Figure \ref{fig:teaser}-Right.  Moreover, we discuss an insight that
directly follows from the proof intuition: if many samples in the
unlabeled dataset are close to the optimal decision boundary (i.e.\ small
$\mean/\sigma$ ratio), the error gap between M-AL and PL increases. We
confirm this intuition 
on real-world data in Section~\ref{sec:exp_verify_theory}.

First, we state the assumptions under which our results hold for
high-dimensional active learning. Formal versions of these conditions can be
found in the Appendix~\ref{sec:appendix_thm}.

\begin{assumption}
  \label{ass:main_ass}
  Let $\nlab$ be the labeling (or query) budget and $\nunl$ the unlabeled
  set size. Assume that $\nlab \ll \nunl$ and $d \gg \nlab$.
  Moreover, consider the distribution
  described in Section~\ref{sec:datadistr} with $\mean/\stdsig < 2$, and let $\Dseed$
  and $\Dunl$ be datasets
  drawn i.i.d.\ from these joint and the marginal distributions, respectively.
\end{assumption}

Our main results provide lower bounds on the gap between the population error
resulting from M-AL versus PL (i.e.\ uniform sampling).  We first state a
theorem characterizing this gap for oracle M-AL and refer to
Appendix~\ref{sec:appendix_thm} for the formal statement and proof.
\looseness=-1

\begin{theorem}[informal]
\label{thm:mixturesoracle}

Consider the setting introduced in~\Cref{ass:main_ass}. Then there exist universal constants
$0<\cone, \epsilon \ll 1$ and $t, \ctwo>0$ such that with probability larger
than $1-e^{-\ctwo t^2/2}$ it holds that:
\vspace{-0.1cm}
\begin{equation*}
\testerr{\thetaor} - \testerr{\thetaunif} > \Psifull\left(\alphaor \right) -
\Psifull\left(\alphaPL \right),
\vspace{-0.1cm}
\end{equation*}
\noindent where $\Psifull$ is a strictly increasing function, defined in Section
\ref{sec:appendix_tgmm}, and
\vspace{-0.1cm}
\begin{equation*}
\vspace{-0.1cm}
\begin{aligned}
\vspace{-0.3cm}
\alphaor &=  \frac{(1-\epsilon)\sqrt{(d-1)/\nlab}}{\passiveratio \left(
\mean_{tr}+ t \stdsig_{tr}/\sqrt{\rho \nlab} \right) + \cone
(1-\rho)\mean_{tr}}-1,\\
    \alphaPL &=
      \frac{(1+\epsilon)\sqrt{(d-1)/\nlab}}{\mean_{tr}-t\stdsig_{tr}/\sqrt{\nlab}}. 
\end{aligned}
\end{equation*}
In particular, if $\passiveratio <  \delta $ for $0< \delta < 1/2$, then
\vspace{-0.1cm}
\begin{equation*}
  \testerr{\thetaor} - \testerr{\thetaunif} > 0.
\vspace{-0.1cm}
\end{equation*}
\end{theorem}
\vspace{-0.2cm}

Note that $\Psifull$ is similar to the cumulative distribution function of the
Gaussian distribution (see Appendix~\ref{sec:appendix_tgmm} for the exact
definition and more details). 

A similar result can be proved for two-stage margin-based active learning. We
state the theorem informally and defer the formal statement and proof to
Appendix~\ref{sec:appendix_thm}.
%


\begin{theorem}[informal]
\label{thm:mixtureUS}

Consider the setting introduced in \Cref{ass:main_ass} and further assume that
$\stdsig > 1$.
Then there exist universal constants $0< \cone, \epsilon \ll 1$ and
$t, \ctwo>0$ such that with probability larger
than $1-e^{-\ctwo t^2/2}$ it holds that:
\begin{equation*}
\testerr{\thetaunc} - \testerr{\thetaunif} > \Psifull\left( \alphaUS \right) -
\Psifull\left( \alphaPL \right),
\end{equation*}
where $\Psifull, \alphaPL$ as in Theorem \ref{thm:mixturesoracle} and
\vspace{-0.2cm}
\begin{align*}
    \alphaUS = \frac{(1-\epsilon)\sqrt{\frac{d-1}{\nlab}}}{ \passiveratio \Cseed
    +(1-\passiveratio) \left(\cone\mean_{tr} + \sqrt{\frac{\log  \nunl}{\Cseed}}
\sqrt[4]{\frac{(1+\epsilon)(d-1)}{\passiveratio \nlab}}\right)}-1,
\vspace{-0.2cm}
\end{align*}
with $|\mean_{tr} - \Cseed | < t\stdsig_{tr}(\passiveratio\nlab)^{-1/2}$. In
particular, for small fixed constants $\cthree, \cfour, \cfive >0$ if
$\passiveratio \nlab < \cthree$, and
\begin{equation*}
\begin{aligned}
  \mean_{tr} > \cfour  \left(\frac{d-1}{\passiveratio\nlab} \right)^{1/6}(\log
\nunl)^{1/3} + \frac{t\stdsig_{tr}}{(\passiveratio \nlab)^{1/2}},\\ 
  \passiveratio <   \cfive \frac{\Cseed}{ \mean_{tr}- \frac{t \stdsig_{tr}}{(\passiveratio\nlab)^{1/2}}
 - \left(\frac{d-1}{\passiveratio \nlab}\right)^{1/6}(\log \nunl)^{1/3}},
\end{aligned}
\end{equation*}
then $\testerr{\thetaunc} - \testerr{\thetaunif} > 0$. 
\end{theorem}

\vspace{-0.2cm}
The proof shares the same intuition and key steps as the proof of
Theorem~\ref{thm:mixturesoracle} and differs only in certain technicalities that
arise from estimating the classifier $\thetaseed$. In particular, 
we additionally need to assume that the mean $\meantr$ is large enough such that
the classifier trained on the seed set has non-trivial prediction error. Note
that it is possible to obtain a large $\meantr$ even for the marginal Gaussian
case when $\mu=0$, by choosing a large $\stdsig$ in
Equation~\eqref{eq:meantr_main}.

\vspace{-0.3cm}
\subsection{Proof sketch and interpretation}
\label{sec:proof_and_intuition}

\vspace{-0.2cm} In this section we present the intuition behind the proofs of
the theorems and discuss the insights revealed by the theory. For simplicity, we
focus primarily on Theorem~\ref{thm:mixturesoracle} for which the phenomenon is
more pronounced.  We note that the same arguments hold for the setting in
Theorem~\ref{thm:mixtureUS}.

\paragraph{Discussion of assumptions.} We now discuss the conditions needed for
the theorems.  Observe that if the ratio $\mu/\sigma$ is large, then even PL 
achieves low error. Hence, we assume settings where $\mean/\stdsig <
2$, which also cover real-world datasets that contain ambiguous samples that lie
near the optimal decision boundary.  If $\mean / \sigma < 2$ then the covariate
distribution has sufficient density in the neighborhood of the optimal decision
boundary. Hence, with high probability, the labeled data acquired with M-AL lies
in this region, leading to an estimator with high population error.  Finally, a
small seed set proportion $\passiveratio$ allows for sufficiently many active
queries, and is common in situations that employ AL.  We refer to
Appendix~\ref{sec:appendix_assumptions} for further arguments supporting the
practical relevance of the setting.

\paragraph{Proof intuition.}  
Figure~\ref{fig:synth_sketch} illustrates the intuition behind the failure of
M-AL in high dimensions using a 2D cartoon. We depict the samples chosen by
oracle M-AL (yellow), which lie close to the optimal decision
boundary (vertical line) and the points selected by uniform sampling (blue),
which are farther away from the optimal decision boundary. Note that for both
sampling strategies, the selected samples are far apart in the non-signal
direction. More specifically, in high dimensions the large distance in the
non-signal components $\xnoise$ is a consequence of sampling $\xnoise$ from a
multivariate Gaussian.
It follows from these facts that the max-$\ell_2$-margin classifier trained on
the samples near the optimal decision boundary (yellow dotted) is more tilted
(and hence, has larger population error) than the one trained on uniformly
sampled points that are further away (blue dashed).
\looseness=-1

\paragraph{Interpretation of theoretical results.}
Theorem \ref{thm:mixturesoracle} characterizes when the population error gap
between oracle M-AL and PL is positive: 
for small labeling budgets (leading to a large $d/\nlab$ ratio), for a small
seed set proportion ($\passiveratio \ll 1$) and for sufficiently many unlabeled
samples near the Bayes optimal decision boundary (implied by $\mean/\stdsig <
2$). In particular, in the regime $d/\nlab \to \infty$ that implies $\eps\to 0$
(as argued in Appendix~\ref{appendix_formal_oracle}), we can observe how for
small $\passiveratio \approx 0$, it holds that $\alphaor \gg \alphaPL$ as $0<
\cone \ll 1$. In turn, $\alphaor \gg \alphaPL$ implies $\testerr{\thetaor} -
\testerr{\thetaunif} \gg 0$ since $\Psifull$ is strictly increasing. In
Figure~\ref{fig:synthetic} we depict how the error bounds, and hence the gap in
Theorem~\ref{thm:mixturesoracle}, depend on the three quantities $\passiveratio,
d/\nlab$ and $\mean/\stdsig$. 


In Figure~\ref{fig:oracleUStheory}, we show the dependence of the bound in
Theorem~\ref{thm:mixturesoracle} on $\nlab$ (and hence, $d/\nlab$), for fixed
$\nseed = 10, d=1000$ and $\mean/\stdsig = 1$. If the query budget $\nlab$ and
the ratio $\passiveratio$ are small (the middle region of the plot on the
horizontal axis), we have a large error gap between oracle M-AL and PL. This
phenomenon is inherently high-dimensional and stops occurring for large sample
sizes $\nlab$ (the right part of the figure).
We also identify these regimes in experiments on real-world data in
Section~\ref{sec:main_results}.  In Appendix~\ref{app:synthetic_experiments}, we
show more evidence that the theoretical bounds closely predict the values from
simulations.  Note that the bounds are loose for extremely small budgets (left
part of the figure).

In Figure~\ref{fig:seedset}, we vary the seed set size $\nseed$ (and hence, the
ratio $\passiveratio$), for fixed $\nlab= 100, d = 1000$.
We observe that increasing the seed set proportion $\passiveratio$ reduces the
error of oracle M-AL (note that $\passiveratio = 1$ corresponds to PL). We
highlight that the dependence of the error on the ratio
$\passiveratio$ is not due to the decision boundary used for M-AL becoming more
meaningful for larger $\passiveratio$, as conjectured by some prior works
\citep{huang14, sener18}: in our case, we sample using the Bayes
optimal decision boundary at every querying step. Instead,
Theorem~\ref{thm:mixturesoracle} captures another failure case of M-AL, specific
to high-dimensional settings.


Moreover, Figure~\ref{fig:seedset} also illustrates the
dependence of the error lower bound on the ratio $\mean/\stdsig$, for oracle
M-AL. In Theorem~\ref{thm:mixturesoracle}, the distribution-dependent ratio
$\mean/\stdsig$ enters the bounds via the quantity $\cone$ which is strictly
increasing in $\mean/\stdsig$ (the exact dependence of $\cone$ on
$\mean/\stdsig$ is presented in Lemma~\ref{lem:dqstar} in
Appendix~\ref{sec:appendix_thm}).
For small $\mean/\stdsig$, the error gap between oracle M-AL ($\passiveratio <
1$) and PL ($\passiveratio=1$) is large.

Finally, this phenomenon is caused by choosing to label samples close
to the Bayes optimal decision boundary, which are, by definition, the points
queried with oracle M-AL.
This suggests that, perhaps surprisingly, oracle M-AL exacerbates this
high-dimensional phenomenon. Indeed, as evident from comparing the bounds in
Theorems~\ref{thm:mixturesoracle} and~\ref{thm:mixtureUS}, oracle M-AL performs
even worse than M-AL that uses the margin of an empirical estimator
$\thetahat$.  We confirm this intuition on real-world datasets
in Appendix~\ref{sec:appendix_heatmap_oracle}.


\vspace{-0.3cm}
\section{Experiments}
\label{sec:experiments}

\vspace{-0.2cm}
In this section, we provide extensive experiments to investigate the
ineffectiveness of low-budget margin-based sampling on high-dimensional
real-world data.
In particular, we train logistic regression with oracle and empirical
M-AL on a wide variety of tabular datasets.

\vspace{-0.3cm}
\subsection{Datasets}
\label{sec:data}

\vspace{-0.2cm}
We select binary classification datasets from OpenML
\citep{OpenML2013} and from the UCI data repository \citep{Dua2019} according to
a number of criteria: i) the data should be high-dimensional ($d > 100$) with
enough samples that can serve as the unlabeled set ($\nunl > \max(1000, 2d)$);
ii) linear classifiers trained on the entire data should have high accuracy
(which excludes most image or text datasets).
A total number of $\numdatasets$ datasets satisfy these criteria and cover a
broad range of applications from finance and ecology to chemistry and histology.
We provide more details about the datasets in
Appendix~\ref{sec:appendix_datasets}.

Like in Section~\ref{sec:theory}, we wish to isolate the effect of
high-dimensionality, 
and hence, we balance the two classes by subsampling the majority class
uniformly at random. Thus we remove confounding effects stemming from applying
M-AL on imbalanced data \citep{ertekin}.
In addition, we mimic the noiseless setting considered in
Section~\ref{sec:theory} using the following procedure: 
after fitting a linear classifier on the entire dataset, we remove the training
samples that are not correctly predicted and use the subsequent smaller subset
as the new dataset. 
We show that, even in the more favorable noiseless setting, 
M-AL is less efficient than PL in high dimensions.
Experiments on the original, uncurated datasets presented in
In Appendix~\ref{sec:appendix_dirty_experiments} reveal a similar trend.


\subsection{Methodology}

\vspace{-0.1cm}
We split each dataset into a test and training set. In all
experiments, we sample a labeled seed set $\Dseed$ of fixed size $\nseed = 6$
from the training set (see Appendix~\ref{sec:appendix_larger_seed} for
experiments with larger $\nseed$). The covariates of the remaining training
samples constitute the unlabeled set $\Dunl$.  \looseness=-1

In practice, one seeks to find the sampling strategy that performs best for a
fixed seed set $\Dseed$ and labeling budget $\nlab$. To provide an extensive
experimental analysis, in this work we compare M-AL (Algorithm~\ref{algo:us})
and PL over a large number of configurations of $(\Dseed, \nlab)$.
We repeatedly draw different seed sets uniformly at random (10 or 100 draws,
depending on the experiment) and consider all integer values in $[\nseed, d/4]$
as the labeling budget $\nlab$, where $d$ is the ambient dimension of the
dataset.\footnote{The value $d/4$ is chosen only for illustration purposes.
  Since for the \emph{real-sim} dataset $d>20,000$, we set the maximum
  labeling budget to $\nlab = 3,000 < d/4$ for computational reasons.
}

At each querying step, we use L-BFGS \citep{liu89} to train a linear classifier
by minimizing the average logistic loss on the labeled dataset collected until
then.  Appendix~\ref{sec:appendix_reg} shows the same high-dimensional
phenomenon for $\ell_1$- or $\ell_2$-regularized classifiers.

\vspace{-0.1cm}
\subsection{Evaluation metrics}

\vspace{-0.1cm} We compare margin-based active learning and passive learning
with respect to two performance indicators. On the one hand, we measure the
probability that PL leads to a smaller test error than M-AL. The probability is
over repeated trials with different seed samples. We compute this probability
for each labeling budget $\nlab \in [\nseed, d/4]$.
On the other hand, we wish to quantify the magnitude of the failure of M-AL. We
compare the most significant gains of M-AL with its most significant losses
across small query budgets for which PL outperforms M-AL with high probability.
In particular, we focus on query budgets smaller than the dataset-dependent
transition point $\ntrans$, where
$\ntrans$ is the largest query budget $\nlab \in [\nseed, d/4]$ for which the
probability of PL outperforming M-AL exceeds $50\%$. If no such budget exists,
$\ntrans = d/4$.
For each query budget size $\nlab \in \{\nseed, ..., \ntrans\}$, we then compute
the gap between the test error obtained with M-AL and with PL, over $100$ draws
of the seed set. For every labeling budget $\nlab$ we report the largest loss of
M-AL (i.e.\ $95^{th}$ percentiles over the draws of $\Dseed$ of the error gap
$\testerr{\thetaunc}-\testerr{\thetaunif}$) and the largest gain of M-AL (i.e.\
$95^{th}$ percentiles of the error gap
$\testerr{\thetaunif}-\testerr{\thetaunc}$).
Then, we depict the distribution of these values of extreme losses/gains over
$\nlab \in \{\nseed, ..., \ntrans\}$. In
Appendix~\ref{sec:appendix_error_vs_budget}
and~\ref{sec:appendix_fraction_budgets}, we present more evaluation metrics
(e.g.\ the dependence of the test error on the budget $\nlab$), which provide
further evidence that M-AL fails to be effective in high dimensions.
\looseness=-1

\begin{figure*}[t]
  \centering
  \begin{subfigure}[t]{0.7\textwidth}
    \centering
    \includegraphics[width=\textwidth]{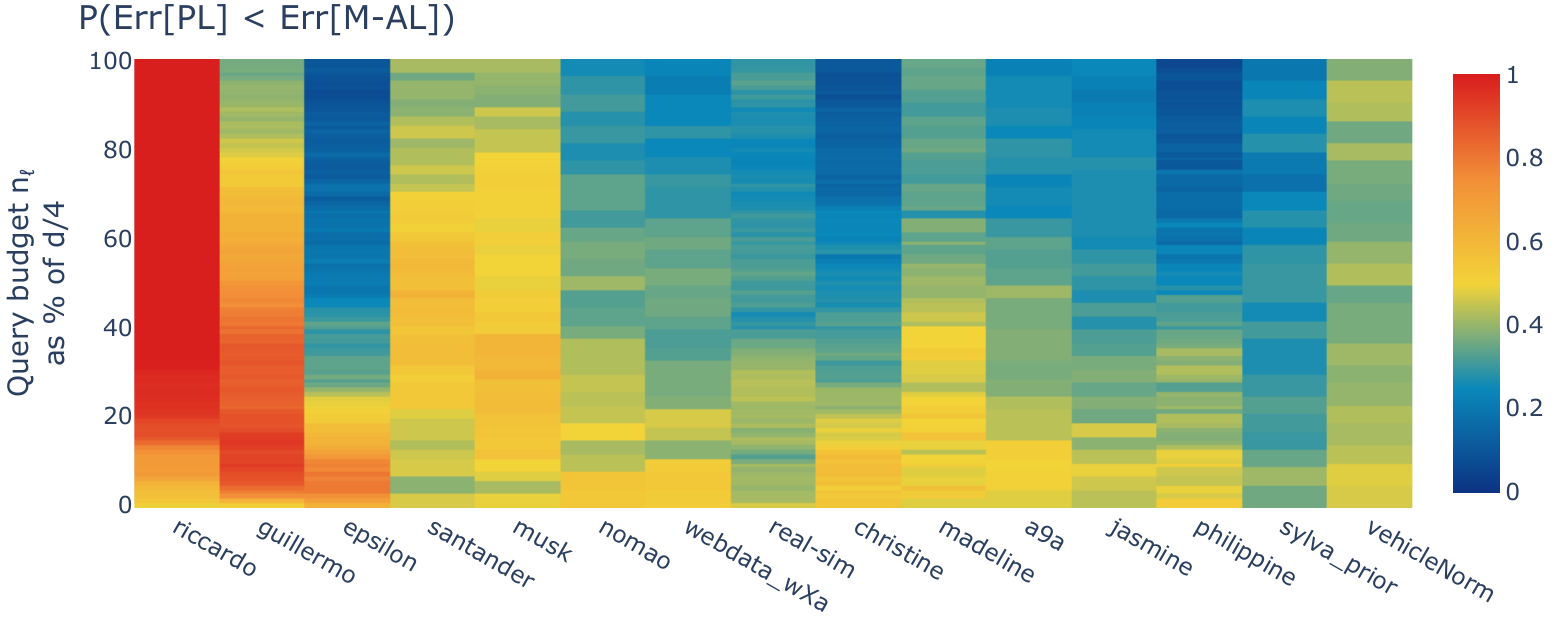}
  \end{subfigure}\\[-0.13cm]
  \begin{subfigure}[t]{0.7\textwidth}
    \centering
    \includegraphics[width=\textwidth]{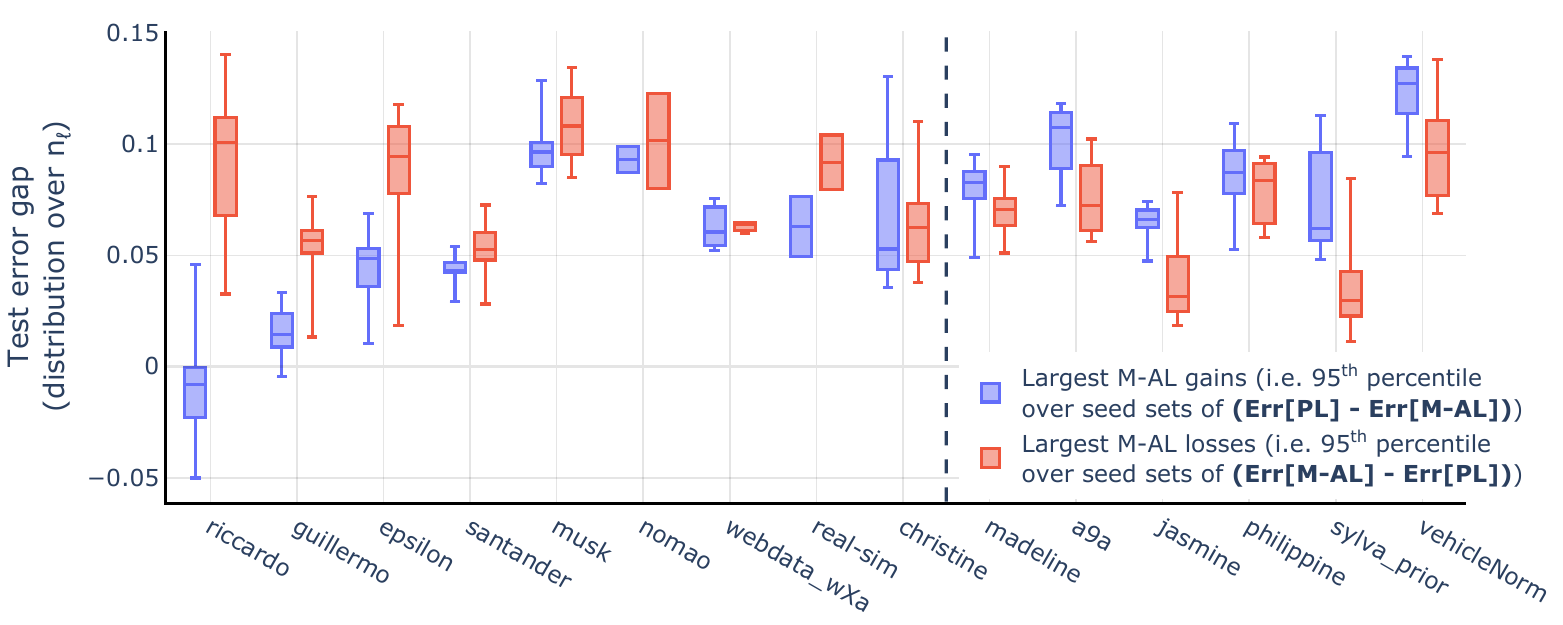}
  \end{subfigure}

  \vspace{-0.2cm} \caption{\small{\textbf{Top:} The probability that the test
      error is lower with PL than with M-AL, over $100$ draws of the seed set.
      PL outperforms M-AL for a significant fraction of $\nlab \in [\nseed,
      d/4]$ (i.e.\ warm-colored regions). See
      Appendix~\ref{sec:appendix_p_al_worse} for more precise numerical values.
      \textbf{Bottom:} Largest gains and losses in test error of M-AL versus PL,
      over $100$ draws of $\Dseed$.  Box plots show distribution over $\nlab \in
      [\nseed, \ntrans]$.
  The sporadic gains of M-AL over PL are generally lower (to left of dashed
  line) or similar to the losses in test error that it can incur.  }}

  \label{fig:main_exp}
\end{figure*}

\vspace{-0.1cm}
\subsection{Main results}
\label{sec:main_results}

In Figure~\ref{fig:main_exp}-Top we show the probability (over
$100$ draws of the seed set) that PL leads to lower test error than M-AL.
The observations match the trend predicted by our theoretical results (see
Theorem~\ref{thm:uncertaintysampling}): decreasing the seed set ratio
$\passiveratio$ (here, by increasing the budget $\nlab$ along the y-axis) leads
to a higher probability that PL outperforms M-AL.
Analogous to the discussion in Section~\ref{sec:proof_and_intuition}, we observe
two regimes:
for small query budgets $\nlab$, M-AL performs poorly with probability larger
than $50\%$ (warm-colored regions). This regime spans a broad range of budgets $\nlab
\in [\nseed, d/4]$ for most datasets.
In the second regime, M-AL eventually outperforms PL for large query
budgets.\footnote{The \emph{riccardo} dataset is particularly challenging for
M-AL and needs more than $d/4$ labeled samples to close the error gap to PL.}
\looseness=-1

Furthermore, Figure~\ref{fig:main_exp}-Bottom shows that in 9 out of 15
datasets,
the median (over budgets $\nlab \in [\nseed, \ntrans]$) of the largest gain of
M-AL is lower than the median loss it can incur, compared to PL.
Intuitively, this indicates that even in the unlikely event that M-AL leads to
better accuracy, the largest gains we can achieve are lower than the potential
losses. 
While the severity of the phenomenon varies with the dataset, we can conclude
after this extensive study
that margin-based AL cannot be used reliably when the dimension of the
data exceeds the size of the query budget.

Finally, recall that in Section~\ref{sec:theory} and in Figure~\ref{fig:teaser}
we show theoretically and empirically that using the margin of the
Bayes optimal classifier exacerbates the failure of M-AL in high dimensions. In
fact, oracle M-AL performs consistently much worse than both PL and vanilla M-AL
in experiments, as indicated in Appendix~\ref{sec:appendix_heatmap_oracle}.


\subsection{Verifying trends predicted by theory for M-AL}
\label{sec:exp_verify_theory}

\vspace{-0.2cm}
The intuition developed in Section~\ref{sec:theory} suggests that the
performance of M-AL in high dimensions may improve for: i) a
larger separation margin between classes (modeled by $\mean/\stdsig$ in
the theorems); 
or ii) a larger seed set (leading to a larger ratio $\passiveratio = \nseed
/ \nlab$). We test whether these insights
also underlie the phenomenon observed in real-world datasets.

\begin{figure*}[t]
\vspace{-0.3cm}
\centering
  \begin{subfigure}[t]{0.65\textwidth}
    \centering
    \includegraphics[width=\textwidth]{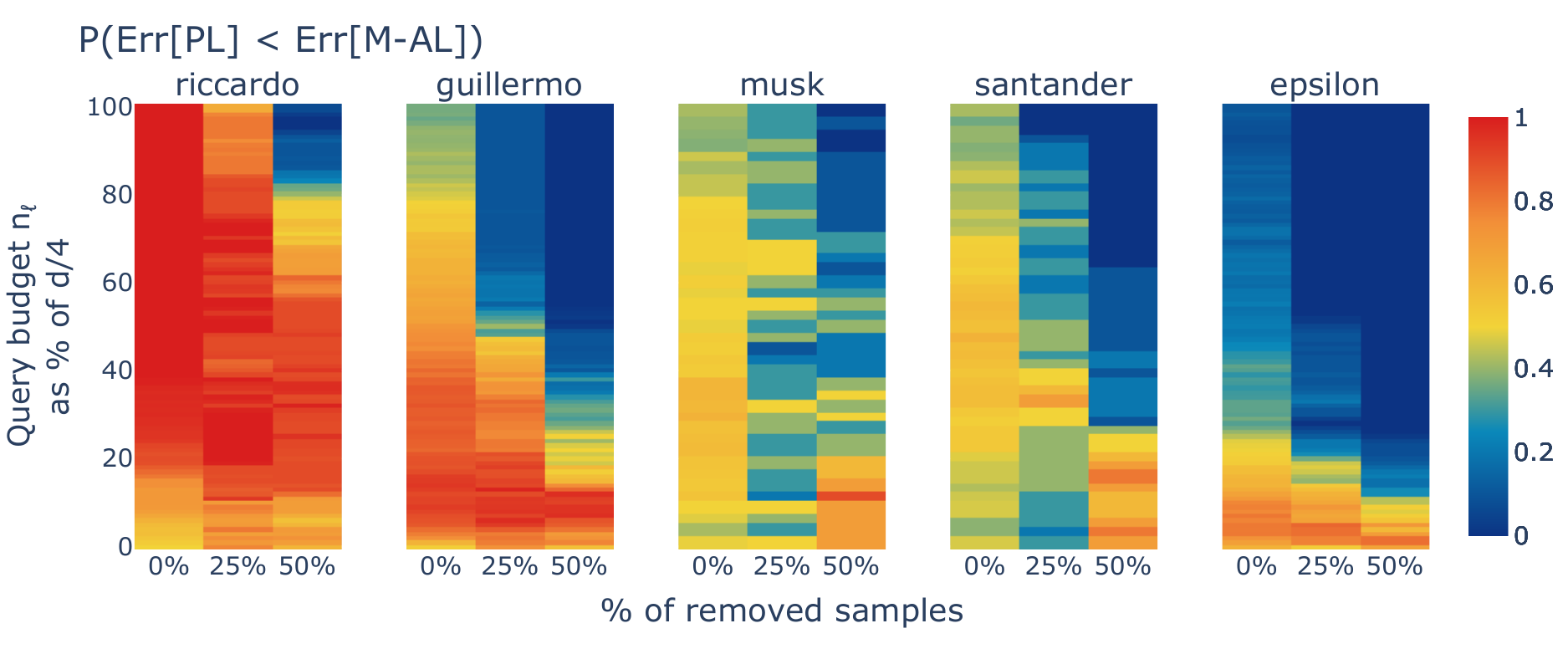}
  \end{subfigure}
  \hfill
  \begin{subfigure}[t]{0.29\textwidth}
    \centering
    \includegraphics[width=\textwidth]{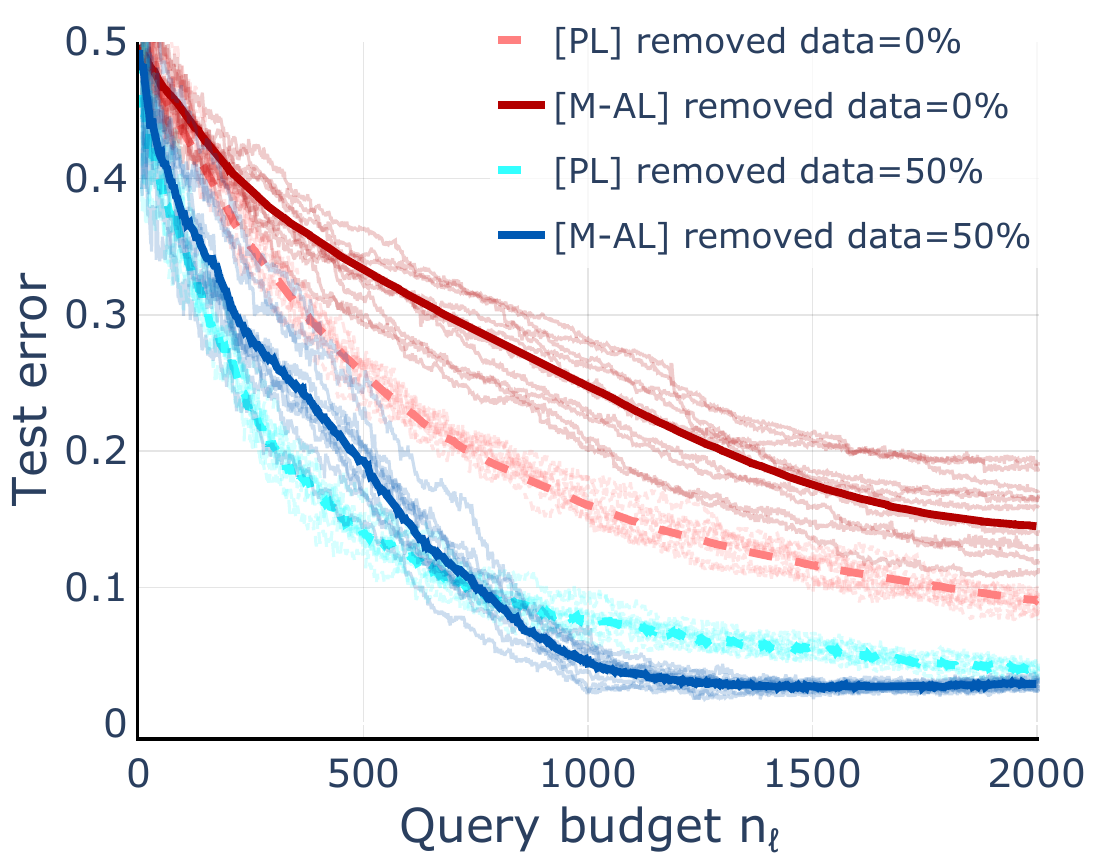}
  \end{subfigure}

\vspace{-0.1cm}
\caption{\small{Increasing the separation between the classes in the unlabeled
    dataset improves the performance of M-AL. Removing the
    $25\%$ or $50\%$ closest points to the Bayes optimal decision boundary improves
    M-AL (\textbf{left}) which now outperforms PL
    for many query budgets (i.e.\ lighter colors), and even on challenging
datasets like \emph{riccardo} (\textbf{right}).}}
  \label{fig:oracle}
  \vspace{-0.2cm}

\end{figure*}

First, we investigate the role of the separation margin. We train a linear
classifier on the full labeled dataset. Then, we artificially increase the
distance between the classes by removing the 25\% or 50\% closest samples to the
decision boundary determined by the classifier trained on the entire dataset.
Indeed, we confirm that removing the 25\% or 50\% most "difficult" samples
improves the performance of M-AL significantly as we show in
Figure~\ref{fig:oracle}, which is in line with the findings of
\citet{sorscher2022}. In particular, after removing the points close to the
Bayes optimal decision boundary, M-AL outperforms PL even on \emph{riccardo},
the most challenging dataset in our benchmark for AL.

Further, we analyze the impact of a larger seed set size on the performance of
M-AL. For a fixed labeling budget $\nlab$, increasing $\nseed$
corresponds to a larger ratio $\passiveratio$, which leads to a smaller gap in
error between M-AL and PL as we explain in Section~\ref{sec:theory}: 
Indeed, we observe that larger seed set sizes can lead to more effective
margin-based AL both on synthetic experiments in
Figure~\ref{fig:synth_seed_size} and on real-world data in experiments presented
in Appendix~\ref{sec:appendix_larger_seed}.

\vspace{-0.3cm}
\subsection{Other AL methods in high dimensions and potential mitigations}
\label{sec:eps_greedy}

\vspace{-0.2cm}
\paragraph{AL strategies effectively equivalent to M-AL.} Finally, we note that
the same failure case occurs for other algorithms, such as margin-based active
learning \citep{Scheffer01, ducoffe18, Mayer20} or entropy sampling
\citep{settles09}, since they effectively sample the same points as M-AL. 


\paragraph{Combining informativeness and representativeness.} Recall that,
by definition, varying the ratio $\passiveratio$ modulates the fraction of the
labeling budget selected with margin-based sampling, with $\passiveratio=1$
corresponding to PL.  Another  way to interpolate between M-AL and
PL is via a strategy that combines informativeness (via
margin-based sampling) and representativeness (via uniform sampling), as proposed
by \citet{brinker03, huang14, yang15, gal17, shui20, farquhar2021} We analyze an
$\epsilon$-greedy scheme that also falls in this family of AL algorithms: at
each querying step, we perform margin-based sampling with probability $1-\eps$ and
uniform sampling with probability $\epsilon$.
In Appendix~\ref{sec:appendix_eps_greedy} we provide evidence that PL continues
to surpass AL for a large fraction of the labeling budgets, even when using the
$\epsilon$-greedy strategy. The intuition for the failure of these algorithms in
high dimensions is the same as the one presented in Section~\ref{sec:theory}.

\paragraph{AL with no margin-based sampling.} Could it be that AL algorithms not
relying on any form of margin-based score, such as \citet{sener18, gissin19,
Hacohen2022}, mitigate this high-dimensional phenomenon? Indeed, we show
experimentally in Figure~\ref{fig:coreset_vs_unc} that coreset-based AL
\citep{sener18} outperforms M-AL with high probability in real-world
applications. However, compared to \emph{PL}, the coreset method is still often
worse (Figure~\ref{fig:coreset_vs_unif}), that is, the high-dimensional
phenomenon outlined in Section~\ref{sec:theory} still persists to a large
extent. In particular, no mechanism prevents the coreset strategy from selecting
points close to the Bayes optimal decision boundary. As highlighted in
Figure~\ref{fig:oracle}, selecting these ``difficult'' samples can hurt the
performance of AL.

%

\paragraph{Discussion on mitigations.} Figure~\ref{fig:oracle} suggests that
M-AL constrained to points far enough from the Bayes optimal
decision boundary might outperform PL in high dimensions. As the
Bayes optimal predictor is not available during training, the closest derived
mitigation strategy would be to not allow selecting 
the points closest to the decision boundary determined by the empirical
estimator $\thetahat$ (instead of the optimal $\thetastar$).
We find that this mitigation strategy does not help to alleviate the negative
effect of M-AL, as it does not effectively remove all the
difficult points from the set of query candidates.  We regard it as important
future work to investigate whether an AL strategy can be provably
effective in high-dimensional settings similar to ours.

\vspace{-0.3cm}
\section{Discussion and future work}

\vspace{-0.2cm}
In this work we show theoretically and through extensive experiments that active
learning, and in particular margin-based sampling, performs worse than uniform
sampling for \emph{linear models} in high dimensions. While we focus on logistic
regression and the max-$\ell_2$-margin solution, we conjecture that the same
intuition outlined in Section~\ref{sec:theory} holds for other linear predictors
like lasso- or ridge-regularized estimators, as indicated by experiments in
Appendix~\ref{sec:appendix_reg}.

Moreover, this phenomenon is more general and also occurs for complex non-linear
models like deep neural networks. Our experiments suggest that M-AL performs
poorly in the context of deep learning on a number of different image
classification tasks (see Appendix~\ref{sec:appendix_image}), corroborating
previous findings by \citet{sorscher2022, Hacohen2022}. We leave as future work
an investigation of whether the insights revealed by our analysis of linear
models transfers to non-linear predictors like deep neural networks. 

Furthermore, for practical purposes, an important question for future work is
whether it is possible to construct a strategy that improves upon uniform
sampling in high dimensions.
Based on Figure~\ref{fig:oracle}, we believe that imposing certain conditions 
on the distributions (e.g.\ vanishing mass close to the decision boundary) could
allow for improvements via active learning.
\looseness=-1


\vspace{-0.4cm}
\section*{Acknowledgements}
\vspace{-0.3cm}
AT was supported by a PhD fellowship from the Swiss
Data Science Center. JC was supported by the Hasler Foundation grant number
21050. We are grateful to Andreas Kirsch, Stephen Mussmann and Ilija Bogunovic
for feedback on the manuscript. We also thank the anonymous reviewers for their
helpful remarks. 

\newpage

\bibliography{al_arxiv.bib}
\bibliographystyle{abbrv}

\newpage

\appendix
\section{Discussion of assumptions for the theory}
\label{sec:appendix_assumptions}

In this section we motivate some of the assumptions made in
Section~\ref{sec:theory}. We argue why these assumptions are not too
constraining and are applicable to a wide variety of practically relevant
settings.

\subsection{Extension to arbitrary Bayes optimal classifier}
\label{sec:appendix_rotation}

In Section~\ref{sec:theory} we introduce the data distribution and consider
noiseless labels determined by a Bayes optimal classifier that takes the form
$\thetastar = [1, 0, ..., 0] \in \sphere^{d-1}$, without loss of generality. We
stress that we only make this choice to ease the exposition and make the
intuition behind the high-dimensional phenomenon more clear.  The distinction
between the signal and non-signal components is merely used to make it easier to
grasp why this high-dimensional phenomenon occurs.  All the arguments in our
proofs hold true for arbitrary Bayes optimal classifiers, including a possibly
rotated $\thetastar$. A simple way to see this is to notice that the intuition
in Figure~\ref{fig:synth_sketch} holds true even if we apply a rotation to the
data and to $\thetastar$: M-AL still tends to sample points close to the Bayes
optimal decision boundary (unlike PL), and hence, the classifier trained on this
labeled set will likely be very tilted compared to $\thetastar$. 

\subsection{Balanced data assumption}
\label{sec:appendix_balanced}

In imbalanced classification problems, it is known that margin-based active
learning tends to sample a more balanced labeled set than uniform sampling
\citep{ertekin}. As a consequence, the classifiers trained on these more
balanced labeled data tend to have better predictive performance. This
phenomenon occurs both in the low-dimensional and in the high-dimensional
regimes.

In our analysis, we want to disentangle effects caused by data imbalance (as the
one described in \citet{ertekin}) from phenomena specific to the
high-dimensional regime. Therefore, we consider balanced data for most of our
analysis. In Appendix~\ref{sec:appendix_dirty_experiments} we also provide
experiments on imbalanced tabular data and see that the failure of M-AL in high
dimensions still occurs, albeit at a lesser extent. We confirm in our
experiments that even in this very low sample regime, M-AL tends to select a
more balanced labeled set.

\subsection{Gaussian mixture models}

As we argue in Section~\ref{sec:datadistr}, GMMs are known to model well data
generated in numerous practical applications \citep{bouguila19}, and hence, have
often been studied in theoretical analyses of machine learning algorithms
\citep{tsipras2018, mahdi20, donhauser21}. Note that the GMM assumption is not
critical for the proofs and similar results can be obtained for other more
general distribution families (e.g.\ sub-exponential)

\subsection{M-AL fails even in beneficial settings}

In this work we characterize a failure of margin-based active learning.
In order to show the extent of this failure case, we wish to prove that it
occurs even in scenarios that are believed to be beneficial for active learning.
In particular, we consider data with noiseless labels and show that the failure
occurs not only for regular M-AL (Algorithm~\ref{algo:us}), but even for M-AL
that uses the margin of the Bayes optimal classifier for sampling.

\subsubsection{Noiseless data}

M-AL tends to select points that are close to the Bayes optimal decision
boundary. For many label noise models (e.g.\ logistic noise) this is exactly the
region of the input space where the noisy data is concentrated. This observation
is also true for many practical applications, for instance when label noise is
caused by ambiguities between the classes (e.g.\ an image that could be assigned
either the class ``wolf'' or ``dog'' because the object is not clear).
Therefore, M-AL tends to be vulnerable to wasting the limited query budget on
acquiring labels that are likely to be incorrect. In contrast, PL samples
uniformly from the data distribution and may be able to overcome this issue.

We want to ensure that the failure that we study in this paper is not linked to
the propensity of M-AL to select noisy samples for labeling. Hence, we consider
noiseless data in our analyses. Having said that, we point out that our results
can be readily extended to settings with label noise and would lead to a more
severe failure of M-AL.

\subsubsection{Oracle M-AL versus Empirical M-AL}
\label{sec:appendix_oracle_mal}

Since we specifically analyze the failure of M-AL in the \emph{low-sample
regime}, one could argue that this drop in performance is due to the cold start
problem: we sample queries using a classifier trained on very few labeled points
\citep{huang14, sener18, Hacohen2022}. To rule out this explanation, we show
that we identify the same failure for oracle M-AL that uses the Bayes optimal
classifier for sampling. In this section we argue why oracle M-AL performs
better than M-AL with an empirical classifier in low dimensions. This good
performance of oracle M-AL justifies our choice to study it in high dimensions
as well. However, we find that in this latter regime, oracle M-AL fails even
more severely than M-AL as we explain in Section~\ref{sec:theory}.

In practice, active learning algorithms use a subjective notion of what the
informative samples are, connected to the current empirical predictor that can
be trained on the labeled data collected so far. However, the optimal sampling
strategy may depend on the Bayes optimal classifier, which, of course, is
unknown in practical applications. For instance, it follows from results in
\citet{chaudhuri15} that a strategy akin to oracle M-AL is optimal in the
context of maximum likelihood estimators in low dimensions.

We now present another intuitive way to see why oracle M-AL is expected to need
significantly fewer labeled samples in low dimensions, compared to M-AL.
Consider the simple setting of learning thresholds in 1D (i.e.\ functions $f:
[0, 1] \rightarrow \{0, 1\}$ of the form $f_t(x)=\indicator[x>t]$). Assume the
covariates are distributed uniformly in $[0, 1]$ and that the labels are
noiseless, i.e.\ there exists $t^\star$ such that $y=\indicator[x>t^\star]$ for
any $x\in [0, 1]$.

We know from standard results in learning theory that uniform sampling requires
$O(1/\epsilon)$ labeled samples to reach error at most $\epsilon$.
\citet{Balcan07} prove that M-AL needs a much smaller labeled set of only
$O(\log(1/\epsilon))$ to achieve the same test error. In contrast, oracle M-AL
requires only $O(1)$ labeled samples to achieve the same performance: we only
need to select the samples closest to the Bayes optimal decision boundary until
we acquire points from both classes.

This intuitive argument can be extended to dimensions larger than 1 as well.
Empirical observations like the one in \citet{sener18} corroborate this
intuition: oracle M-AL outperforms M-AL in the context of neural networks
trained on image data.

It is important to note that in the discussion in this section we rely on a
labeling budget much larger than the dimensionality. Prior to our work, the
behavior of oracle M-AL has not been studied in the low-sample regime.
Therefore, it is justified to ask how oracle M-AL behaves in the
high-dimensional regime.


\section{Formal statements and proofs of the main results}
\label{sec:appendix_thm}

In this section we state and give the proofs of the main results. We first
discuss some preliminaries regarding the mixture of truncated Gaussians
distribution after which we state the formal results and the proofs.

\vspace{-0.2cm}
\subsection{Properties of a mixture of truncated Gaussians} 
\label{sec:appendix_tgmm}

Recall that we consider data drawn from a multivariate mixture of two Gaussians,
and we partition the covariates into noise dimensions $\xnoise \in \RR^{d-1}$
sampled according to $\PP_{\xnoise} = \gauss(0,I_{d-1})$ and a signal dimension
$\xsig \in \RR$.  As detailed in Section~\ref{sec:datadistr}, the signal $\xsig$
is drawn from a univariate mixture of two Gaussians with means $-\mean$ and
$\mean$ for $\mean \in \RR$ and standard deviation $\stdsig > 0$. The components
correspond to one of two classes $y\in \{-1, 1\}$, and they are truncated such
that the data is noiseless. We denote the univariate truncated Gaussian mixture
distribution by $\ptgmm$ and observe that it determines, by definition, the
joint distribution $\PP_{y\xsig}$. To state the formal theorems, we now discuss
some properties of the univariate distribution $\ptgmm$ that will also be used
throughout the proofs in this section. 


\vspace{-0.2cm}
\paragraph{Mean and standard deviation of a univariate truncated Gaussian.} For
completeness, we now introduce the known formulas for the mean and standard
deviation of a truncated Gaussian random variable. A positive one-sided
truncated Gaussian distribution is defined as follows: we restrict the support
of a normal random variable with parameters $(\mean, \stdsig)$ to the
interval $(0, \infty)$. Clearly the mean of the univariate truncated Gaussian is
slightly larger than $\mean$ and the standard deviation slightly smaller than
$\stdsig$.  Let $\pdf$ be the probability density function of the standard
normal distribution and denote by $\Phi$ the cumulative distribution function.
Then we find that the mean of a positive one-sided truncated Gaussian
distribution is

\begin{equation}
\label{eq:meantruncated}
\mean_{tr} = \mean+\frac{\stdsig\pdf(-\mean/\stdsig)}{1-\cdf(-\mean/\stdsig)}, 
\end{equation}

\noindent and the standard deviation is given by

\begin{equation*}
\label{eq:stddevtruncated}
\stdsig_{tr} =
\stdsig\left(1-\frac{\mean}{\stdsig} \cdot
\frac{\pdf(-\mean/\stdsig)}{1-\cdf(-\mean/\stdsig)}-\left(\frac{\pdf(-\mean/\stdsig}{1-\cdf(-\mean/\stdsig)}\right)^2\right).
\end{equation*}

\paragraph{Error of a linear classifier.} We now derive the closed form of the
population error of a linear classifier evaluated on data drawn from a
distribution with $\PP_{y\xsig}=\ptgmm$ and $\PP_{\xnoise|y}=\gauss(0,
I_{d-1})$.  Consider without loss of generality a classifier induced by a vector
$\theta \in \mathbb{R}^{d}$, with $\theta = [1, \alpha\thetatilde]$, where
$\|\thetatilde\|_2=1$. We use the notation $\alpha(\theta)$ to denote the
$\alpha$-parameter of  $\theta = [1, \alpha\thetatilde]$.
By definition, the error of a classifier $\theta=[1, \alpha\thetatilde]$ is
given by

\begin{equation}
  \begin{aligned}[b]
\label{eq:testerror}
\err(\theta) &= \PP\left[y\langle\theta, x\rangle  < 0\right] = \PP\left[
y\alpha\sum_{i=2}^d \thetatilde_i x_i  < -y\xsig\right]\\
&=\frac{1}{2\pi\alpha
\stdsig(1-\Phi(-\mean/\stdsig))}\int_{0}^{\infty}\int_{t}^{\infty}e^{-\frac{(t-\mean)^2}{2\stdsig^2}}e^{-\frac{l^2}{2
\alpha^2}}dldt \eqqcolon \Psifull(\alpha),
\end{aligned}
\end{equation}

\noindent where we use the fact that the sum of Gaussian random variables is again a
Gaussian random variable, and hence, $y\sum_{i=2}^d \theta_i x_i$ is normally
distributed with mean zero and standard deviation $\alpha$. For the final
identity, since all coordinates are independent, we use the known formula for
the probability density function of a truncated Gaussian.

Note that the expression in Equation~\eqref{eq:testerror} can easily be
approximated numerically. Moreover, note that $\Psifull$ resembles the
cumulative density function of a standard normal random variable, as illustrated
in Figure~\ref{fig:psialpha}: it grows exponentially in $\alpha$, for small values of
$\alpha$, and approaches its maximum asymptotically.
For convenience we state two properties of the error function $\Psifull$:

\begin{enumerate}
\item $\Psifull$ is a monotonically increasing function of $\alpha$.
\item $\Psifull$ is monotonically decreasing in $\mean$.
\end{enumerate}

Therefore, for fixed distributional parameters $\mean$ and $\stdsig$, we have that
$\alpha$ fully characterizes the error of the classifier. Hence, proving a gap
between the values of $\alpha$ obtained with margin-based sampling and with
uniform sampling is sufficient to show a gap in the population error. 

\subsection{Formal statement of Theorem~\ref{thm:mixturesoracle}}
\label{appendix_formal_oracle}

In this section, we state the formal version of
Theorem~\ref{thm:mixturesoracle}. The proof of the theorem can be found in
Section~\ref{sec:appendix_thm_proofs}. We start with formally introducing the setting
and assumptions. Thereafter, we state the theorem that compares oracle
M-AL with PL.

\paragraph{Setting.} We look at the same setting as in Section~\ref{sec:theory},
namely pool based active learning with margin-based sampling strategies. We
consider the typical setting where we start with a small labeled dataset
$\Dseed$ of $\nseed$ samples and a large unlabeled dataset $\Dunl$ with $\nunl$
samples all i.i.d.\ drawn from the truncated Gaussian distribution. We denote
by $\rho = \frac{\nseed}{\nlab}$ the fraction of samples we query using uniform sampling. Recall that
we define $\thetaunif$ and $\thetaor$ as the classifiers obtained after querying
$(1-\passiveratio)\nlab$ samples either uniformly or with oracle M-AL,
respectively. 

We now introduce some assumption on the setting. In practice, an unlabeled
dataset is available that is much larger than the number of queries. Moreover,
most real-world datasets contain some hard examples that are difficult to
classify even by human experts. The equivalent synthetic counterpart is the
existence of unlabeled samples close to the Bayes optimal decision boundary.
Finally, we consider high-dimensional settings, where the dimensionality is
larger than the labeling budget. To state our theorem formally, we make these
conditions precise in the following assumption.

\begin{assumption}
\label{ass:nunl}
We assume that $\nunl > \max(10^3\nlab, 10^5)$, $\mean/\stdsig < 2$ and $d -1> \nlab$.
\end{assumption}
With Assumption~\ref{ass:nunl}, we are now ready to state the theorem.

\begin{theorem}[Oracle margin-based sampling]
\label{thm:oracleuncertainty}
For a small constant $c > 0$ independent of the
dimension and under Assumption~\ref{ass:nunl}, it holds with
probability at least $0.99(1-2e^{-t^2/2})^5(1-e^{-c d/\nlab})$
that:
\begin{equation*}
\label{eq:ortestbound}
\testerr{\thetaor} - \testerr{\thetaunif} \geq \Psifull(\alphaor) -
\Psifull(\alphaPL),
\end{equation*}
where
\begin{align*}
  \alphaPL = \frac{\sqrt{\frac{d-1}{\nlab}+\frac{2t}{\sqrt{\nlab}}}}{\mean_{tr}
  -t\stdsig_{tr}\nlab^{-1/2}}, &&
  \alphaor = \frac{\sqrt{\frac{d-1}{\nlab}}-1-t}{\passiveratio (\mean_{tr}+t
  \stdsig_{tr}(\rho\nlab)^{-1/2})+ (1-\passiveratio)6.059 \cdot 10^{-2}
\mean_{tr}}-1.
\end{align*}
\end{theorem}

We note that $\alphaPL, \alphaor$ are upper
and lower bounds of $\alpha(\thetaunif)$ and $\alpha(\thetaor)$, respectively.
The term $(1-e^{-c d/\nlab})$ lower bounds the probability that all
points of each sampling strategy are support points (see
Lemma~\ref{lem:SPuniform} for the explicit probability) -- in particular, as
the ratio $\frac{d}{\nlab}$ grows, this probability grows.  Another consequence
of Theorem~\ref{thm:oracleuncertainty} is that for high-dimensional data (i.e.\
$d \gg \nlab$) and for a small ratio of uniformly sampled seed points
$\passiveratio \ll 1$, it holds with high probability that $\testerr{\thetaor} -
\testerr{\thetaunif}>0$. We state this observation precisely in
Corollary~\ref{cor:oracleuncertainty} and prove it in
Section~\ref{sec:cororacleproof}. To state the corollary, we first define a
quantity that is independent of the dimension. Denote by $\denomor$ the denominator
of $\alphaor$, i.e.  $\denomor = \passiveratio(\mean_{tr}+t \stdsig_{tr}(\rho \nlab)^{-1/2})+(1-\passiveratio)6.059 \cdot 10^{-2}\mean_{tr}$.
Note that $\denomor \ll \mean_{tr}$. We are now ready to state the corollary.

\begin{corollary}
\label{cor:oracleuncertainty}
Under the same assumptions and with the
same probability as in Theorem~\ref{thm:oracleuncertainty} it holds that
$\testerr{\thetaor} - \testerr{\thetaunif} >0$
if 
\begin{align*}
  \frac{d-1}{\nlab} > 4\left(1+\denomor+t+\sqrt{t(4\nlab)^{-1/2}}\right)^2, &&
 \passiveratio < \frac{1}{2}-\frac{1+\sqrt{2}}{2} \frac{t \stdsig_{tr}}{\sqrt{\nlab} \mean_{tr}}.
\end{align*}
\end{corollary}

The condition on $\passiveratio$ ensures that enough samples are queried using
oracle M-AL such that the difference to passive learning is
large enough.

We obtain the first informal statement in Theorem~\ref{thm:mixturesoracle}
directly from Theorem~\ref{thm:oracleuncertainty}  by choosing 
\begin{equation}
  \label{eq:epsilonexpl}
\epsilon = \max \left( \sqrt{\frac{\nlab}{d-1}} (1+t), \sqrt{1+\frac{2 t
\nlab^{1/2}}{d-1}}-1 \right),
\end{equation}
and the constants $c_1,c_2$ correspondingly.
The second statement then follows from Corollary
\ref{cor:oracleuncertainty}.
%

\subsection{Formal statement of Theorem~\ref{thm:mixtureUS}}

In this section, we state the formal version of Theorem~\ref{thm:mixtureUS}
which shows an error gap between passive learning and active learning using the
margin of the empirical classifier $\thetahat$. Before we state the
theorem, we first discuss two-stage margin-based sampling, a slight modification
of Algorithm~\ref{algo:us}.

\paragraph{Two-stage margin-based sampling.}
We consider the same modification of the margin-based sampling procedure as
\citep{chaudhuri15, mussmann18}.
Instead of the iterative process of labeling a point and updating the estimator
$\thetahat$, we use a two-stage procedure: 1) we obtain $\thetaseed$ using the
initial small seed set; and 2) we use $\thetaseed$ to select a batch of
$(1-\passiveratio)\nlab$ samples to query from the unlabeled set.  Without this two-stage
strategy, the estimator $\thetahat$ at a certain iteration is not independent of
the unlabeled set, which makes the analysis more challenging, as also noted by
\citep{mussmann18}. Moreover, \citet{chaudhuri15} show that a two-stage strategy
similar to ours achieves the optimal convergence rate in the context of maximum
likelihood estimators, and hence, it is no worse than the iterative procedure in
Algorithm~\ref{algo:us}.
We stress that we do not need this simplification for the analysis of oracle
M-AL, since with this strategy the queried points are
independent of the estimators $\thetahat$. Moreover, the two-stage procedure is
necessary only for one step of the proof highlighted in
Section~\ref{sec:lemalphabounds}.

We now state the main theorem for empirical M-AL:

\begin{theorem}
\label{thm:uncertaintysampling}
For a small constant $c > 0$ independent of the dimension and under
Assumption~\ref{ass:nunl} with $\stdsig>1$, it holds with probability at least
$0.99(1-2e^{-t^2/2})^5(1-e^{-c d/\nlab})$ that:
\begin{equation*}
\testerr{\thetaunc} - \testerr{\thetaunif} \geq \Psifull(\alphaUS) - \Psifull(\alphaPL),
\end{equation*}
where $\alphaPL$ is defined as in Theorem~\ref{thm:oracleuncertainty} and
\begin{equation*}
  \alphaUS = \frac{\sqrt{\frac{d-1}{\nlab}}-\sqrt{2\log \nunl} - 1-t}{\rho\Cseed +
(1-\rho)\left( 0.061 \mean_{tr}+ \sqrt{\frac{2 \log
\nunl}{\Cseed}}\left(\frac{d-1+\stdsig_{tr} t}{\rho
\nlab}\right)^{1/4}+t\right)}-1,
\end{equation*}
with $\Cseed$ a constant that satisfies
\begin{equation*}
  \mean_{tr}-t \stdsig_{tr}\nseed^{-1/2} \leq \Cseed \leq
  \mean_{tr}+t\stdsig_{tr}\nseed^{-1/2} .
\end{equation*}
\end{theorem}

Theorem~\ref{thm:uncertaintysampling} gives a high probability bound for the
error gap between passive learning and two-stage empirical margin-based
sampling.
As before, we state precise conditions when this gap is positive in
Corollary~\ref{cor:cortwostageproof} and provide the proof in
Section~\ref{sec:cortwostageproof}. Similarly as for Corollary
\ref{cor:oracleuncertainty}, we first define a quantity. Let $\denomuncer$ be
the denominator of $\alphaUS$ and recall that $\denomuncer \ll \mean_{tr}$ if
the max-$\ell_2$-margin classifier of the seed set $\Dseed$ has reasonable
accuracy. 

\begin{corollary}
\label{cor:cortwostageproof}
Under the same assumptions and with the same probability as in
Theorem~\ref{thm:uncertaintysampling} it holds that 
$\testerr{\thetaor} - \testerr{\thetaunif} >0$ if the following conditions are
satisfied:
\begin{enumerate}

  \item (high-dimensional regime) $\frac{d-1}{\nlab} > 4\left(\sqrt{2 \log \nunl}
      +1+\denomuncer +t
    +\sqrt{t}(4\nlab)^{-1/4} \right)$.

  \item (large signal-to-noise ratio) $\mean_{tr}  \geq
    \left(\frac{d+\stdsig_{tr} t}{\passiveratio \nlab}\right)^{1/6}(\log \nunl)^{1/3} +
    \frac{t\stdsig_{tr}}{(\passiveratio \nlab)^{1/2}}$.

  \item (numerous margin-based queries) $ \passiveratio <   \frac{2  \Cseed}{0.878 \mean_{tr}- \frac{t \stdsig_{tr}}{(\passiveratio\nlab)^{1/2}}
 - \left(\frac{d-1+\stdsig_{tr} t}{\passiveratio \nlab}\right)^{1/6}(\log \nunl)^{1/3} -t}$.

\end{enumerate}
\end{corollary}

The second condition is necessary to ensure that the classifier $\thetaseed$
trained on the seed set has non-trivial error. Like in
Section~\ref{appendix_formal_oracle}, the third condition guarantees that the
influence of the uniformly sampled seed set is reduced. To get an explicit condition on $\rho$ on the right hand side, we note that $\passiveratio \nlab \geq 2$ by definition.

Furthermore, similar to Section~\ref{appendix_formal_oracle}, the term $(1-e^{-c
d/\nlab})$ lower bounds the probability that all points are support points for
each of the two sampling strategies. The explicit expression for this
probability can be found in Lemma~\ref{lem:SPuniform}.

Finally, observe that Theorem~\ref{thm:uncertaintysampling} and Corollary
~\ref{cor:cortwostageproof} are together the formalization of
Theorem~\ref{thm:mixtureUS}. More specifically, by setting $\epsilon$ similar as
in Equation~\eqref{eq:epsilonexpl} and considering large $d/\nlab$, we find the
informal statement.

\subsection{Proofs of Theorems~\ref{thm:oracleuncertainty} and~\ref{thm:uncertaintysampling}}
\label{sec:appendix_thm_proofs}

Recall that, without loss of generality, we consider predictors $\theta=[1,
\alpha \thetatilde]$, with $\|\thetatilde\|_2 = 1$, which makes the population
error of $\theta$ be a strictly increasing function of $\alpha$. Therefore, to
prove Theorems~\ref{thm:oracleuncertainty} and~\ref{thm:uncertaintysampling} we
derive bounds on $\alpha$ for uniform and oracle/empirical margin-based sampling.
We split the proof into three main steps.
In the first step we bound the $\alpha$-parameter of the max-$\ell_2$-margin
classifier of a dataset obtained using an arbitrary sampling strategy. The
bounds we obtain are a function of certain geometric quantities that we
introduce in this section. The second step then bounds these geometric
quantities for the specific sampling strategies that we are interested in.
Lastly, in the third step we develop these bounds further for the special case
of a mixture of truncated Gaussians.  Our results also hold for a marginal Gaussian
distribution that is usually analyzed in the active learning literature
\citep{Hanneke13}.

We would like to reemphasize that we focus on separable data, a setting
that benefits active learning.  As described in
Section~\ref{sec:datadistr}, we consider a Bayes optimal predictor
$\thetastar$ with vanishing population error and choose without loss
of generality $\thetastar = e_1 = [1, 0, ..., 0] \in
\RR^d$.\footnote{If $\thetastar \neq e_1$, we can rotate and translate the data
in order to get $\thetastar=e_1$.} We can write the covariates as $x = [\xsig,
\xnoise]$, where we explicitly separate the coordinates of $x$ into a signal
$\xsig \in \RR$ and non-signal component $\xnoise \in \RR^{d-1}$.  The marginal
distribution of the covariates takes the form $\PP_{X}=\PP_{\xsig} \cdot
\PP_{\xnoise}$, where $\PP_{\xnoise} = \gauss(\xnoise; 0, I_{d-1})$ is the
distribution of the non-signal dimensions.

We point out that the first two steps of the proof of
Theorems~\ref{thm:oracleuncertainty} and~\ref{thm:uncertaintysampling} hold for
any arbitrary distribution $\PP_{y\xsig}$.
If the joint distribution $\PP_{y\xsig}$ is a univariate mixture of truncated
Gaussians, then $\PP_{y\xsig} = \ptgmm$, which in turn corresponds to a marginal
Gaussian for $\mu \to 0$.
\paragraph{Step 1: Characterizing $\thetahat$ for arbitrary sampling
strategies.}
To state the key lemma that characterizes the max-$\ell_2$-margin solution
$\thetahat$ for any labeled dataset $\Dlab \subset \RR^d \times \{-1, 1\}$  of
size $\nlab$, we first introduce three geometric quantities.
Recall that we consider classifiers of the form $\theta = [1, \alpha
\thetatilde]$ with $\|\thetatilde\|_2 = 1$. We denote the max-$\ell_2$-margin of
$\Dlab$ in the last $d-1$ coordinates by $\margin$, and write:

\begin{equation}
  \label{eq:margin}
  \margin = \max_{\thetatilde \in \sphere^{d-2}} \min_{(x, y) \in \Dlab} y\langle
\xnoise, \thetatilde\rangle.
\end{equation}

Note that $\margin$ is in fact the maximum min-$\ell_2$-margin of $\Dlab$ in the
last $d-1$ coordinates.
Similarly, the maximum average-$\ell_2$-margin of $\Dlab$ in the last $d-1$
coordinates is defined as

\begin{equation}
  \label{eq:avgmargin}
  \margin_{avg} = \max_{\thetatilde \in \sphere^{d-2}} \frac{1}{\nlab} \sum_{(x,y)
\in \Dlab} y\langle \xnoise, \thetatilde\rangle.
\end{equation}

Lastly, we define the average distance to the decision boundary of the optimal
classifier induced by $\thetastar$ as

\begin{equation*}
  \dstar = \frac{1}{\nlab} \sum_{(x, y) \in \Dlab} y x_{1}.
\end{equation*}

We now state the lemma that bounds the parameter $\alpha$ of the
max-$\ell_2$-margin classifier trained on an arbitrary labeled set. We provide
the proof of the lemma in Section~\ref{app:proof_optimalclassifier}.

\begin{Lemma}[Bound on the max-$\ell_2$-margin classifier for active learning]
\label{lem:optimalclassifier}

Let $\Dlab$ be a labeled set such that all samples are support
vectors of the max-$\ell_2$-margin classifier $\thetahat$ of $\Dlab$. Then the
$\alpha$-parameter of $\thetahat$ is bounded as follows:

\begin{equation*}
\frac{\margin}{\dstar} -1 \leq \alpha \leq \frac{\margin_{avg}}{\dstar}.
\end{equation*}
\end{Lemma}

Once equipped with Lemma~\ref{lem:optimalclassifier}, the next step is to derive
bounds on $\margin$, $\margin_{avg}$ and $\dstar$ for uniform and
oracle/empirical margin-based sampling.

\paragraph{Step 2: Bounding $\dstar$, $\margin$ and $\avgmargin$ for specific
sampling strategies.} In this step, we derive concrete bounds for the key
quantities in Lemma~\ref{lem:optimalclassifier} for specific sampling
strategies, namely uniform and oracle/empirical margin-based sampling. 
For this purpose we introduce further geometric quantities that now also depend
on the seed set $\Dseed$.
First, we denote by $\dqstar$ the maximal distance of the newly sampled queries
to the decision boundary of the Bayes optimal classifier $\thetastar$:

\begin{equation*}
\dqstar = \max_{(x,y) \in \Dlab \setminus \Dseed}  y x_{1}.
\end{equation*}

Furthermore, let $\thetaseed$ be the parameter vector of the max-$\ell_2$-margin
classifier of the seed set with $\alpha$-parameter $\alpha_{seed}$. We define
$\dqhat$ as the maximal distance of the newly queried points to the decision
boundary determined by $\thetaseed$:

\begin{equation*}
\dqhat = \max_{(x,y) \in \Dlab \setminus \Dseed}  
\frac{\left|\langle\thetaseed, x\rangle\right|}{\| \thetaseed\|_2}. 
\end{equation*}
Lastly, we define $\Cseed$ as the average distance to the decision boundary of $\thetastar$ of the samples in the seed set:
\begin{equation*}
\Cseed = \frac{1}{\nseed} \sum_{(x, y) \in \Dseed} y x_{1}.
\end{equation*}

Finally, recall that in pool-based active learning one has access to an
unlabeled set $\Dunl$ drawn i.i.d.\ from $\PP_X$ and a small labeled seed set
$\Dseed$ where the covariates are drawn i.i.d.\ from $\PP_{XY}$.
We collect a labeled set $\Dlab$ that includes the uniformly sampled $\Dseed$
and $(1-\passiveratio)\nlab$ labeled points whose covariates are selected from
$\Dunl$ according to a sampling strategy. We are now ready to state the
following lemma, which bounds the quantities that show up in
Lemma~\ref{lem:optimalclassifier}, namely $\dstar$, $\margin$ and $\avgmargin$.
The proof of the lemma is presented in Section~\ref{sec:lemalphabounds}.

\begin{Lemma}[Bounds on $\dstar$, $\margin$ and $\avgmargin$]
\label{lem:alphabounds}
Consider the standard pool-based active learning setting in which we collect a
labeled set $\Dlab$ and assume $\nlab < d-1 < \nunl$ where $\nlab=|\Dlab|$ and
$\nunl = |\Dunl|$. Then, the following are true:

\begin{enumerate}
\item If $\Dlab$ is collected using \thmemph{uniform sampling}, then with a
  probability greater than $1-2e^{-t^2/2}$, it holds that
\begin{equation}
  \label{eq:alphaunif}
\begin{aligned}
\dstar = \frac{1}{\nlab} \sum_{(x, y)\in\Dlab} y\xsig,&&
\avgmargin < \sqrt{\frac{d-1}{\nlab} + \frac{2 t}{\sqrt{\nlab}}}.
\end{aligned}
\end{equation}
\item If $\Dlab$ is collected using \thmemph{oracle margin-based sampling}, then
  with probability greater than $1-2e^{-t^2/2}$ it holds that
\begin{equation}
  \label{eq:alphaoracle}
\begin{aligned}
  \dstar < \passiveratio \Cseed+(1-\passiveratio)\dqstar, &&
  \margin > \sqrt{\frac{d-1}{\nlab}} -1 -t.
\end{aligned}
\end{equation}
\item If $\Dlab$ is collected using \thmemph{two-stage empirical margin-based 
  sampling}, then with probability greater than $(1-2e^{-t^2/2})^2$, it holds that
\begin{equation}
  \label{eq:alphauncert}
\begin{aligned}
  \dstar  < \passiveratio \Cseed + (1-\passiveratio)(\dqhat + \sqrt{2
  \alpha_{seed}\log \nunl}+t), &&
  \margin > \sqrt{\frac{d-1}{\nlab}} -\sqrt{2 \log \nunl} -1 -t.
\end{aligned}
\end{equation}
\end{enumerate}
\end{Lemma}

Plugging these bounds into the result of Lemma~\ref{lem:optimalclassifier}, we
find high-probability bounds on the $\alpha$-parameter of each of these three
sampling strategies.  As explained in Section~\ref{sec:appendix_thm_proofs},
these bounds hold for any arbitrary joint distribution of the signal component
and the label, $\PP_{y\xsig}$. 
In what follows we derive the bounds on $\alpha$ further for a mixture of
truncated Gaussians. 


\paragraph{Step 3: Bounding $\dstar,\dqstar, \dqhat, \alphaseed$, $\Cseed$ and
the probability that all samples are support vectors for a mixture of truncated
Gaussians.} In order to use Lemma~\ref{lem:alphabounds} to prove the theorem, we
first derive the probability that all samples in $\Dlab$ are support points for
$\PP_{y\xsig} = \ptgmm$  for uniform and oracle/empirical margin-based sampling.
This result is summarized in Lemma~\ref{lem:SPuniform} (see
Appendix~\ref{app:proof_sp} for the proof). 

\begin{Lemma}[All samples are support points]
\label{lem:SPuniform}
Let $\Dlab$ be a dataset of $\nlab < d-1$ samples drawn via either uniform sampling,
margin-based sampling or oracle margin-based sampling from a large unlabeled
dataset.
The unlabeled data is drawn i.i.d.\ from the multivariate mixture of truncated Gaussians
distribution. Then, for a constant $c(\mean_{tr},
\stdsig_{tr}, \nunl)>0$ independent of $d$ and
$\nlab$ it holds with probability larger than
$1-2e^{-(\sqrt{(d-1)/\nlab}-\sqrt{\log \nlab}-c)^2}$ that all samples in
$\Dlab$ are support points of the max-$\ell_2$-margin classifier of
$\Dlab$.
\end{Lemma}

Next, we bound $\dqstar, \dqhat, \alphaseed$ and $\Cseed$ which finalizes the
proof.
We note that for uniform sampling, $\dstar$ is the average of $\nlab$ i.i.d.\
samples from a one-sided truncated Gaussian, which is a sub-Gaussian random
variable with mean $\mean_{tr}$ and standard deviation $\stdsig_{tr}$. Hence,
with probability larger than $1-2e^{-t^2/2}$, it holds via Hoeffding's
inequality that

\begin{equation}
  \label{eq:upperbounddstar}
  \dstar > \mean_{tr} - t \stdsig_{tr}\nlab^{-1/2}.
\end{equation}
By the same argument it holds
with probability greater than $1-2e^{-t^2/2}$ that
\begin{equation}
\label{eq:upperboundcseed}
\Cseed < \mean_{tr} + t \stdsig_{tr}\nseed^{-1/2}.
\end{equation}


To bound the remaining quantities, we treat each of the three sampling strategies separately.

\paragraph{(a) Uniform sampling.} Plugging the bounds on $\dstar$
(Equation~\eqref{eq:upperbounddstar}) and $\avgmargin$
(Lemma~\ref{lem:alphabounds}) into Lemma~\ref{lem:optimalclassifier} and
multiplying the independent probability statements yields the expression of the
upper bound $\alphaPL \geq \alpha(\thetaunif) $ that appears in
Theorems~\ref{thm:oracleuncertainty} and~\ref{thm:uncertaintysampling}.




\paragraph{(b) Oracle margin-based sampling.} We now bound $\dqstar$ using the
following lemma which we prove in Section~\ref{sec:proofdqstar}.

\begin{Lemma}[Bound on $\dqstar$]
\label{lem:dqstar}
Let $\Dunl$ and $\Dseed$ be the unlabeled set and the labeled seed set,
respectively, with covariates drawn i.i.d.\ from the multivariate mixture of truncated Gaussians distribution.
Then, with probability larger than $1-e^{-t^2}$, we have that
\begin{equation*}
  \dqstar < \stdsig \left( \Phi^{-1}\left(\left(t(2
  \nunl)^{-1/2}+(1-\passiveratio)\nlab/\nunl\right)\left(1-\Phi(-\mean/\stdsig)\right)+\Phi(-\mean/\stdsig)\right)
\right)+\mean.
\end{equation*}

Moreover, if $\nunl > \max(10^5, 10^3\nlab)$ and $\mean/\stdsig < 2$ then with probability greater than 0.99
\begin{equation*}
  \label{eq:upperbounddqstar}
\dqstar < 6.059 \cdot 10^{-2} \mean_{tr}.
\end{equation*}
\end{Lemma}

We now argue that the conditions required for Equation~\eqref{eq:upperbounddqstar}
to hold are not too restrictive. Indeed, it is standard in practical active
learning settings that the unlabeled set is orders of magnitude larger than the
labeling budget.
Moreover, in most real-world datasets there exist ambiguous samples, close to
the optimal decision boundary, which can be be difficult to classify even for
human experts. The condition $\mean/\stdsig < 2$ ensures that that is the case
in our setting as well, with high probability.




Invoking the probability bound in Lemma~\ref{lem:SPuniform},
plugging the bounds on $\Cseed$ (Equation~\eqref{eq:upperboundcseed}), $\margin$
(Lemma~\ref{lem:alphabounds}) and $\dqstar$ (Lemma~\ref{lem:dqstar}) into
Lemma~\ref{lem:optimalclassifier} gives the expression for lower bound $\alphaor
\leq \alpha(\thetaor)$ that appears in Theorem~\ref{thm:oracleuncertainty}.
Invoking all the probability statements involved and combining this result with
the previous derivation of $\alphaPL$ finishes the proof of
Theorem~\ref{thm:oracleuncertainty}.

\paragraph{(c) Two-stage margin-based sampling.}
For bounding $\dqhat$ 
we can use a similar technique as in Lemma~\ref{lem:dqstar}, if we assume
further that $\stdsig \geq 1$. This condition ensures that, with high
probability, there exist examples with a high signal component for any $\mean
\ge 0$. The following lemma states the bound
on $\dqhat$ (see Section~\ref{sec:dqhatproof} for the proof).

\begin{Lemma}[Bound on $\dqhat$]
\label{lem:dqhat}
Let $\Dunl$ and $\Dseed$ be the unlabeled set and the labeled seed set,
respectively, with covariates drawn i.i.d.\ from the multivariate mixture of truncated Gaussians distribution.
If $\stdsig \geq 1$ then it holds with probability larger than $1-e^{-t^2}$ that
\begin{equation*}
  \dqhat < \stdsig \left( \Phi^{-1}\left(\left(t(2
  \nunl)^{-1/2}+(1-\passiveratio)\nlab/\nunl\right)\left(1-\Phi(-\mean/\stdsig)\right)+\Phi(-\mean/\stdsig)\right)
\right)+\mean.
\end{equation*}
Moreover, if $\nunl > \max(10^5, 10^3\nlab)$ and $\mean/\stdsig < 2$ then with probability greater than 0.99
\begin{equation*}
\dqhat < 6.059 \cdot 10^{-2} \mean_{tr}.
\end{equation*}
\end{Lemma}

Using Lemma~\ref{lem:dqhat}, we now derive an upper bound on $\alphaseed$. We
note that the seed set is drawn i.i.d.\ from the data distribution. Hence, we
can use the bound for uniform sampling of Lemma~\ref{lem:optimalclassifier} and
set $\Dlab=\Dseed$ to arrive at $\alphaseed < \avgmargin/\dstar$.  By a similar
argument, we use the bound of Equation~\eqref{eq:upperbounddstar} to lower bound
$\Cseed$ and we obtain that $\Cseed > \mean_{tr} - t\stdsig_{tr}(\rho
\nlab)^{-1/2}$.  Then, by Lemma~\ref{lem:alphabounds}, we have that $\avgmargin <
\left(\frac{d-1}{\rho \nlab} + \frac{2
t}{(\rho\nlab)^{1/2}}\right)^{1/2}$ with probability greater than
$1-2e^{-t^2/2}$. Note that if all uniform samples are support points of the
max-$\ell_2$-margin classifier, then all samples in the seed set are as well for
the max-$\ell_2$-margin classifier of the seed set.  Putting everything together, we
find that, with probability greater than $(1-2e^{-t^2/2})^2$, it holds that

\begin{equation}
\label{eq:alphaseedbound}
\alphaseed \leq \left(\frac{d-1}{\rho \nlab} + \frac{2 t}{(\rho
\nlab)^{1/2}}\right)^{1/2} \left(\mean_{tr}-\frac{t\stdsig_{tr}}{(\rho
\nlab)^{1/2}}\right)^{-1}.
\end{equation}



Plugging the bounds on $\Cseed$ (Equation~\eqref{eq:upperboundcseed}), $\margin$
(Lemma~\ref{lem:alphabounds}) , $\alphaseed$
(Equation~\eqref{eq:alphaseedbound}) and $\dqhat$ (Lemma~\ref{lem:dqhat}) into
Lemma~\ref{lem:optimalclassifier} gives the expression for the lower bound
$\alphaUS \leq \alpha(\thetaunc)$ that appears in
Theorem~\ref{thm:uncertaintysampling}.  Invoking all the probability statements
involved and combining this result with the previous derivation of $\alphaPL$
finishes the proof of Theorem~\ref{thm:uncertaintysampling}.

\section{Proofs of main Lemmas}
\label{app:prooflemmas}

In this section, we provide proofs for the lemmas needed to prove the main
theoretical results presented in Section~\ref{sec:appendix_thm}.

\subsection{Proof of Lemma~\ref{lem:optimalclassifier}}
\label{app:proof_optimalclassifier}

Recall that we can consider parameter vector that are normalized such that
\begin{equation*}
\thetahat = [1, \alpha \thetatilde],
\end{equation*}

\noindent for some $\alpha\geq0$
with $\| \thetatilde \|_2 = 1$.
Further, recall that we decompose covariates as $x= [\xsig,\xnoise]$. For
convenience of notation, we define $\bar{a} = \frac{1}{\nlab}\sum_{(x,y) \in
\Dlab} y \langle \thetatilde, \xnoise \rangle$, namely the average margin in the
$d-1$ noise dimensions $\xnoise$ of points in the dataset $\Dlab$.

From the conditions of the lemma we have that all points in $\Dlab$ are support
points. Since, the distance to the decision boundary induced by $\thetahat$ is
the same for all support points, we can write the max-$\ell_2$-margin
$\allmargin$ as the following average:

\begin{equation*}
\begin{aligned}
  \allmargin &= \frac{1}{\|\thetahat\|_2\nlab} \sum_{(x, y) \in \Dlab} \left(y
x_1 + \alpha y \langle \thetatilde, \xnoise \rangle \right)  = \frac{1}{\sqrt{1
+ \alpha^2}} \left( \dstar + \alpha \margavg \right).
\end{aligned}
\end{equation*}

Maximizing over $\alpha$, we find that the max-$\ell_2$-margin classifier
$\thetahat$ is determined by the following $\alpha$-parameter: 

\begin{equation*}
  \alpha = \frac{\margavg}{\frac{1}{\nlab} \sum_{(x,y) \in \Dlab} y x_{1}} =
  \frac{\margavg}{\dstar}.
\end{equation*}

Hence, the max-$\ell_2$-margin classifier can be written as $\thetahat = \left[
1, \frac{\margavg}{\dstar} \thetatilde \right]$ and the margin is given by
$\allmargin = \sqrt{\dstarsq+ \margavg^2}$.

Now we prove that we can use the maximum average-$\ell_2$-margin $\avgmargin$
and the max-$\ell_2$-margin $\margin$ in the $d-1$ noise coordinates to sandwich
$\margavg$ as follows: $\avgmargin \geq \margavg \geq
\sqrt{\margin^2-\dstarsq}$.

The first inequality follows directly from the definition of the maximum
average-$\ell_2$-margin $\avgmargin$ in Equation~\eqref{eq:avgmargin}. 
We now prove the second inequality.  Let $\mmnoise$ be the max-$\ell_2$-margin
classifier in the $d-1$ noise coordinates, with $\|\mmnoise\|_2 = 1$.
By the definition of $\mmnoise$, it holds that $y \langle \mmnoise,
\xnoise \rangle \geq \margin$.  Moreover, consider the classifier determined by
$\thetapart = [\min_{(x, y) \in \Dlab} y \xsig, \margin \mmnoise] \in
\mathbb{R}^{d}$.  Since the max-$\ell_2$-margin $\allmargin$ is maximal it holds
that

\begin{equation*}
  \allmargin \ge
  \min_{(x,y)\in \Dlab} \frac{y \langle \thetapart, x\rangle}{\|\thetapart\|_2}
  \geq \sqrt{\left(\min_{(x,y)\in \Dlab}
  y x_1\right)^2 + \margin^2}.
\end{equation*}

Using that $\allmargin = \sqrt{\dstarsq+ \margavg^2}$, and solving for $\margavg$, we find that

\begin{equation*}
  \alpha  = \frac{\margavg}{\dstar} \geq \frac{\sqrt{\margin^2+\left(\min_{(x,y)\in \Dlab} y
  x_1\right)^2-\dstarsq}}{\dstar} \geq \frac{\margin}{\dstar}-1,
\end{equation*}
which concludes the proof.

\subsection{Proof of Lemma~\ref{lem:alphabounds}}
\label{sec:lemalphabounds}

Lemma~\ref{lem:alphabounds} consists of three statements: a lower bound on the
value of $\alpha$ for margin-based sampling and for oracle margin-based
sampling, and an upper bound on $\alpha$ for passive learning.


Recall that by Lemma~\ref{lem:optimalclassifier} we have that for a labeled set $\Dlab$
collected through any sampling strategy, the $\alpha$-parameter of the
max-$\ell_2$-margin classifier of $\Dlab$ is lower and upper bounded by

\begin{equation*}
\label{eq:alphabounds}
\frac{\margin}{\dstar}-1 \leq \alpha \leq \frac{\avgmargin}{\dstar},
\end{equation*}

\noindent where $\dstar = \frac{1}{\nlab} \sum_{(x, y) \in \Dlab}  y \xsig$.  To
prove Lemma~\ref{lem:alphabounds}, we apply Lemma~\ref{lem:optimalclassifier}
and bound $\margin, \avgmargin$ and $\dstar$ for each sampling strategy
separately.


\subsubsection{Key margin results}

To prove Lemma~\ref{lem:alphabounds} we need to bound the $\ell_2$-margin in the
$d-1$ noise coordinates of a dataset. In this section we give high probability
bounds for the $\ell_2$-margin, which hold if the conditions of
Lemma~\ref{lem:alphabounds} are satisfied, i.e.\ the max-$\ell_2$-margin
classifier exists.



\begin{Lemma}[Bounds for margins $\marginlem, \avgmarginlem$.]
  \label{lem:noisemargins}
  Let $\Dlabnoiselem$ be a dataset of size $\nlablem < d$ with i.i.d.\ inputs
$\xnoiselem \sim \NN(0, \eye_{d})$ and arbitrary labels such that the
max-$\ell_2$-margin solution exists.  Then it holds with probability at least $1-2e^{-t^2 /2}$ that
\begin{itemize}
  \item    the maximum average-$\ell_2$-margin $\avgmarginlem$ of $\Dlabnoiselem$ is upper bounded by
    \begin{equation}
      \label{eq:averagemarg}
\avgmarginlem \leq \sqrt{\frac{d}{\nlablem}+\frac{2 t}{\sqrt{\nlablem}}}.
\end{equation}
\item the max-$\ell_2$-margin $\marginlem$ of $\Dlabnoiselem$ is upper and lower bounded by 
  \begin{equation}
    \label{eq:nonsigmarginUnif}
\sqrt{\frac{d}{\nlablem}}-1-t \leq \marginlem \leq \sqrt{\frac{d}{\nlablem}}+1+t .
\end{equation}
\end{itemize}

Let $\Dlabnoiselem$ be a labeled dataset of size $\nlablem < d$ where
$\{\xnoiselem : (\xnoiselem,y) \in \Dlabnoiselem\}$ is an arbitrary subset among
$\nunllem$ i.i.d. samples $\xnoiselem\sim \NN(0, \eye_{d})$ with arbitrary labels such that 
the max-$\ell_2$-margin solution exists. Then, with probability at least $1-e^{-t^2/2}$,
\begin{itemize}
  \item the max-$\ell_2$-margin $\marginlem$ of $\Dlabnoiselem$ is upper and lower bounded by
    \begin{equation}
      \label{eq:nonsigmarginAL}
\sqrt{\frac{d}{\nlablem}}-\sqrt{2\log \nunllem}-1-t \leq \marginlem \leq
\sqrt{\frac{d}{\nlablem}}+\sqrt{2\log \nunllem}+1+t.
\end{equation}
\end{itemize}
\end{Lemma}


The proof of the lemma can be found in Section~\ref{sec:marginlemmaproofs}.

\subsubsection{Proof of Lemma~\ref{lem:alphabounds}}

We now use Lemma~\ref{lem:noisemargins} to prove Lemma~\ref{lem:alphabounds}.
We do so by replacing $\Dlabnoiselem$ in Lemma~\ref{lem:noisemargins} by the set
$\Dlabnoise := \{(\xnoise, y) : (x, y) \in \Dlab\text{, where } x=[\xsig,
\xnoise]\}$, where $\Dlab$ is collected with one of the three sampling
strategies that we consider. By the conditions of Lemma~\ref{lem:alphabounds} we
have that the max-$\ell_2$-margin solution exists.  

\paragraph{Uniform sampling}

\begin{itemize}[label={},leftmargin=0cm]
  \vspace{-0.3cm}

\item For uniform sampling, Lemma~\ref{lem:noisemargins} directly yields the upper
bound for $\avgmargin$: it suffices to replace $\avgmarginlem$ with $\avgmargin$
to arrive to arrive at $\avgmargin \leq \sqrt{\frac{d-1}{\nlab}+\frac{2
t}{\sqrt{\nlab}}}$, which
proves Equation~\eqref{eq:alphaunif}.

\end{itemize}


\paragraph{Oracle margin-based sampling}

\begin{itemize}[label={},leftmargin=0cm]
  \vspace{-0.3cm}
\item \emph{Bound for $\margin$.} We now argue that the set
  $\Dlabnoise := \{(\xnoise, y) : (x, y) \in \Dlab\text{, where } x=[\xsig,
  \xnoise]\}$ satisfies the assumptions of Lemma~\ref{lem:noisemargins} when
  $\Dlab$ is collected with oracle M-AL. The bound on $\margin$
  then follows directly.  
  Note that oracle M-AL queries the
  $(1-\passiveratio)\nlab$ closest points to the optimal decision boundary.
  Importantly, the Bayes optimal classifier is independent of the $d-1$ noise
  coordinates of the covariates.  Therefore, $\Dlabnoise$ selected with oracle
  M-AL is drawn i.i.d.\ from a standard normal distribution, and
  hence, satisfies the conditions of Lemma~\ref{lem:noisemargins}.

Since, we have that $d > n-1$ and we consider linear classifiers with an
intercept at the origin, the max-$\ell_2$-margin classifier always exists.
Therefore, we are in the conditions of Lemma~\ref{lem:noisemargins} and we get
that
$\margin \geq \sqrt{(d-1)/\nlab}-1-t$ with probability greater than $1-2e^{-t^2/2}$.

\item \emph{Bound for $\dstar$.} Next, observe that using the definitions of $\passiveratio =
\frac{\nseed}{\nlab}$ and $\Cseed=\frac{1}{\nseed} \sum_{(x,y) \in \Dseed} y \xsig$ we can write
\begin{align*}
  \dstar &=\frac{1}{\nlab} \sum_{(x, y) \in \Dlab}  y x_1 = \passiveratio \Cseed + \frac{1}{\nlab} \sum_{(x, y) \in \Dlab \setminus \Dseed}  y x_1\\
  &=\passiveratio \Cseed + (1-\passiveratio) \frac{1}{\nlab-\nseed} \sum_{(x, y)
  \in \Dlab \setminus \Dseed}  y x_1 \leq \passiveratio \Cseed +
  (1-\passiveratio) \dqstar,
\end{align*}

\noindent where the inequality follows from the definition of 
$\dqstar := \max_{(x,y) \in \Dlab \setminus \Dseed}  y x_{1}$. This concludes the
proof of the inequalities in Equation~\eqref{eq:alphaoracle}.

\end{itemize}




\paragraph{Empirical margin-based sampling}

\begin{itemize}[label={},leftmargin=0cm] \vspace{-0.3cm} \item \emph{Bound for
  $\margin$.} We argue now that $\margin$ corresponds to $\marginlem$ in
  Equation~\eqref{eq:nonsigmarginAL}. In particular, since the inequalities in
  Equation~\eqref{eq:nonsigmarginAL} hold for \emph{any} subset of $\Dunl$, they
  also hold for the set $\Dlab$ collected with empirical margin-based sampling.
  Therefore, we find that with probability greater than $1-e^{-t^2/2}$, the
  max-$\ell_2$-margin in the $d-1$ noise coordinates is lower bounded by
  $\sqrt{(d-1)/\nlab} - \sqrt{2 \log \nunl}-1-t$.



\item \emph{Bound for $\dstar$.} 
We stress that this is the only step in the entire proof of
Theorem~\ref{thm:uncertaintysampling} where we use the two-stage margin-based 
sampling procedure (instead of the iterative process described in
Algorithm~\ref{algo:us}).

Similar to oracle margin-based sampling, the key is to derive a bound
for $\frac{1}{\nlab-\nseed} \sum_{(x, y) \in \Dlab \setminus \Dseed}  y x_1 $.
Recall that $\thetaseed= [1, \alphaseed \thetaseedtilde]$ with
$\|\thetaseedtilde\|_2=1$ is the max-$\ell_2$-margin classifier of
$\Dseed$.
Further, due to the two-stage procedure, $\thetaseedtilde$ is independent of
all the samples in the unlabeled dataset. Using this fact together
with the union bound and that the labels are independent of the $d-1$ last
coordinates, we find that $\max_{(x,y) \in \Dunl} \alphaseed
\langle \thetaseedtilde, y \xnoise\rangle < \sqrt{2 \alphaseed \log
  \nunl} + t$ 
Therefore, together with the definition of $\dqhat$, we have that with probability greater than $1-2e^{-t^2/2}$:
\begin{align*}
  \frac{1}{\nlab-\nseed} \sum_{(x, y) \in \Dlab \setminus \Dseed}  y x_1 &= \frac{1}{\nlab-\nseed} \sum_{(x, y) \in \Dlab \setminus \Dseed} y\langle\thetaseed, x\rangle - \alphaseed
\langle \thetaseedtilde, y \xnoise\rangle  \\
  &\leq  \dqhat + \sqrt{2 \alpha_{seed} \log \nunl} + t, 
\end{align*}
from which the bound for $\dstar$ in Equation~\eqref{eq:nonsigmarginAL} follows.

\end{itemize}

\subsection{Proof of Lemma~\ref{lem:SPuniform}}
\label{app:proof_sp}

We now prove this lemma for oracle margin-based sampling. The result for the
other strategies follows a nearly identical argument if we use the respective
bounds on $\margin$ as in Lemma~\ref{lem:noisemargins}.



Recall that, by definition, all support points have the same $\ell_2$-distance
to the decision boundary of the max-$\ell_2$-margin classifier, denoted by
$\allmargin$.
Clearly, the max-$\ell_2$-margin of $\Dlab$ is lower bounded
by the max-$\ell_2$-margin in the last $d-1$ coordinates, i.e.


\begin{equation}
\label{eq:allmarginlower}
\allmargin \geq \margin.
\end{equation}

Now, let $\Dsupport \subset \Dlab$ be the subset containing all support points,
then by Lemma~\ref{lem:optimalclassifier}, the normalized max-$\ell_2$-margin
classifier can be written as follows:

\begin{equation*}
  \thetahat = \frac{1}{\sqrt{\thetahatone^2 + a^2}}\left[\thetahatone,
  a \thetatilde \right],
\end{equation*}

\noindent with $a> \sqrt{\margin^2 - \thetahatone^2}$ where $\|\thetatilde\|_2=1$ and we use the notation $\thetahatone =
\frac{1}{\left|\Dsupport\right|} \sum_{(x,y) \in \Dsupport}  y x_1$. Therefore,
it holds that $\allmargin = y\langle \thetahat, x\rangle \ge
\sqrt{\thetahatone^2 + \margin^2}=\margin$ for all $(x, y) \in \Dsupport$.
After rewriting it follows that $\Dlab \subseteq \Dsupport$, if
the following condition is satisfied:


\begin{equation}
  \label{eq:sufcondsupp}
\max_{(x,y) \in \Dlab} \margin   y \langle \thetatilde , \xnoise \rangle  
 \leq \allmargin^2-y\thetahatone \xsig.
\end{equation}

We now take steps to give a more restrictive sufficient condition that implies
the one in Equation~\eqref{eq:sufcondsupp}, and hence, also implies that
$\Dlab \subseteq \Dsupport$. For an arbitrary pair $(x, y) \in
\Dlab \setminus \Dseed$ the following inequality holds with a probability larger
than $1-2e^{-t^2/2}$:

\begin{equation*}
\begin{aligned}
  \margin^2- y \thetahatone \xsig &\overset{i}{\geq} \margin^{2} - y
  \thetahatone \xsig\\
  &\overset{ii}{\geq} \margin^{2}-y\thetahatone(\mean_{tr}+t\stdsig_{tr})
\end{aligned}
\end{equation*}
\noindent where inequality (i) follows from Equation~\eqref{eq:allmarginlower} and
(ii) holds as $\dqstar<\mean_{tr}$ and the concentration bound in the first
coordinate of uniformly
sampled queries. Now, note that $\thetahatone \leq \mean_{tr}+t \stdsig_{tr} =
c(\mean_{tr}, \stdsig_{tr})$. Hence, for all samples in
$\Dlab\setminus\Dseed$ to be support points it is sufficient that the following
holds:


%

\begin{equation*}
\label{eq:minimumspeq}
\max_{(x,y) \in \Dlab\setminus\Dseed} \margin  y \langle \thetatilde , \xnoise
\rangle  \leq \margin^2-c(\mean_{tr}, \stdsig_{tr})^2.
\end{equation*}

Since $\xnoise$ is distributed according to a multivariate standard Gaussian and
$\|\thetatilde\|_2=1$, we know that 

\begin{equation*}
\label{eq:SPgaussianmax}
 \max_{(x,y) \in \Dlab\setminus\Dseed} y \langle \thetatilde , \xnoise \rangle
 \leq \sqrt{2 \log \nlab} + t,
\end{equation*}

\noindent with probability at least $1-2e^{-t^2/2}$. By combining
Equations~\eqref{eq:SPgaussianmax} and~\eqref{eq:minimumspeq}, we get that
$\Dlab \setminus \Dseed \subseteq \Dsupport$ with probability at least
$1-2e^{-0.5\left(\margin-\tilde{c}(\mean_{tr}, \stdsig_{tr})^2 \margin^{-1} -
\sqrt{2 \log \nlab}\right)^2}$ for a constant $\tilde{c}$ independent of $n$ and
$d$.

Moreover, by Lemma~\ref{lem:noisemargins} we have that $\margin >
\sqrt{\frac{d}{\nlab}}-1-t$. Hence for a constant $0<\tilde{c}(\mean_{tr},
\stdsig_{tr})$ independent of $n$ and
$d$, we have that
$\Dlab\setminus\Dseed \subset \Dsupport$, with a probability greater than
$1-2e^{-0.5\left(\sqrt{d/\nlab} - \sqrt{2 \log \nlab} -\tilde{c}\right)^2}$.

\subsection{Proof of Lemma~\ref{lem:noisemargins}}
\label{sec:marginlemmaproofs}

We now prove Lemma~\ref{lem:noisemargins} which bounds the maximum average- and
min-$\ell_2$-margin of a dataset $\Dlabnoiselem$. The data is either drawn
i.i.d.\ from a multivariate standard normal, or consists of an arbitrary subset
of a larger dataset drawn i.i.d.\ from a multivariate standard normal.

\subsubsection{Bound for the maximum average-$\ell_2$-margin for i.i.d.\
standard normal data}
The maximum average-$\ell_2$-margin can be written as

\begin{equation*}
\begin{aligned}
\avgmarginlem &= \max_{\theta \in \mathcal{S}^{d-1}}
\frac{1}{\nlablem}\sum_{(x, y) \in \Dlabnoiselem} y \langle \theta, x \rangle \\
& = \max_{\theta \in \mathcal{S}^{d-1}}  \frac{\theta_1}{\nlablem} \sum_{(x, y)
\in \Dlabnoiselem} y x_1 + ... + \frac{\theta_d }{\nlablem}\sum_{(x, y) \in
\Dlabnoiselem} y x_d \\
&= \frac{1}{\sqrt{\nlablem}}\max_{\theta \in \mathcal{S}^{d-1}}  \langle \theta,
z\rangle,
\end{aligned}
\end{equation*}

\noindent where in the last equation $z$ is a $d$-dimensional vector distributed
according to a standard Gaussian (note that we consider the samples in
$\Dlabnoiselem$ to be random
variables). By Cauchy-Schwarz, the maximum is found by setting $\theta =
z/\| z \|_2$. Using Chernoff's bound, we find that $\|z\|_2 <
\sqrt{d(1+t)}$ with probability larger than $1-2e^{-d t^2/8}$.
Multiplying by $1/\sqrt{\nlablem}$ yields the result in
Equation~\eqref{eq:averagemarg}.

\subsubsection{Bound for the maximum min-$\ell_2$-margin}


The proof of Equations~\eqref{eq:nonsigmarginUnif} and~\eqref{eq:nonsigmarginAL}
consists of two parts. We first upper and lower bound the max-$\ell_2$-margin
$\marginlem$ by the scaled maximum and minimum singular values of the matrix
$\xnoisematrixlem \in \RR^{d\times \nlablem}$ whose columns are given by $y x$
for $(x, y) \in \Dlabnoiselem$. Then, we use matrix concentration results to
bound these singular values for the two different cases that correspond to
Equations~\eqref{eq:nonsigmarginUnif} and~\eqref{eq:nonsigmarginAL}.


\paragraph{Step 1: Bounding the max-$\ell_2$-margin.}
We use the upper and lower bounds on the extremal singular values of the
data matrix to derive upper and lower bounds on the max-$\ell_2$-margin of the
dataset.

The existence of a max-$\ell_2$-margin solution implies that there exist vectors
$\theta \in \mathcal{S}^{d-1}$ and $v \in \RR^\nlablem$ such that $\xnoisematrixlem^\top \theta = v$ with
$v\geq c\mathbb{1}_{\nlablem}$ element-wise for a $c>0$. For any $\theta \in
\mathcal{S}^{d-1}$, we know that $
\|\xnoisematrixlem^\top \theta\|\leq s_{\text{max}}(\xnoisematrixlem)$.
The (minimum) margin is equivalent to $\marginlem = \max_{\theta} c$ such that
$v>c\mathbb{1}_n$ and hence $\|v\|_2 \geq
\sqrt{\nlablem} \marginlem$. Since $\|v\|_2 = \|\xnoisematrixlem^\top
\theta\|\leq  s_{\text{max}}(\xnoisematrixlem)$, we readily have $\marginlem
\leq \frac{s_{\text{max}}(\xnoisematrixlem)}{\sqrt{\nlablem}}$. We now prove the
lower bound.
Note that as $Z \in \mathbb{R}^{d\times n}$ is a random matrix of standard normal
random variables with $n<d$, it has almost surely a rank of $n$. Hence, there
exists a $\theta' \in \mathbb{R}^{d}$  such that $Z^{\top} \theta' = \mathbb{1}_{n} c
>0$. Moreover, the smallest non-zero singular value of $Z^{\top}$ equals the
smallest singular value of its transpose $Z$. Therefore, using the fact
that any vector $\theta'$ in the span of eigenvectors corresponding to
non-zero singular values satisfies
$\|Z\theta\|_2 \geq s_{\text{min}}(Z)$ and the existence of a solution $\theta'$
for $Z^{\top} \theta' = c \mathbb{1}_n > 0$, we have that there exists a $\theta'$ with
$\|\theta'\|_2=1$ in the span of the eigenvectors corresponding to non-zero
singular values of $Z^{\top}$ such that 
\begin{equation*}
  \margin \geq \min_{j \in [n]} |Z^{\top}\theta|_{j} \geq
  s_{\text{min}}(Z)\frac{1}{\sqrt{n}}.
\end{equation*}

\paragraph{Step 2: Bounding the singular values of $\xnoisematrixlem$.} It
remains to bound the maximum and minimum singular values of $\xnoisematrixlem$
for the two different scenarios considered in Lemma~\ref{lem:noisemargins}.

\begin{itemize}[label={},leftmargin=0cm]
  \vspace{-0.3cm}

\item \emph{(i) For i.i.d.\ samples:} To prove the i.i.d.\ case in
  Equation~\eqref{eq:nonsigmarginUnif}, we use the following set of inequalities
  on the maximal and minimal singular values
  of any random matrix $M \in \mathbb{R}^{d \times \nlablem}$ with $d>\nlablem$ and
  i.i.d.\ standard normal entries:

\end{itemize}
  \vspace{-0.3cm}

\begin{equation}
\label{eq:verschynin}
\sqrt{d}-\sqrt{\nlablem} - t \leq s_{\text{min}}(M) \leq  s_{\text{max}}(M)\leq
\sqrt{d}+\sqrt{\nlablem} + t.
\end{equation}

These inequalities hold with probability greater than $1-2e^{-t^2/2}$ (see e.g.\
Corollary~5.35 in \citet{vershynin10}).  Recall that the columns of the matrix
$\xnoisematrixlem$ are given by $y \xnoiselem$ for $\xnoiselem \in
\Normal(0,\eye_{d})$ and arbitrary $y$. Therefore, $\xnoisematrixlem$ is a
matrix with standard normal entries, which concludes the proof of
Equation~\eqref{eq:nonsigmarginUnif}.

\begin{itemize}[label={},leftmargin=0cm]
  \vspace{-0.3cm}

\item \emph{(ii) For an arbitrary subset of $\nlablem$ samples from $\Dunl$:}
  Let $M_u \in \R^{d\times \nunl}$ be a random matrix with i.i.d.\ standard
  Gaussian entries with $\nunl > \nlablem$.  We define the set of data matrices
  corresponding to all possible subsets of columns of $M_u$ of size $\nlablem$:

\end{itemize}
  \vspace{-0.3cm}

\begin{equation*}
\allsubsets := \{ M \in \RR^{d \times \nlablem} :
\text{ columns of } M \text{ are a subset of the columns of } M_u\}.
\end{equation*}

In order to account for arbitrary subsets, we use the
union bound over the cardinality of the set $|\allsubsets|=\numsubsets =
\frac{\nunl!}{(\nunl-\nlablem)!\nlablem!}  \leq \nunl^{\nlablem}$. Hence, we
obtain:

\begin{equation*}
\begin{aligned}
P\left[\max_{M \in \allsubsets}s_{\text{max}}(M) > (\sqrt{2 \log
\nunl}+1)\sqrt{\nlablem} + \sqrt{d} + t\right]&\\
\leq \numsubsets P\big{[}s_{\text{max}}(M) & >  \left(\sqrt{2 \log
\nunl}+1\right)\sqrt{\nlablem} + \sqrt{d} + t\big{]}.
\end{aligned}
\end{equation*}

Using Equation~\eqref{eq:verschynin} then yields

\begin{equation*}
\begin{aligned}
  \numsubsets P\left[s_{\max}(M) >  (\sqrt{2 \log \nunl}+1)\sqrt{\nlablem} +
  \sqrt{d} + t\right] &\leq e^{\log 2 \numsubsets}e^{-(\sqrt{ 2 \nlablem \log \nunl} + t)^2/2}\\
&= e^{\log(\numsubsets) + \log(2) - \nlablem \log (\nunl) -\sqrt{2\log
\nunl}\sqrt{\nlablem} - t^2/2} \\
&\leq e^{-t^2/2}.
\end{aligned}
\end{equation*}

This proves the upper bound on the maximum singular value of an arbitrary subset
of $\nlablem$ columns of $M_u$, where $M_u \in \RR^{d \times \nunl}$ with
standard normal entries.
Plugging this result into $\marginlem \le
\frac{s_{\text{max}}(\xnoisematrixlem)}{\sqrt{\nlablem}}$ allows us to bound the
max-$\ell_2$-margin of $\Dlabnoiselem$.
Observe that by symmetry of the random variable, the same derivation holds for
the minimal singular value as well.
This concludes the proof of the inequalities in
Equation~\eqref{eq:nonsigmarginAL}.


\section{Technical Proofs}
\subsection{Proof of Lemma~\ref{lem:dqstar}}
\label{sec:proofdqstar}
Define $\nquery := (1-\passiveratio)\nlab$ to be the number of queries made with
margin-based sampling. Recall that $\dqstar$ is defined as the distance to the decision boundary
determined by $\thetastar$ of the $\nquery^{th}$ closest sample from the
unlabeled dataset. Note that the unlabeled dataset is drawn i.i.d.\ from the
mixture of truncated Gaussians distribution described in
Section~\ref{sec:datadistr}. Let $x_q$ be the $\nquery^{th}$ closest sample to
the decision boundary determined by $\thetastar$. Let $\Phi_{tr}$ denote the
cumulative distribution function of a Gaussian with mean $\mean$ and standard
deviation $\stdsig$ truncated to the interval $(0, \infty)$:

\begin{equation*}
\label{eq:cdftrgaus}
\Phi_{tr}(t) = \frac{\Phi \left(\frac{t-\mean}{\stdsig}\right)-\Phi\left(-
\frac{\mean}{\stdsig}\right)}{1 - \Phi \left( - \frac{\mean}{\stdsig}\right)}.
\end{equation*}

We find that for some $t>0$ the following holds: 

\begin{equation*}
\label{eq:startalsofordqhat}
P[\dqstar < t] =
1-\sum_{i=1}^{\nquery-1}\binom{\nunl}{i}\Phi_{tr}(t)^{i}(1-\Phi_{tr}(t)^{\nunl-i}).
\end{equation*}

Using Hoeffding's inequality, we can bound the probability as

\begin{equation*}
P[\dqstar< t] \geq 1-e^{-2 \nunl \left(\Phi_{tr}(t)-\nquery/\nunl\right)^2}.
\end{equation*}

After the change of variable $\tilde{t} = \Phi_{tr}^{-1}\left(\frac{t}{\sqrt{2
\nunl}}+\frac{\nquery}{\nunl}\right)$ we arrive at:

\begin{equation*}
P[\dqstar < \Phi_{tr}^{-1}(\tilde{t}/\sqrt{2 \nunl}+\nquery/\nunl)] \geq
1-e^{-\tilde{t}^2}.
\end{equation*}

Now plugging in the definition of the inverse of the CDF of the positive-sided
truncated Gaussian yields that, with probability of at least $1-e^{-t^2}$, the
following holds:

\begin{equation}
\label{eq:dqstar}
\dqstar < \stdsig \left( \Phi^{-1}\left(\left(\frac{t}{\sqrt{2 \nunl}}
+\frac{\nquery}{\nunl}\right)\left(1-\Phi
\left(-\frac{\mean}{\stdsig}\right)\right)+\Phi
\left(-\frac{\mean}{\stdsig}\right)\right) \right)+\mean.
\end{equation}

Observe that the right-hand side of Equation~\eqref{eq:dqstar} is monotonically
increasing in $\frac{1}{\nunl}$ and $\frac{\nquery}{\nunl}$.  From the
assumptions required for Lemma~\ref{lem:dqstar} we have that
$\frac{\nquery}{\nunl} < \frac{\nlab}{\nunl} < 10^{-3}$ and $\frac{1}{\nunl} <
10^{-5}$. Fixing $t$ such that the probability is $0.99$, we can further write
the upper bound as follows:

\begin{equation*}
\dqstar < \stdsig \left( \Phi^{-1}\left(c\left(1-\Phi
\left(-\frac{\mean}{\stdsig}\right)\right)+\Phi
\left(-\frac{\mean}{\stdsig}\right)\right) \right)+\mean,
\end{equation*}

\noindent where $c$ is a small positive constant that depends on $t$ and which
can be computed numerically for fixed $t$. For convenience, we define $\beta :=
\frac{\mean}{\stdsig}$. Then, using the formula for $\mean_{tr}$ from
Equation~\eqref{eq:meantruncated} we arrive at:

\begin{equation*}
\frac{\dqstar}{\mean_{tr}} < \left(1-\Phi \left( -\beta\right)
\right)\frac{\left( \Phi^{-1}\left(c\left(1-\Phi \left(-\beta\right)\right)+\Phi
\left(-\beta\right)\right) \right)+\beta}{\beta \left(1- \Phi \left(-\beta
\right) \right) + \phi \left(-\beta\right)} =
\frac{\beta+\Phi^{-1}\left(c\left(1-\Phi \left(-\beta\right)\right)+\Phi
\left(-\beta\right)\right)}{\beta  + \frac{\phi
\left(-\beta\right)}{1-\Phi(-\beta)}}.
\end{equation*}

Taking the derivative and an algebraic exercise shows that the right-hand side is an increasing function of $\beta$. Hence, we can plug in the numerical value of $c$ and the condition $\beta = \mean/\stdsig < 2$ to find the desired upper bound and conclude the proof.

\subsection{Proof of Lemma~\ref{lem:dqhat}}
\label{sec:dqhatproof}

Define $\nquery := (1-\passiveratio)\nlab$ to be the number of queries made with
margin-based sampling. Recall that $\dqhat$ is defined as the distance to the
decision boundary determined by $\thetaseed$ of the $\nquery^{th}$ closest
sample from the unlabeled dataset. Note that the unlabeled dataset is drawn
i.i.d.\ from the mixture of truncated Gaussians distribution described in
Section~\ref{sec:datadistr}. Let $x_q$ be the $\nquery^{th}$ closest sample to
the decision boundary determined by $\thetaseed$ and define $\p = P[|\langle
\thetaseed, x\rangle |<t]$ for a sample $x$ drawn from the multivariate mixture
of truncated Gaussians distribution and a constant $t>0$. Clearly

\begin{equation*}
P[\dqhat < t] = 1-\sum_{i=1}^{\nquery-1}\binom{\nunl}{i}\p^{i}(1-\p)^{\nunl-i}.
\end{equation*}
Using Hoeffding's inequality, we can bound the probability as
\begin{equation*}
P[\dqhat< t] \geq 1-e^{-2 \nunl (\p-\nquery/\nunl)^2}.
\end{equation*}

Now using the definition of the mixture of truncated Gaussians distribution and
recalling that $\thetaseed = [1, \alphaseed \thetaseedtilde]$, we find that

\begin{equation*}
  \p = P\left[\left|x_1 + \alphaseed\langle \thetaseedtilde, \xnoise \rangle
  \right| < t \sqrt{1 + \alphaseed^{2}}\right].
\end{equation*}

We note that $\langle \thetaseedtilde, \xnoise \rangle$ is distributed according to
a standard normal and $x_1$ according to a mixture of univariate Gaussians
truncated at $0$ with mean $y \mean$ and variance $\stdsig^2$. Denote by
$\Phi_{tr}$ the cumulative distribution function of the truncated Gaussian
distribution. Then $x_1 < t$ with probability $\Phi_{tr}(t)$. If
$\stdsig > 1$, then $P[x_1 < t] < P[\langle \thetaseedtilde, \xnoise \rangle < t]$
for all $t>0$. In that case, we find that 

\begin{equation*}
\p \leq  P\left[|x_1| < t \right].
\end{equation*}

Hence, we can take $p = P\left[|x_1| < t \right]$ as an upper bound and use the
derivation in the proof of Lemma~\ref{lem:dqstar} from Equation~\eqref{eq:startalsofordqhat} onwards.


\subsection{Proof of Corollary~\ref{cor:oracleuncertainty}}
\label{sec:cororacleproof}

In Theorem~\ref{thm:oracleuncertainty}, let us denote the numerators of the
expressions of $\alphaor$ and $\alphaPL$ by $\marginminor$ and $\marginminPL$,
respectively. Similarly, we use the notation $\Mor, \MPL$ for the denominators
of $\alphaor$ and $\alphaPL$, respectively.  Since the function $\Psifull$
defined in Equation~\eqref{eq:testerror} is monotonic in $\alpha$, it follows from
Theorem~\ref{thm:oracleuncertainty} with high probability that oracle
M-AL performs worse than passive learning if $\alphaor =
\frac{\marginminor-\Mor}{\Mor} > \frac{\marginminPL}{\MPL} = \alphaPL$. We observe
that

\begin{equation*}
\frac{\marginminor-\Mor}{\Mor}> \frac{\marginminPL}{\MPL} \iff
\frac{\marginminor-\Mor}{\marginminPL} > \frac{\Mor}{\MPL}.
\end{equation*}

For an $\eta \in (0, 1)$ and using the expressions for $\marginminor$ and
$\marginminPL$ from Theorem~\ref{thm:oracleuncertainty}, we have that

\begin{equation*}
\frac{\marginminor}{\marginminPL} > \eta \iff d-1 > \nlab \frac{\left(1+\Mor+t+\eta\sqrt{2
t \nlab^{-1/2}}\right)^2}{(1-\eta)^2}.
\end{equation*}

Moreover, for $\eta > \frac{\Mor}{\MPL}$, we find that
\begin{equation*}
\eta > \frac{\Mor}{\MPL} \iff \mean_{tr} > \frac{\passiveratio(\mean_{tr}+t
  \stdsig_{tr}\nseed^{-1/2}) +  6.059(1-\passiveratio)\cdot 10^{-2} \mean_{tr} +
t \stdsig_{tr} \nlab^{-1/2} \eta }{\eta}.
\end{equation*}

Hence, by plugging in $\eta = 0.5$, we have that $\frac{\marginminor}{\Mor}> \frac{\marginminPL}{\MPL}$ if 
\begin{align*}
  \frac{d-1}{\nlab} &> 4 \left(1+t+\Mor+\sqrt{t (4\nlab)^{-1/2}}\right)^2
  \label{eq:cond_d} \text{ and}\\
  \mean_{tr} &> 2 \passiveratio(\mean_{tr}+t\stdsig_{tr}\nseed^{-1/2}) +
  2\cdot 6.059(1-\passiveratio)\cdot 10^{-2} \mean_{tr} + 2t \stdsig_{tr}
  \nlab^{-1/2}.
\end{align*}

Using $\passiveratio < 0.5$ we arrive at 
\begin{equation*}
\mean_{tr}> \frac{ t \stdsig_{tr} (1 + 2\passiveratio^{1/2})\nlab^{-1/2}}{1
-2\passiveratio}.
\end{equation*}

We now solve for $\passiveratio$ and find

\begin{equation*}
\sqrt{\passiveratio} < q\left(\sqrt{1+\frac{1}{2 q^2}-\frac{1}{q}}-1\right),
\end{equation*}

\noindent with $q = \frac{t \stdsig_{tr}}{2 \mean_{tr} \sqrt{\nlab}}$.
Using the fact that $\sqrt{\passiveratio} < \frac{1}{\sqrt{2}}$ and ignoring
negligible terms, we get the following bound on $\passiveratio$

\begin{equation*}
\passiveratio < \frac{1}{2}-\frac{1+\sqrt{2}}{2} \frac{t
\stdsig_{tr}}{\sqrt{\nlab} \mean_{tr}},
\end{equation*}
which concludes the proof.

\subsection{Proof of Corollary~\ref{cor:cortwostageproof}}
\label{sec:cortwostageproof}

Similar to Section~\ref{sec:cororacleproof}, let us denote in
Theorem~\ref{thm:uncertaintysampling} the numerators of the expressions of
$\alphaUS$ and $\alphaPL$ by $\marginminAL$ and $\marginminPL$, respectively.
Similarly, we use the notation $\MUS, \MPL$ for the denominators of $\alphaUS$
and $\alphaPL$, respectively.

By Theorem~\ref{thm:uncertaintysampling} margin-based sampling leads to a
classifier with a lower test error than uniform sampling, if $\alphaUS =
\frac{\marginminAL-\MUS}{\MUS} \geq \frac{\marginminPL}{\MPL} = \alphaPL$. Similar to
the proof of Corollary~\ref{cor:oracleuncertainty}, we find that

\begin{equation*}
\frac{\marginminAL-\MUS}{\MUS} \geq \frac{\marginminPL}{\MPL} \iff
\frac{\marginminAL-\MUS}{\marginminPL} \geq \frac{\MUS}{\MPL}.
\end{equation*}

Let $\eta \in (0, 1)$. Then it holds that

\begin{equation*}
\begin{aligned}
  \frac{\marginminAL-\MUS}{\marginminPL} \geq \eta &\iff \sqrt{\frac{d-1}{\nlab}}-\sqrt{2
  \log \nunl} -1 -\MUS -t \geq \eta \left( \sqrt{\frac{d-1}{\nlab}+2 t\nlab^{-1/2}} \right)\\
  &\Rightarrow \sqrt{\frac{d-1}{\nlab}}-\sqrt{2 \log \nunl} -1 -\MUS -t \geq \eta \left(
  \sqrt{\frac{d-1}{\nlab}}+\sqrt{\frac{2 t}{\sqrt{\nlab}}} \right)\\
  &\iff \frac{d-1}{\nlab} > \frac{\sqrt{2 \log \nunl} + 1+\MUS+ t +\eta\sqrt{2
t\nlab^{-1/2}}}{(1-\eta)^2}.
\end{aligned}
\end{equation*}

Choosing $\eta = 0.5$ yields

\begin{equation*}
  \frac{d-1}{\nlab} > 4\left(\sqrt{2 \log \nunl} + 1+ \MUS +t +\sqrt{ t (4\nlab)^{-1/2}}\right).
\end{equation*}

Similarly, we have that $\frac{\MUS}{\MPL}< \eta$. Plugging in the expressions given in Theorem~\ref{thm:uncertaintysampling}, we find that

\begin{equation*}
\label{eq:cseed}
\passiveratio \Cseed + (1-\passiveratio) \left(6.059 \cdot 10^{-2}\mean_{tr} + \left(
    \frac{2 \log \nunl}{\Cseed}\right)^{1/2}\left(\frac{d-1}{\passiveratio \nlab}+\frac{2 \stdsig_{tr}t}{\passiveratio \nlab}\right)^{1/4}+t
  \right)< \eta \left( \mean_{tr} - \frac{t\stdsig_{tr}}{\sqrt{\nlab}} \right).
\end{equation*}

Recalling the bound on $\Cseed$ in Equation~\eqref{eq:upperboundcseed}, we find that if

\begin{equation*}
\label{eq:meanbound1}
\mean_{tr}  \geq \left(\frac{d-1}{\passiveratio  \nlab}+ \frac{2 \stdsig_{tr}
t}{\rho \nlab} \right)^{1/6}(2 \log \nunl)^{1/3} +
\frac{t\stdsig_{tr}}{(\passiveratio  \nlab)^{1/2}},
\end{equation*}

\noindent then the following condition on $\passiveratio$ suffices in order to
guarantee that $\alphaPL < \alphaUS$:

\begin{equation*}
  \passiveratio <   \frac{0.878 \mean_{tr}-t \stdsig_{tr}(\passiveratio
\nlab)^{-1/2} - \left(\frac{d-1}{\passiveratio  \nlab}+ \frac{\stdsig_{tr} t}{
\passiveratio  \nlab} \right)^{1/6}(2 \log \nunl)^{1/3} -t}{\Cseed}.
\end{equation*}

\begin{figure}[t]
  \centering
  \begin{subfigure}{0.33\textwidth}
    \centering
    \includegraphics[width=\textwidth]{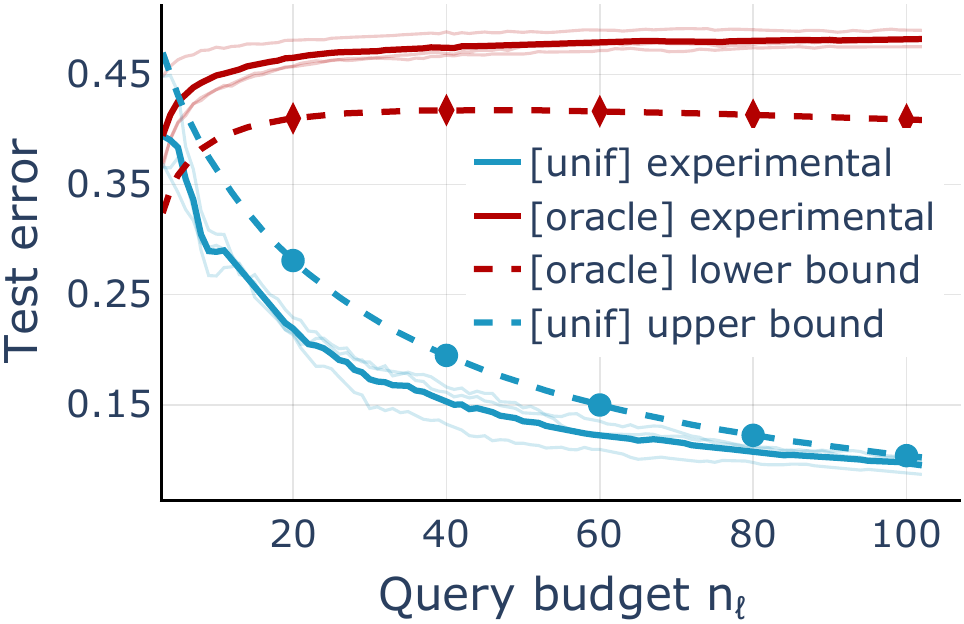}
    \caption{Test error gap vs query budget}
    \label{fig:oracleUSynthexp}
  \end{subfigure}
  \begin{subfigure}{0.3\textwidth}
    \centering
    \includegraphics[width=\textwidth]{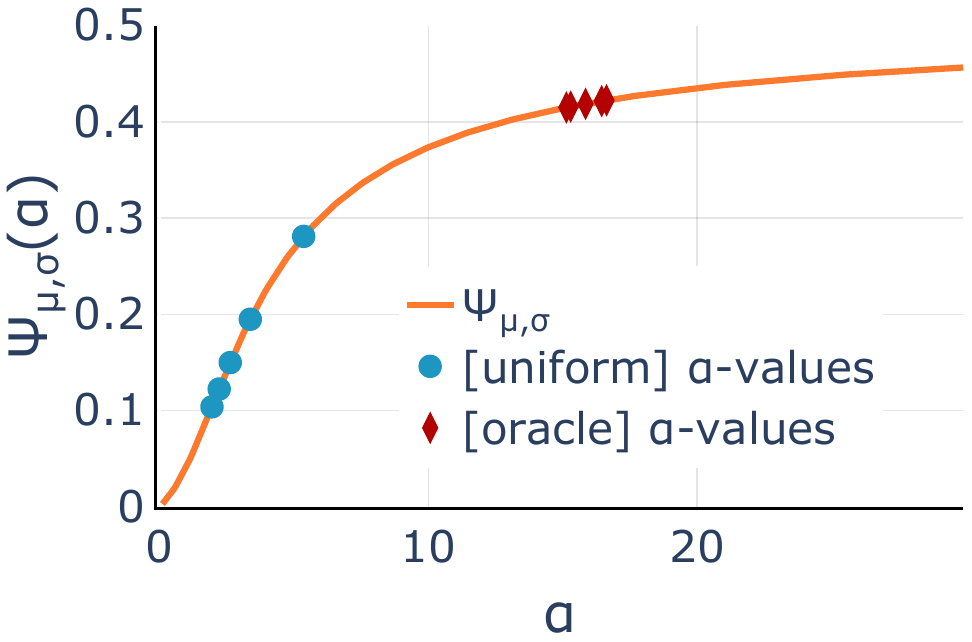}
    \caption{ $\Psifull(\alpha)$ }
    \label{fig:psialpha}
  \end{subfigure}
  \begin{subfigure}{0.3\textwidth}
    \centering
    \includegraphics[width=\textwidth]{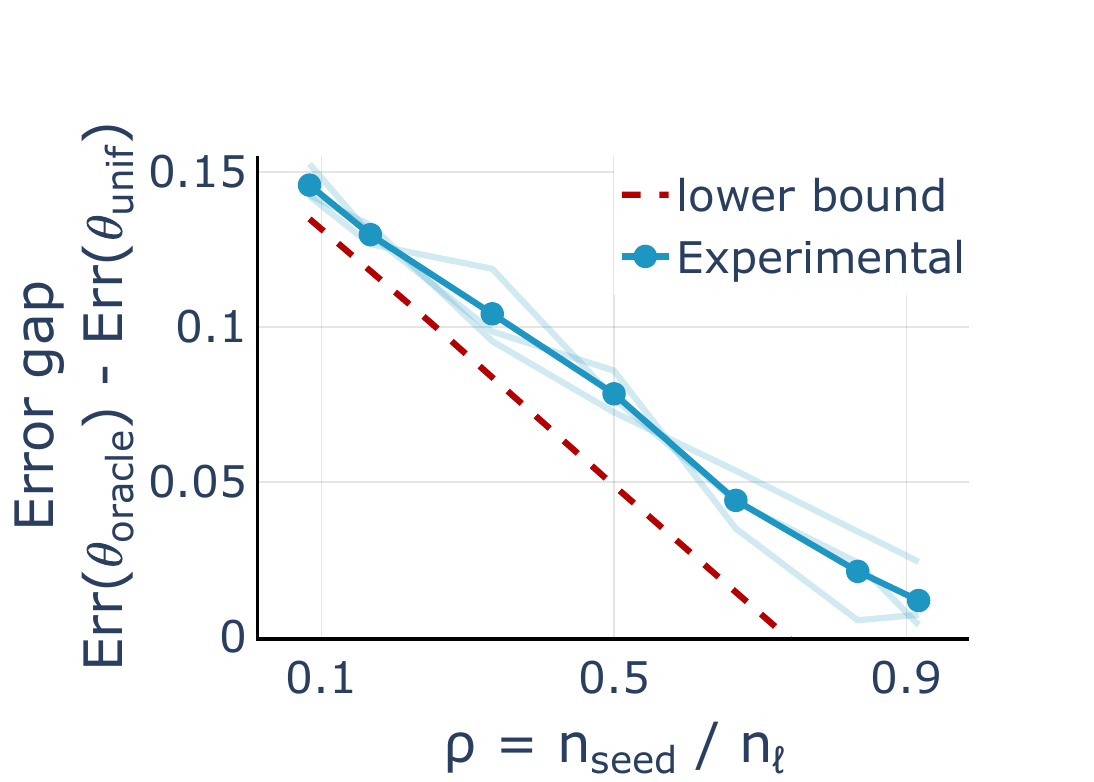}
    \caption{Test error gap vs $\passiveratio$}
    \label{fig:seedsetapp}
  \end{subfigure}
  \caption{\small{For large $d/\nlab$ the bounds of Theorem~\ref{thm:oracleuncertainty}
      are close to tight. (a)~The bounds in Theorem~\ref{thm:oracleuncertainty} (dashed lines) show
      that M-AL leads to lower test error compared to uniform
      sampling.  The lighter color lines correspond to one of $3$ runs with different
      draws of the seed set, while the solid line indicates their mean.
      (b)~The function
      $\Psifull(\alpha)$ is monotonically increasing in $\alpha$. The markers show
      the $(\alpha, \Psifull(\alpha))$ values corresponding to the query budgets
      indicated in figure~(a).
(c)~By the bound in Theorem~\ref{thm:mixtureUS} (dashed line),
      the error gap between M-AL and PL decreases as the proportion of seed
      samples grows.}} 
      \label{fig:syntheticapp}
\end{figure}

\section{Synthetic experiments on the mixture of truncated Gaussians
distribution}
\label{app:synthetic_experiments}

In this section, we give the experimental details to the synthetic experiments
in Figures~\ref{fig:oracleUStheory} and~\ref{fig:seedset}. Further, we
show empirically that for large $d/\nlab$ the theoretical bounds closely predict
the experimental values. Lastly, we further empirically discuss the dependency
on the distributional parameters $\stdsig$ and $\mean$ of the truncated Gaussian
mixture model for \textbf{empirical} margin-based sampling. 

\subsection{Experimental details to Figures~\ref{fig:oracleUStheory} and
\ref{fig:seedset}}

In both Figures~\ref{fig:oracleUStheory} and
\ref{fig:seedset}, we plot the theoretical upper and lower bounds of Theorem
\ref{thm:oracleuncertainty} with $\nunl = 10^5$, $t=3$ and compute $\Psifull$ by integrating using
scikit-learn's function "dblquad". 

In Figure~\ref{fig:oracleUStheory}, we
set $d=1000$, $\mean = 2$, $\stdsig = 2$, $\nseed = 10$ and vary $\nlab$ from
$\nseed$ to $1000$. On the other hand, in Figure~\ref{fig:seedset} we set
$d=1000$, $\stdsig = 2$ and vary the mean-parameter $\mean$ in $\{1,2,3\}$ and the
seed set size $\nseed$ in  $[1, ..., \nlab]$.

\subsection{Verifying the bounds in Theorem~\ref{thm:oracleuncertainty} on
synthetic data}

We now experimentally confirm the bounds in Theorem~\ref{thm:oracleuncertainty}.
Recall that for large $d/\nlab$, the bounds on $\margin$ of
Lemma~\ref{lem:noisemargins} are tight. Therefore we consider two
settings where $d/\nlab$ is large. 

First, in Figure~\ref{fig:oracleUStheory}, we set $d=3k$, $\nunl = 10^5$, $\stdsig =2$
and $\mean = 3$. Then we vary $\nlab$ from $\nseed$ to $100$. We plot the
results of $3$
independent experiments for each setting along with the theoretical lower and
upper bounds given in Theorem~\ref{thm:oracleuncertainty}. Observe that the
theoretical bounds
closely predict both the test error of passive learning as well as the test
error of oracle M-AL.

Further, for completeness, in Figure~\ref{fig:psialpha} we also plot the
function $\Psifull$ with corresponding
$\alpha$-values from the setting in Figure~\ref{fig:oracleUStheory}. Observe
that for small $\alpha$ the function $\Psifull$ increases fast.

Lastly, in Figure~\ref{fig:seedsetapp}, we set $d=10k$,$\nunl = 10^5$, $\mean = 0$, $\stdsig=
3$, $\nlab = 60$ and vary $\nseed$ from $5$ to $55$. Observe that the
theoretical bound closely predicts the test error gap. Moreover, observe that the test error gap
monotonically decreases in $\passiveratio$ both experimentally and according to
the theoretical bound. 

\vspace{-0.2cm}
\paragraph{Logistic regression implementation.}
In all synthetic experiments, we use the SGDClassifier of the Scikit-learn
library \citep{scikit-learn} with the following settings: we set the learning
rate to be a constant of $10^{-4}$ and train for at least $10^4$ epochs without
regularization. Moreover, we set the tolerance parameter to $10^{-5}$ and the
maximum number of epochs to $10^6$. In all experiments, we consider regular
margin-based sampling as defined in Algorithm~\ref{algo:us}.

\begin{figure*}[t]
\centering
  \begin{subfigure}{0.27\textwidth}
    \centering
   \includegraphics[width=\textwidth]{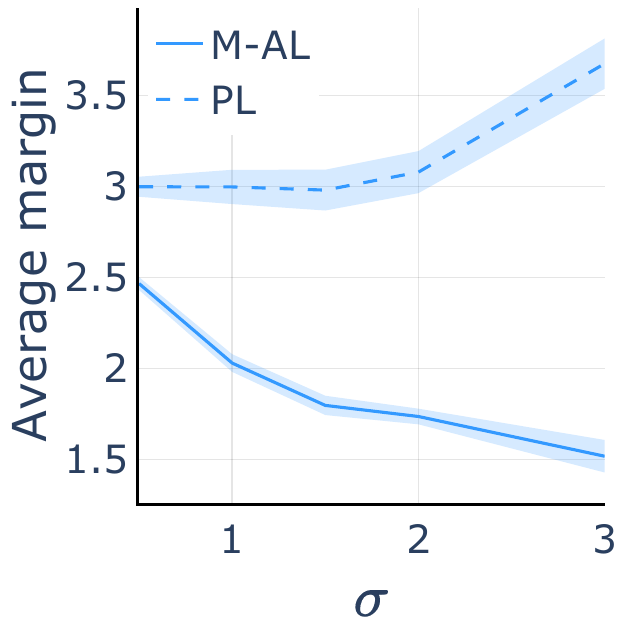}
    \caption{Average margin in $\stdsig$}
    \label{fig:synth_margin}
  \end{subfigure}
  \hspace{1cm}
  \begin{subfigure}{0.53\textwidth}
    \centering
    \includegraphics[width=\textwidth]{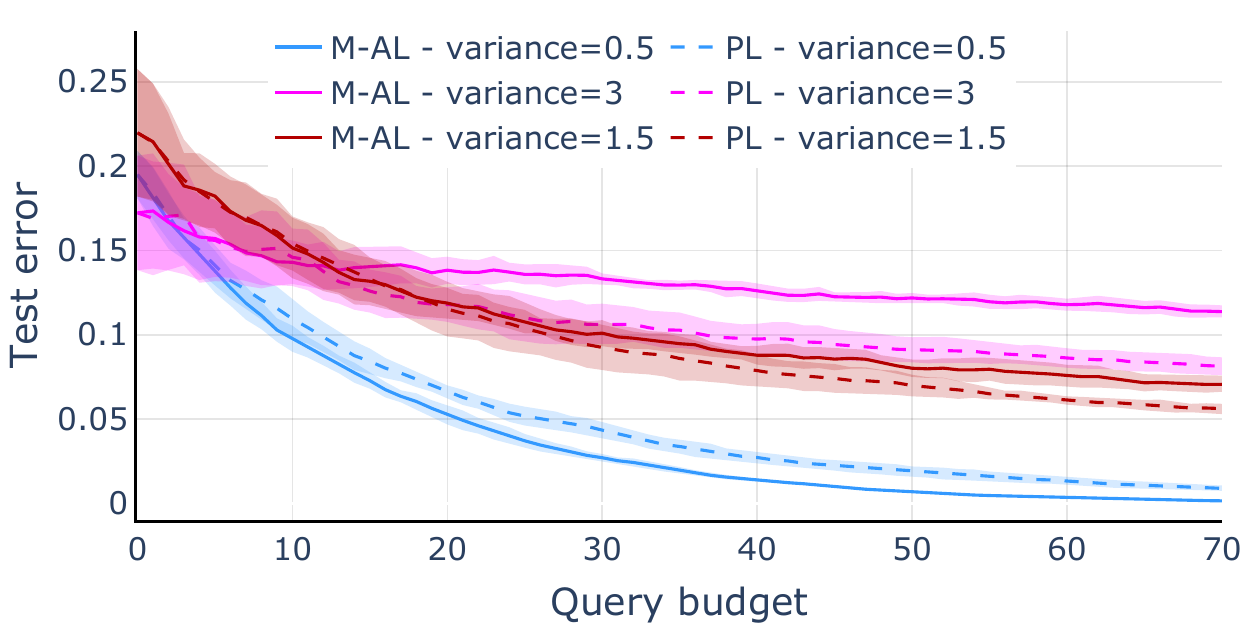}
    \caption{Test error for varying $\stdsig$}
    \label{fig:synth_test_error}
  \end{subfigure}
  \caption{\small{ (a) M-AL collects labeled sets with
       a smaller average margin in the signal component as $\stdsig$ increases.
       (b) As the average margin of a set acquired via M-AL is
       smaller for increasing $\stdsig$, the test error deteriorates, as
       predicted by Lemma~\ref{lem:alphabounds}. The test error of M-AL
       can even be larger than that of PL, for large
       enough $\stdsig$.  We use $d=1000$, $\mean=3$ and $\nunl=10^5$ for all the
       experiments in this figure. The shaded areas indicate one standard deviation bands around the mean error, computed over 5 random draws of the seed set.}}
  \label{fig:synthvarmargin}
\end{figure*}

\subsection{Dependence on the standard deviation $\stdsig$ and seed size
$\nseed$ for regular margin-based sampling}
For completeness, we illustrate the dependence on the variance and the seed size
of regular margin-based sampling with following experiments. To simulate realistic settings, we set $d=1000$, $\nunl = 10^5$, $\nseed = 10$, $\stdsig = 3$ and $\mean = 3$. 

First, we perform a set of experiments to analyze the dependence on the variance
$\stdsig$ and to also confirm to main intuition empirically. We compute the
average distance to the decision boundary of the ground truth $\thetastar$ of a
labeled set acquired via M-AL and PL. Indeed, in Figure \ref{fig:synth_margin}
we see that the average margin of M-AL decreases with increasing $\sigma$.
Moreover, we note that M-AL indeed queries points close to the optimal decision
boundary. In Figure \ref{fig:synth_test_error} we observe that, as predicted by
Lemma~\ref{lem:alphabounds}, the decrease of the average margin gap is directly
correlated with an increase of the error gap between M-AL and PL. Hence, our
main intuition is also here empirically verifiable: M-AL queries points
relatively close to the ground truth, which causes in high dimensions the
max-$\ell_2$-margin classifier to rely more on the non-signal components to
classify the training data. 

\begin{figure*}[t]
    \centering
    \includegraphics[width=0.5\textwidth]{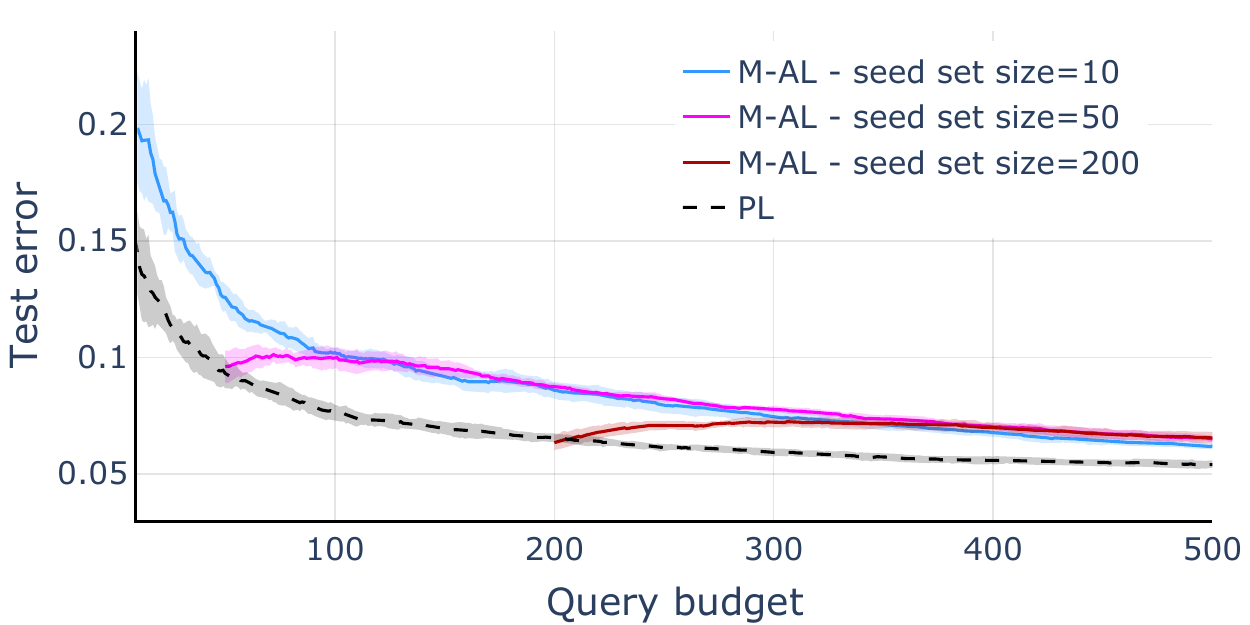}
    \caption{\small{ We set $d=1000$, $\mean=3$ and $\nunl=10^5$. The shaded
    areas indicate one standard deviation bands around the mean error, computed
over 5 random draws of the seed set. Observe that for increasing seed size the
gap between M-AL and PL closes, but does not vanish. Note that we study the
high-dimensional regime and hence only consider seed sizes up to $d/4$.
Therefore, seed size larger than $d$ may fully close the test error gap between
AL and PL.}}
    \label{fig:synth_seed_size}
  \end{figure*}

Secondly, we perform a set of experiments to an analyze the dependence on the
seed size $\nseed$. In Figure~\ref{fig:synth_seed_size}, we see that the test
error gap between M-AL and PL closes slowly for an increasing seed size.
However, we note that the gap remains non-zero for all seed sizes up to $d/4$.
\looseness=-1

\vspace{-0.2cm}
\section{Experiment details for tabular data}
\label{sec:appendix_exp_details}

\subsection{Datasets}
\label{sec:appendix_datasets}

To assess how suitable margin-based sampling is for high-dimensional data, we
conduct experiments on a wide variety of real-world datasets. 
We select datasets from OpenML \citep{OpenML2013} and from the UCI data
repository \citep{Dua2019} according to a number of criteria. In particular, we
focus on datasets for binary classification that are high-dimensional ($d >
100$) and which have enough samples that can serve as the unlabeled set ($\nunl
> \max(1000, 2d)$).  We do not consider text or image datasets where the
features are sequences of characters or raw pixels as estimators other than
linear models are better suited for these data modalities (e.g.\ CNNs,
transformers etc). Instead we want to analyze M-AL in a simple
setting and thus focus on datasets that are (approximately) linearly separable.
Moreover, we discard datasets that have missing values. Finally, we are left
with $\numdatasets$ datasets that cover a broad range of applications from
finance and ecology to chemistry and histology. We provide more details about
the selected datasets in Appendix~\ref{sec:appendix_datasets}.

To disentangle the effect of high-dimensionality from other factors such as
class imbalance, we subsample uniformly at random the examples of the majority
class, in order to balance the two classes. In addition, to ensure that the data
is noiseless, we fit a linear classifier on the entire dataset, and remove the
samples that are not interpolated by the linear estimator. This noiseless
setting is advantageous for active learning, since we are guaranteed to not
waste the limited labeling budget on noisy samples.  However, as we show later,
even in this favorable scenario, the performance of M-AL suffers in
high-dimensions.  For completeness, we also compare M-AL and PL on the original,
uncurated datasets in Appendix~\ref{sec:appendix_dirty_experiments} and observe
similar trends as in this section.

\begin{table}[h]
\scriptsize
\centering

\begin{tabular}{llrrrrr}
\toprule
Dataset name &    $d$ &  Training set size &  Test set size &  Majority/minority ratio &  Linear classif. training error \\
\midrule
                          a9a &    123 &              39074 &           9768 &                 3.17&                          0.1789 \\
                  vehicleNorm &    100 &              78823 &          19705 &                 1.00&                          0.1415 \\
                        nomao &    118 &              27572 &           6893 &                 2.50&                          0.0531 \\
santander                     &    200 &             160000 &          40000 &                 8.95&                          0.2188 \\
                  webdata\_wXa &    123 &              29580 &           7394 &                 3.16&                          0.1813 \\
                  sylva\_prior &    108 &              11516 &           2879 &                15.24&                          0.0011 \\
                     real-sim &  20958 &              57848 &          14461 &                 2.25&                          0.0027 \\
                     riccardo &   4296 &              16000 &           4000 &                 3.00&                          0.0007 \\
                    guillermo &   4296 &              16000 &           4000 &                 1.49&                          0.2536 \\
                      jasmine &    144 &               2388 &            596 &                 1.00&                          0.1867 \\
                     madeline &    259 &               2512 &            628 &                 1.01&                          0.3405 \\
                   philippine &    308 &               4666 &           1166 &                 1.00&                          0.2445 \\
                    christine &   1636 &               4335 &           1083 &                 1.00&                          0.1408 \\
                         musk &    166 &               5279 &           1319 &                 5.48&                          0.0438 \\
                      epsilon &   2000 &              48000 &          12000 &                 1.00&                          0.0947 \\
\bottomrule
\end{tabular}

\caption{Some characteristics of the uncurated datasets considered in our
experimental study.}
\label{table_stats}

\vspace{-0.5cm}
\end{table}

\paragraph{More dataset statistics.} Table~\ref{table_stats} summarizes some
important characteristics of the datasets. The datasets span a wide range of
applications (e.g.\ ecology, finance, chemistry, histology etc). All datasets
are high-dimensional ($d \ge 100$) and have sufficiently many training samples
that will serve as the unlabeled set. The test error is computed on a holdout
set, whose size we report in Table~\ref{table_stats}. We also present the
class-imbalance of the original, uncurated datasets and the training error of a
linear classifier trained on the entire dataset, which indicates the degree of
linear separability of the data.
\looseness=-1

\subsection{Methodology}

\vspace{-0.2cm}
We split each dataset in a test set and a training set. The covariates of the
training samples constitute the unlabeled set. We assume that the labels are
known for a small seed set of size $\nseed = 6$ (see
Appendix~\ref{sec:appendix_larger_seed} for experiments with larger seed sets).
For each experiment and dataset, we repeat the draw of the seed set 10 or 100
times, depending on the experiment.

For illustration purposes, we set the labeling budget to be equal to a quarter
of the number of dimensions.\footnote{Since the real-sim dataset has over 20,000
  features, we set a labeling budget lower than $d/4$, namely of only 3,000
  queries, for computational reasons.} We query one point at a time and select
  the sample whose label we want to acquire either via uniform sampling (i.e.\
  passive learning) or using margin-based sampling (i.e.\ active learning).

We use L-BFGS \citep{liu89} to train linear classifiers by minimizing the
logistic loss on the labeled dataset. In Appendix~\ref{sec:appendix_reg} we show
that the same high-dimensional phenomenon occurs when using $\ell_1$- or
$\ell_2$-regularized classifiers.



\vspace{-0.3cm}
\section{Additional experiments on tabular data}

\vspace{-0.1cm}
\subsection{Experiments on uncurated data}
\label{sec:appendix_dirty_experiments}

\vspace{-0.2cm}
For completeness, in this section we provide experiments on the original,
uncurated datasets. We distinguish two scenarios: 1) balanced data, but not
necessarily linearly separable; and 2) possibly imbalanced and not linearly
separable data. In both cases, we use the same methodology described in
Section~\ref{sec:experiments} to plot the probability (over draws of the seed
set) that the error with PL is lower than with M-AL and the losses/gains of M-AL
compared to PL.

\vspace{-0.2cm}
\paragraph{Balanced, but non-linearly separable data.} As indicated in
Appendix~\ref{sec:appendix_datasets}, not all datasets are originally linearly
separable. For clarity, in the experiments in the main text we curate the data
such that a linear classifier can achieve vanishing training error. This is
provides a clean test bed for comparing margin-based and uniform sampling in
high-dimensions.

\begin{figure}[H]
  \centering
  \begin{subfigure}[t]{\textwidth}
    \centering
    \includegraphics[width=0.7\textwidth]{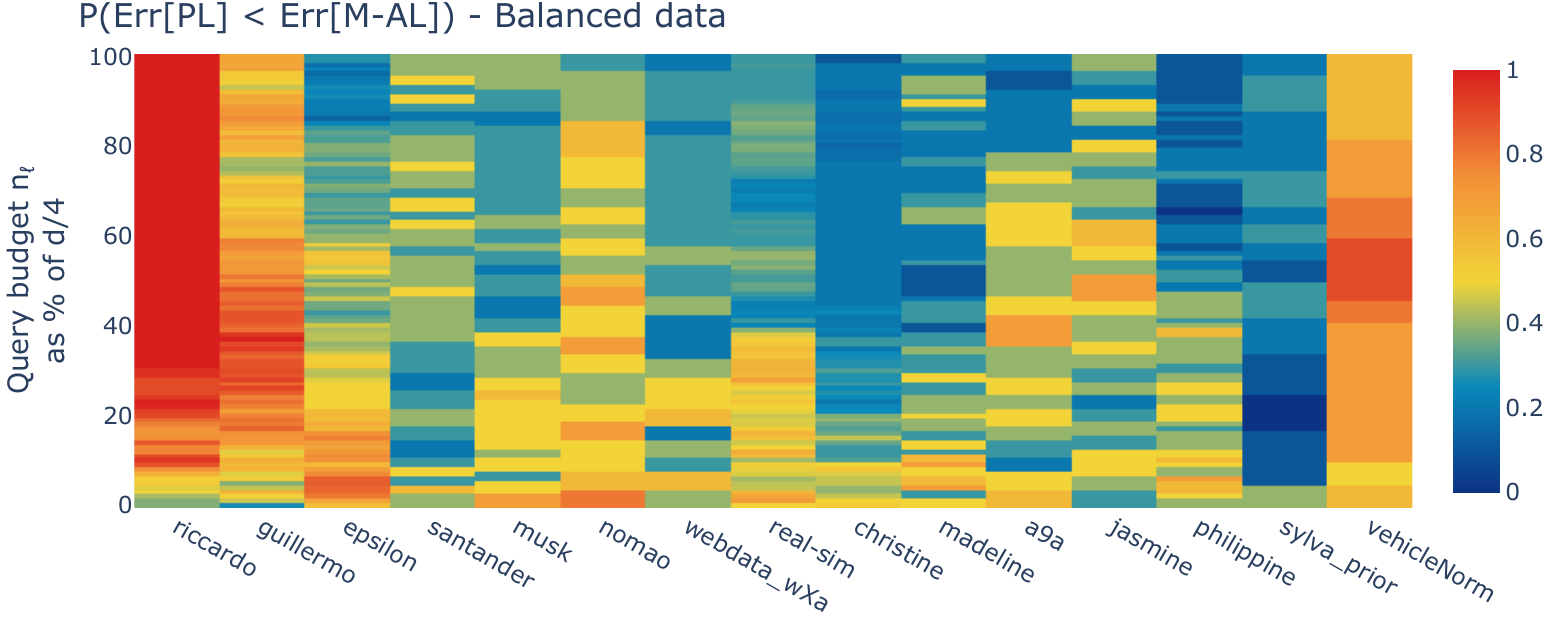}
  \end{subfigure}\\[-0.1cm]
  \begin{subfigure}[t]{\textwidth}
    \centering
    \includegraphics[width=0.7\textwidth]{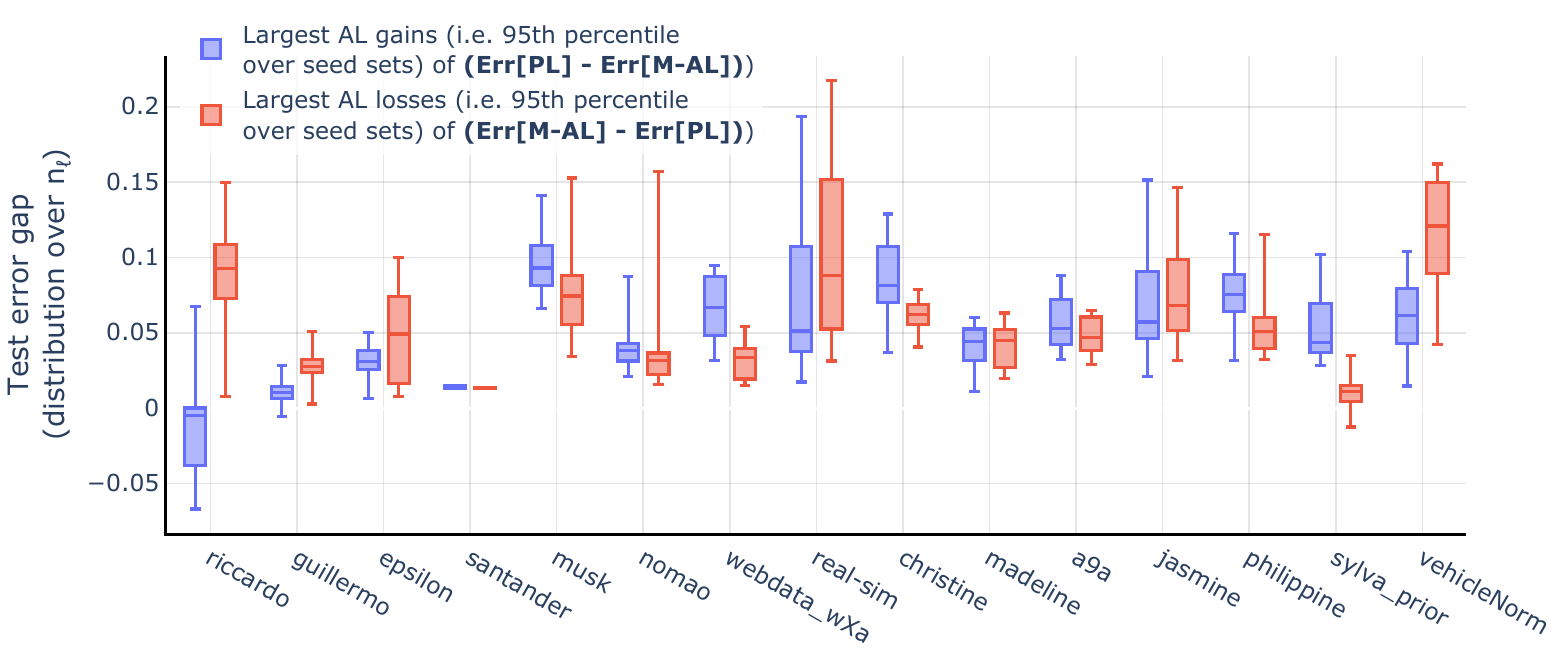}
  \end{subfigure}

  \caption{\textbf{Top:} The probability that the test error is lower with
    PL than with M-AL, over $10$ draws of the seed
    set. Data is class-balanced, but potentially not linearly separable.
  \textbf{Bottom:} For the range of budgets where M-AL does poorly with
    high probability, its sporadic gains over PL are generally
    similar or lower than the losses it can incur in terms of increased test
    error. Data is class-balanced, but potentially not linearly separable.}
  \label{fig:heatmap_dirty}

\vspace{-0.6cm}
\end{figure}

In Figure~\ref{fig:heatmap_dirty} we keep the datasets class-balanced, but allow
them to be potentially not linearly separable. We observe similar trends as the
ones illustrated in Figure~\ref{fig:main_exp} for the noiseless versions of the
datasets.

\begin{figure}[t]
  \centering
  \begin{subfigure}[t]{\textwidth}
    \centering
    \includegraphics[width=0.78\textwidth]{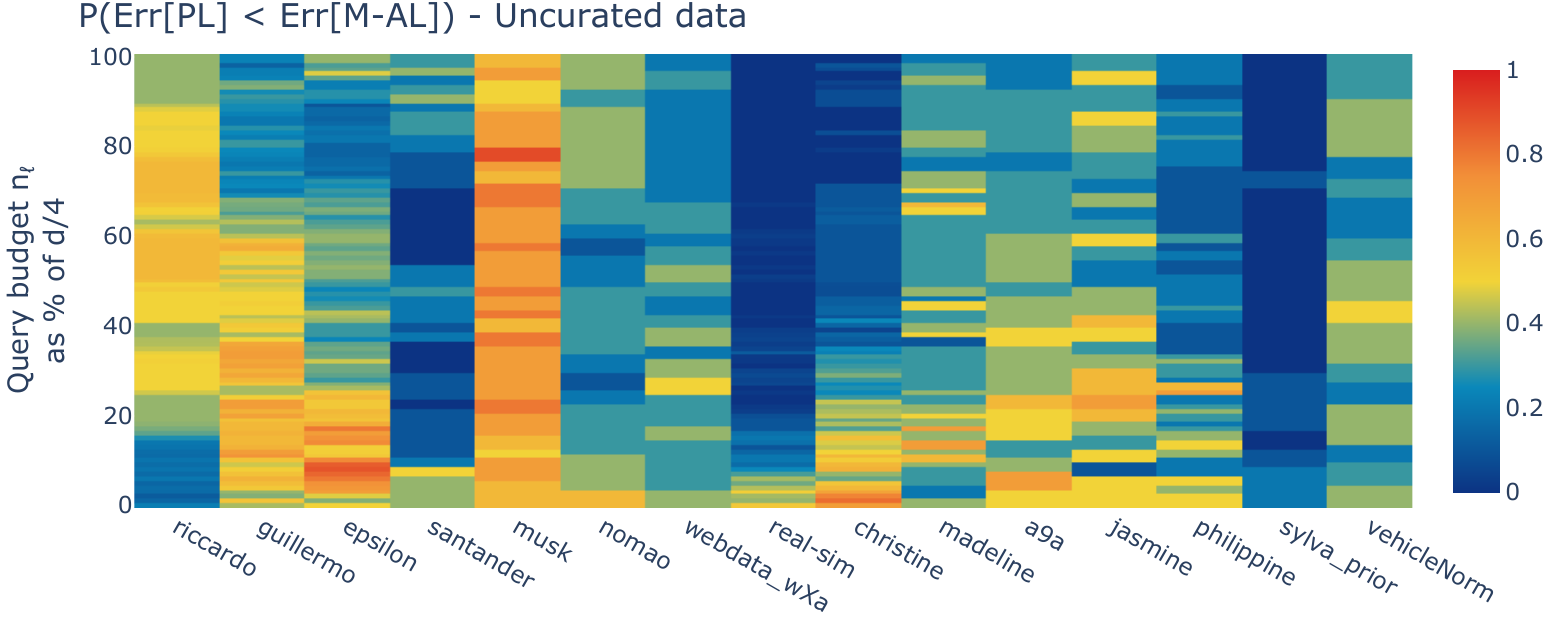}
  \end{subfigure}\\[-0.1cm]
  \begin{subfigure}[t]{\textwidth}
    \centering
    \includegraphics[width=0.78\textwidth]{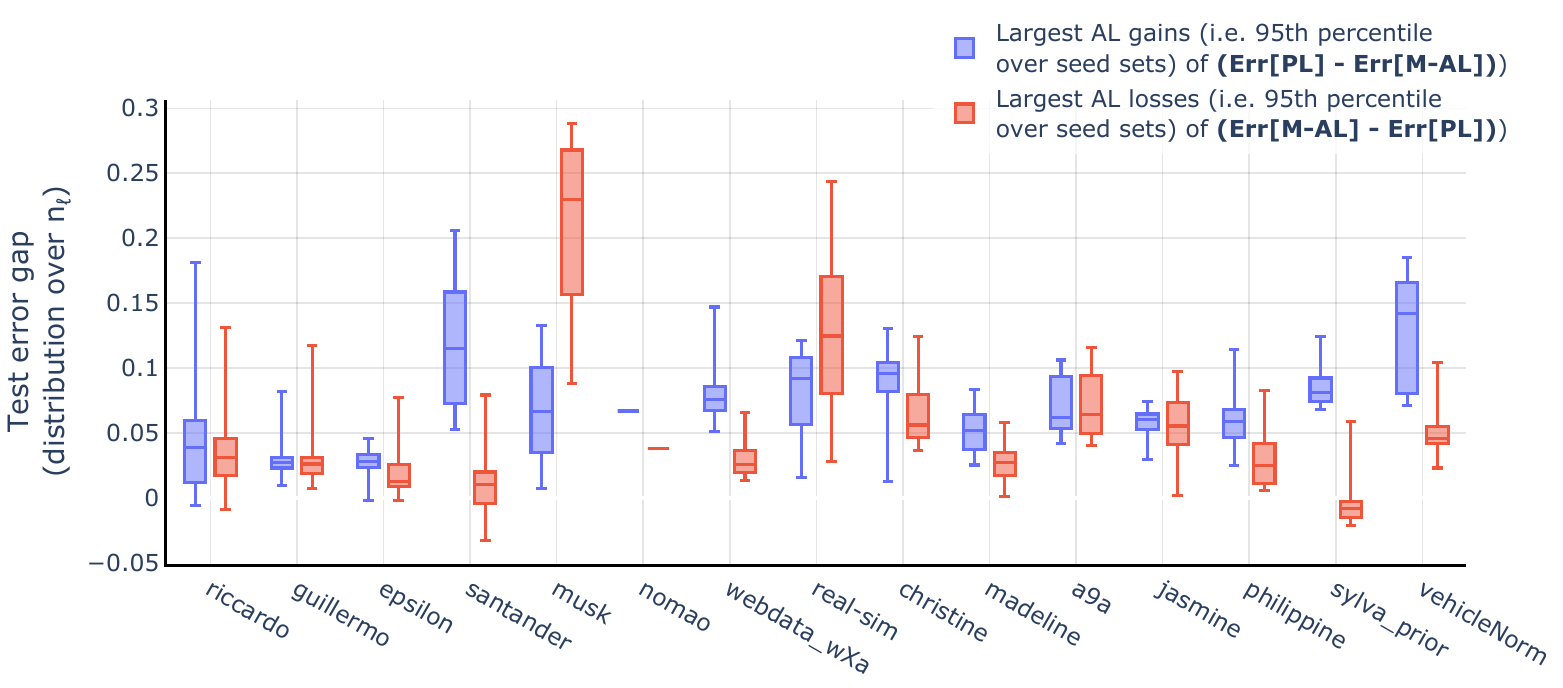}
  \end{subfigure}
\vspace{-0.4cm}
  \caption{\textbf{Top:} The probability that the test error is lower with
    PL than with M-AL, over $10$ draws of the seed
    set. Data is potentially class-imbalanced and not linearly separable.
  \textbf{Bottom:} For the range of budgets where M-AL does poorly with
    high probability, its sporadic gains over PL are generally
    similar or lower than the losses it can incur in terms of increased test
    error.}
  \label{fig:heatmap_og}

\vspace{-0.3cm}
\end{figure}

\vspace{-0.2cm}
\paragraph{Imbalanced and non-linearly separable data.} Margin-based sampling
brings about surprising benefits when applied on high-dimensional imbalanced
data. In particular, Figure~\ref{fig:heatmap_og} shows that for a broad range of
query budgets margin-based sampling leads to better test error than uniform
sampling. For these experiments we did not alter the original datasets in any
way, and kept all the training samples.

These results reveal a perhaps unexpected phenomenon. When the unlabeled
data is imbalanced (see Appendix~\ref{sec:appendix_datasets} for the exact
imbalance ratio of each dataset), M-AL tends to achieve better
predictive performance compared to passive learning.
This phenomenon has also been previously observed by \citet{ertekin}.

%
%



\vspace{-0.3cm}
\subsection{Uniform sampling versus oracle margin-based sampling}
\label{sec:appendix_heatmap_oracle}

\vspace{-0.3cm} In this section we provide the counterpart of
Figure~\ref{fig:main_exp}, but now we use the distance to the Bayes optimal
decision boundary for the active learning algorithm. Recall that for oracle M-AL
we first train a classifier on the entire labeled dataset (this estimator will
act as a stand-in for the Bayes optimal predictor). Then we use the distance to
the decision boundary determined by this approximation of the Bayes optimal
classifier to select points to query.

\begin{figure}[H]
  \centering
  \includegraphics[width=0.8\textwidth]{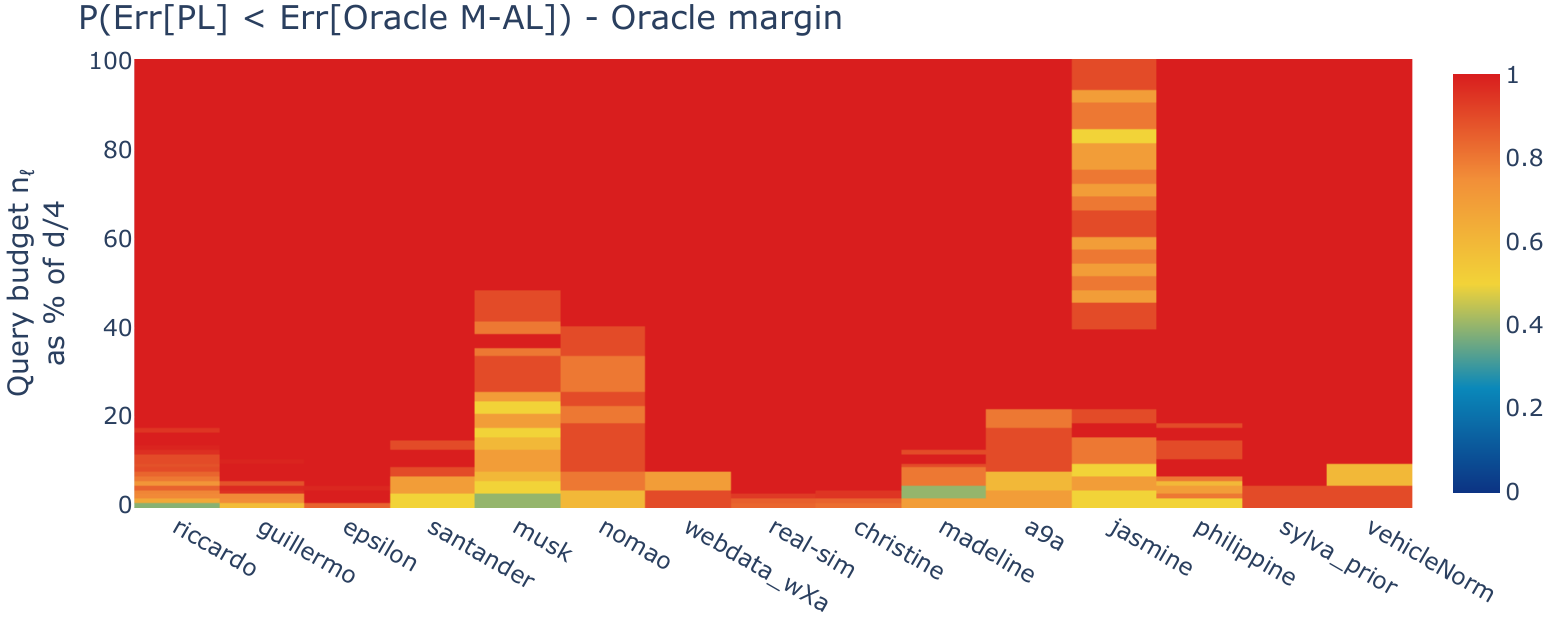}
\vspace{-0.2cm}
  \caption{The probability that the test error is lower with PL
  than with \textbf{oracle M-AL}, over $10$ draws of the seed
set. Oracle M-AL performs consistently worse than PL (warm-colored regions).}
\label{fig:appendix_oracle}

\vspace{-0.8cm}
\end{figure}

Figure~\ref{fig:appendix_oracle} reveals that the gap between M-AL and PL is
even more significant when using the oracle margin, which is in line with the
intuition provided in Section~\ref{sec:theory}. Oracle margin-based sampling will
select samples close to the Bayes optimal decision boundary (i.e.\ the yellow
points in Figure~\ref{fig:synth_sketch}).  Hence, the decision boundary of the
classifier trained on the labeled set collected with active learning will be
tilted compared to the optimal predictor, as long as the query budget is
significantly smaller than the dimensionality.

\vspace{-0.3cm}
\subsection{Test error at different query budgets -- more datasets}
\label{sec:appendix_error_vs_budget}

\begin{figure}[H]
  \centering
  \includegraphics[width=0.8\textwidth]{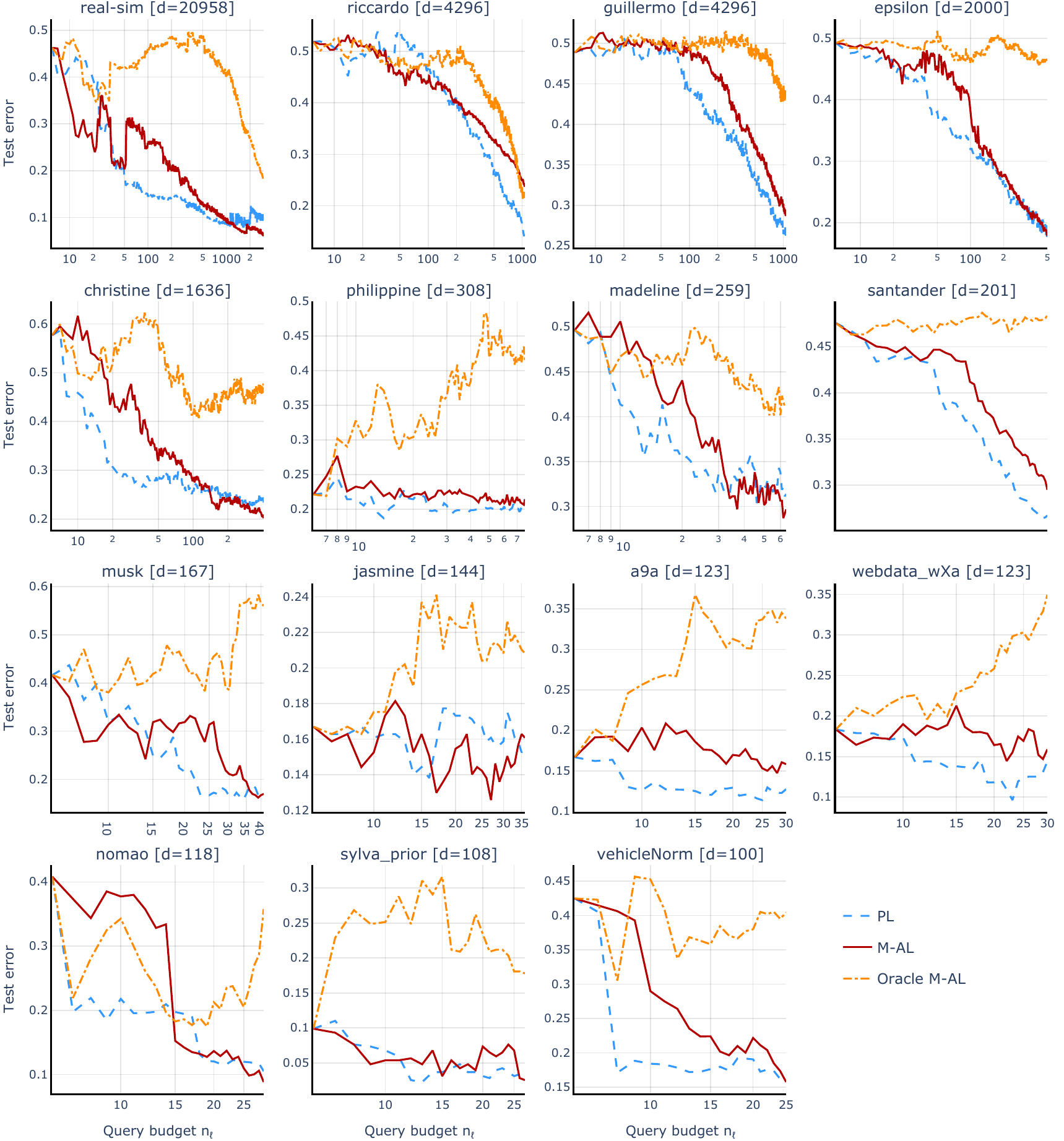}

  \caption{The test error using M-AL (with or without using the oracle margin)
    is often higher than what is achieved with uniform
sampling, for all the datasets that we consider.}
\label{fig:appendix_teaser}

\vspace{-0.9cm}
\end{figure}

\vspace{-0.2cm} We compare the test error of PL and M-AL, similar to
Figure~\ref{fig:teaser}, but for more real-world datasets. For margin-based
sampling, we use both the oracle margin and the margin of $\fhat$ as shown in
Algorithm~\ref{algo:us}. Figure~\ref{fig:appendix_teaser} show that oracle M-AL
consistently leads to larger test error compared to passive learning on all
datasets. In addition, using the distance to the decision boundary determined by
the max-$\ell_2$-margin classifier also leads to worse prediction performance,
in particular on the high-dimensional datasets and for small query budgets. For
illustration and computational purposes, we limit the query budget to
$\min(3000, d/4)$.

\pagebreak
\subsection{Another perspective on Figure~\ref{fig:main_exp}}
\label{sec:appendix_p_al_worse}

\begin{figure}[H]
  \centering
  \includegraphics[width=0.85\textwidth]{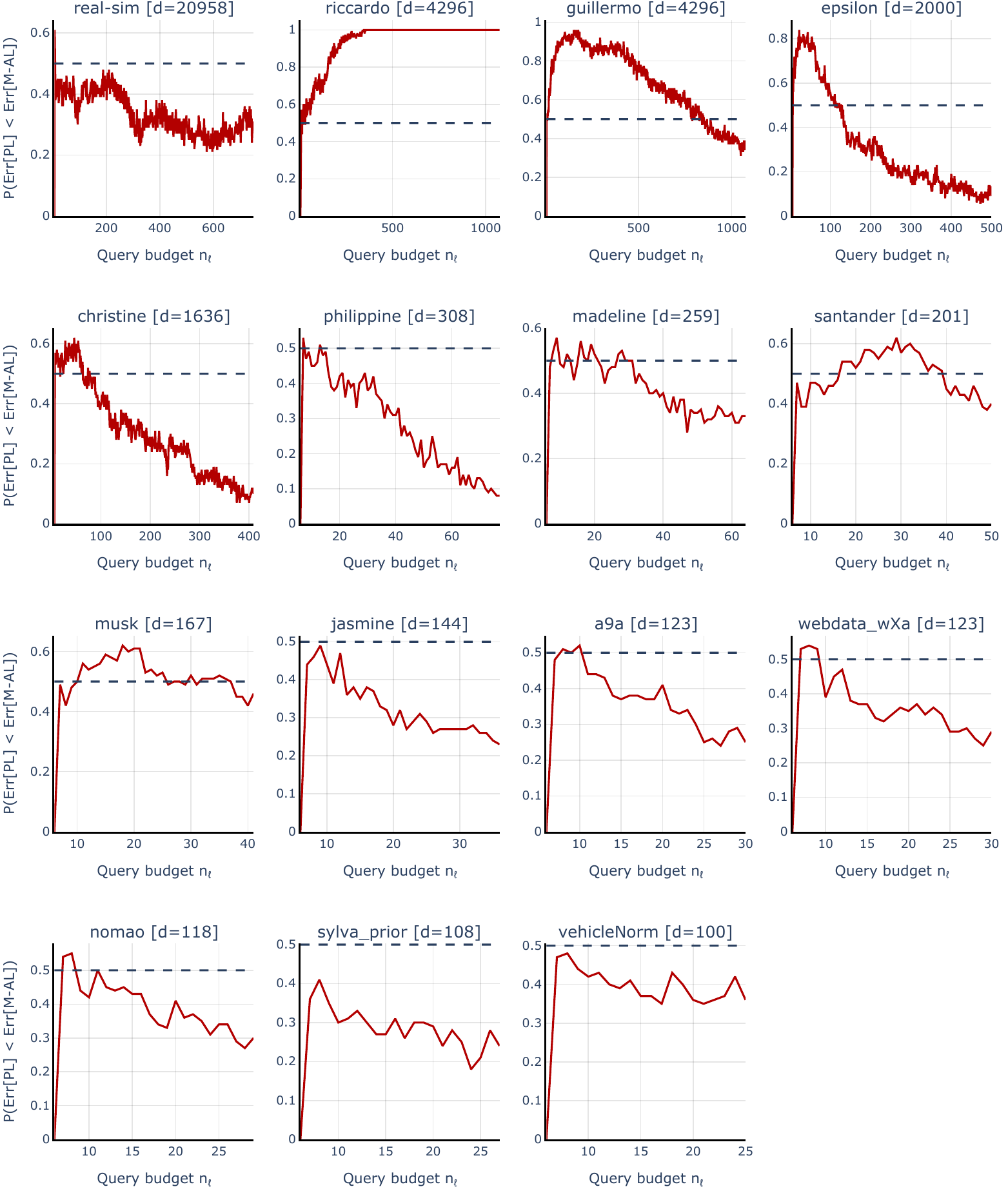}
  \caption{The probability that M-AL
  performs worse than PL at different query budgets. The
empirical probability is computed over $100$ draws of the initial seed set.}
  \label{fig:all_p_al_worse}
  \vspace{-0.9cm}
\end{figure}

\vspace{-0.2cm}
In Figure~\ref{fig:main_exp}-Top we provide an overview of the gap that exists
between M-AL and PL in high-dimensions. Here, we provide a
more detailed perspective of the same evaluation metric. Each panel in
Figure~\ref{fig:all_p_al_worse} corresponds to one column in
Figure~\ref{fig:main_exp}. The horizontal dashed line indicates the $50\%$
threshold at which the event that M-AL performs better is
equally likely to its complement. Notice that in all figures the solid line
starts at $0$, since before any queries are made, both uniform sampling and
margin-based sampling yield the same test error, namely the error of the
max-$\ell_2$-margin classifier trained on the seed set.

\vspace{-0.2cm}

We note that the spikes in the lines in Figure~\ref{fig:all_p_al_worse} come
from the fact that for different seed sets, M-AL may start to
underperform at different iterations. Hence, aggregating over several seed sets
leads to the non-smooth lines in the figure.

\begin{figure}[H]
  \centering
  \includegraphics[width=0.9\textwidth]{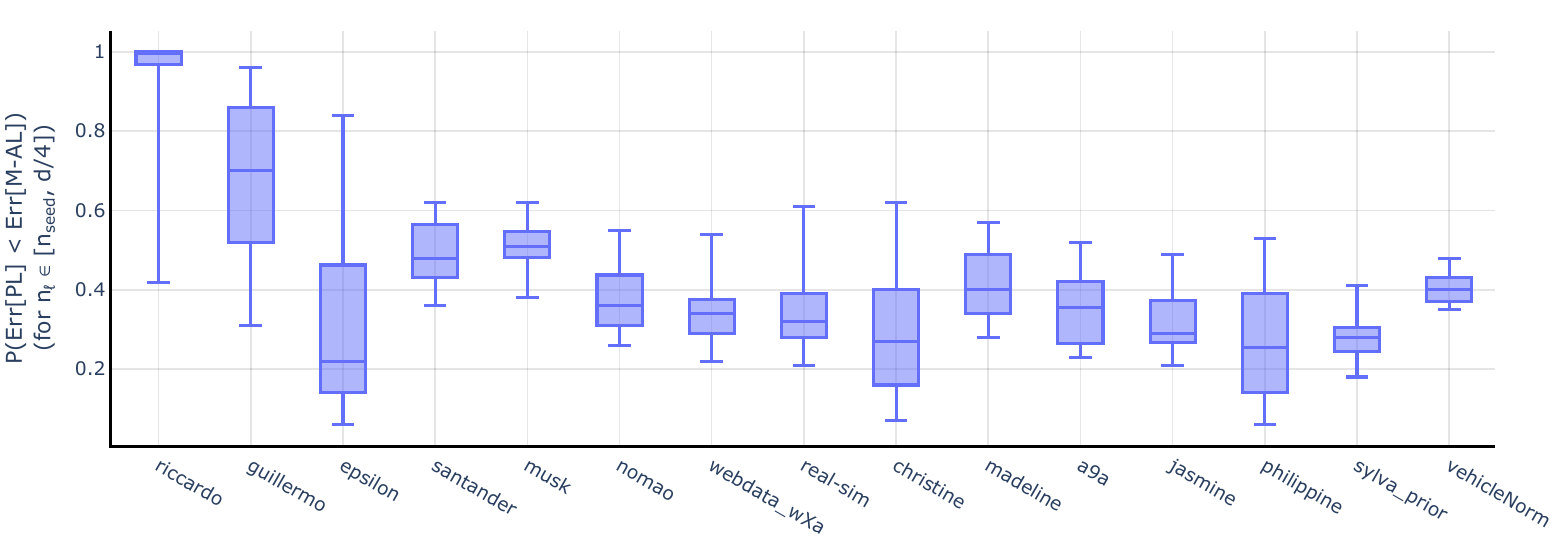}

  \caption{Box plot of the distribution of P(Err[PL] $<$ Err[AL]) over query budgets $\nq \in \{\nseed, ..., d/4\}$.}
  \label{fig:boxplot_p_al_worse}
  \vspace{-0.7cm}
\end{figure}

In addition, in Figure~\ref{fig:boxplot_p_al_worse} we summarize each of the
panels in Figure~\ref{fig:all_p_al_worse} in a box plot that offers yet another
perspective on this experiment. Notably, the boxes are fairly concentrated for
all datasets, confirming that the gap between the test error with uniform and
margin-based sampling stays roughly the same for any query budget $\nq \in
\{\nseed, ..., d/4\}$. Note that here the probability is over the draws of the
seed set, and the box plots show percentiles of the distribution over query
budgets for each dataset.

For these experiments we use the distance to the decision boundary determined by
the estimator $\fhat$ as shown in
Algorithm~\ref{algo:us}. In Figure~\ref{fig:main_exp}-Bottom we show the largest
gains and losses of M-AL for query budgets $\nq \in \{\nseed,
..., \ntrans\}$, where $\ntrans$ is defined as the budget after which
M-AL is always better than PL with probability at
least $50\%$. In other words, one can read $\ntrans$ off
Figure~\ref{fig:all_p_al_worse} as the leftmost point on the horizontal axis for
which the solid line intersects the horizontal dashed line. For datasets that
never intersect the $50\%$ dashed line, we take $\ntrans=d/4$ conservatively.
This is more advantageous for M-AL, as larger query budgets tend
to lead to larger gains over PL.



\subsection{Fraction of budgets for which active learning underperforms}
\label{sec:appendix_fraction_budgets}

\begin{figure}[H]
  \centering
  \includegraphics[width=0.9\textwidth]{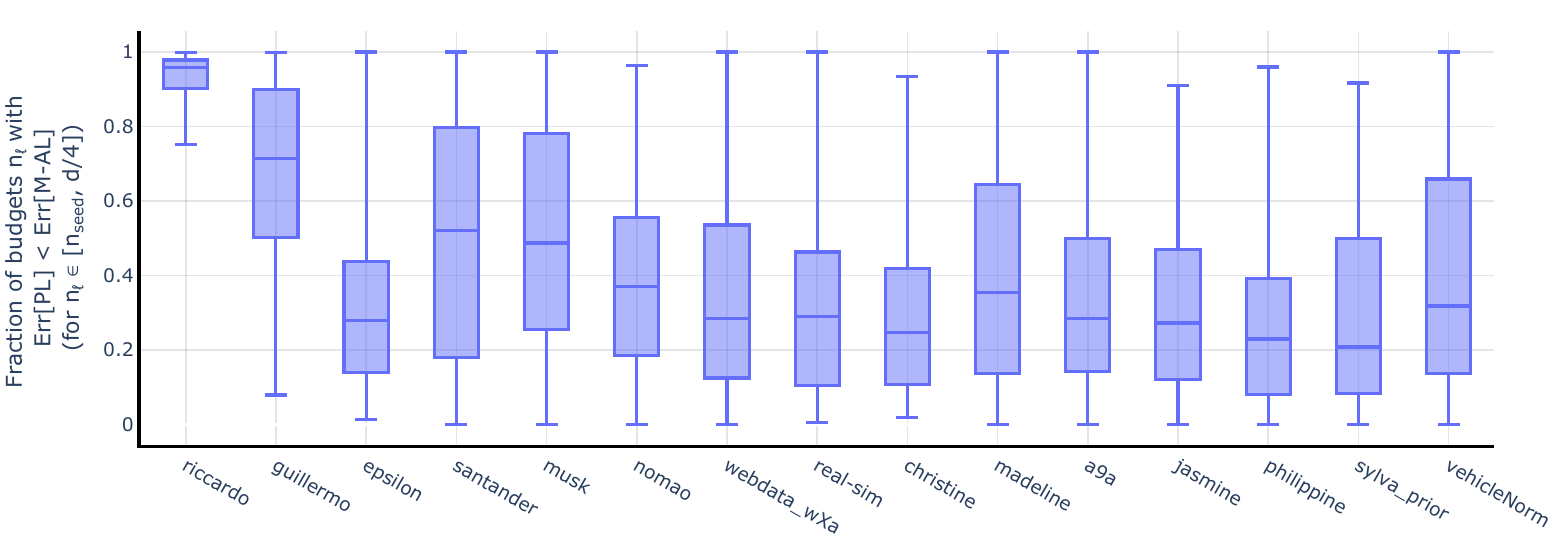}

  \vspace{-0.2cm}
  \caption{Fraction of the query budgets between $\nseed$ and $d/4$ for which
  the error with M-AL is worse than with PL. The
box plot indicates the distribution over $100$ draws of the seed set (median,
lower and upper quartiles).}
  \label{fig:fraction_budgets}

  \vspace{-0.7cm}
\end{figure}

An alternative to using the metric illustrated in Figure~\ref{fig:main_exp}-Top
and in Appendix~\ref{sec:appendix_p_al_worse} is to instead compute the fraction
of the query budgets for which active learning performs worse than passive
learning.  In Figure~\ref{fig:fraction_budgets} we present this evaluation
metric for all the datasets that we consider. The box plot indicates the
distribution over $100$ draws of the initial seed set. For all datasets and with
high probability over the draws of the seed data M-AL
underperforms on a large fraction of the query budgets between $\nseed$ and
$d/4$.

  \vspace{-0.2cm}

Note that the fences of the box plots that almost cover the entire $[0, 1]$
range are a consequence of having a large number of runs (i.e.\ $100$). The
whiskers indicate the minimum and maximum values and they will be more extreme,
the larger the set over which we take the minimum/maximum is.
\looseness=-1


\subsection{Experiments with regularized estimators}
\label{sec:appendix_reg}

The failure case of margin-based active learning that we discuss in this
paper is not limited to the situation when we use interpolating estimators.
Indeed, as we show here, even regularization in the form of an $\ell_2$ or
$\ell_1$ penalty still leads to classifiers with high test error when the data
is collected using margin-based sampling.

We not that, in what follows, a small coefficient $C$
corresponds to stronger regularization, since we employ the scikit-learn
\citep{scikit-learn} implementation of penalized logistic regression. Therefore
$C \to 0$ implies the predictive error term in the loss is ignored, while $C\to
\infty$ leads to no regularization (note that unless otherwise specified, all
results throughout the paper are reported for the unregularized
max-$\ell_2$-margin classifier).

\begin{figure}[H]
  \centering
  \includegraphics[width=\textwidth]{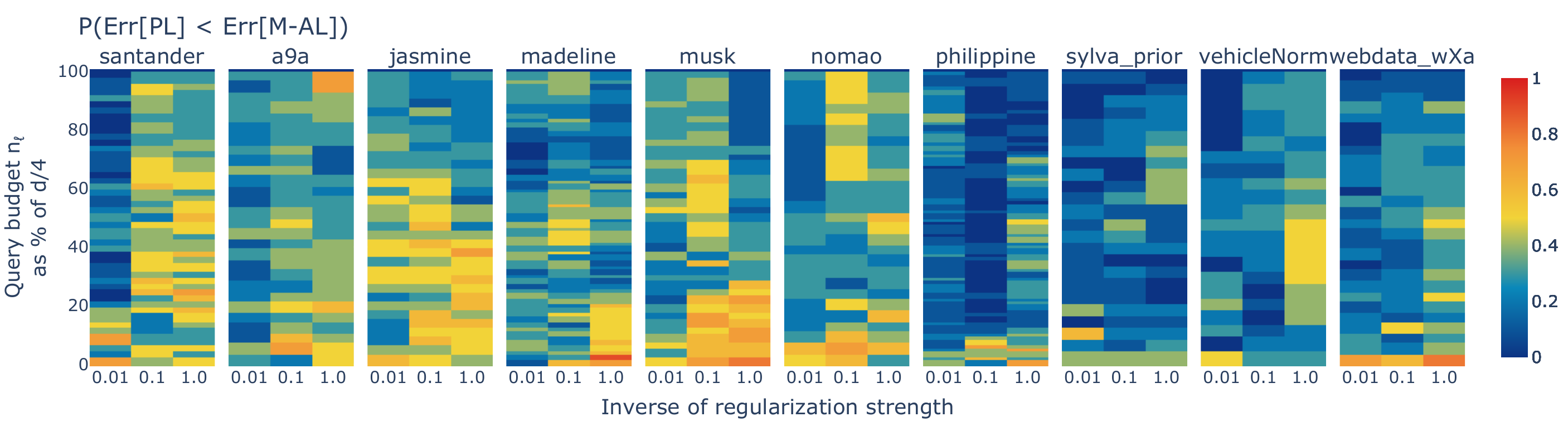}

  \vspace{-0.2cm}
  \caption{The probability that the test error is lower with PL
  than with M-AL, over $10$ draws of the seed
  set. We use an \textbf{$\ell_2$-regularized} classifier for both prediction and
margin-based sampling. Note that smaller values along the x-axis correspond to
stronger regularization. If we regularize too much (e.g.\ for a coefficient of
$0.01$), the prediction error is poor for both PL and M-AL, which explains the
light-colored regions.}
  \label{fig:l2}

\end{figure}

\begin{figure}[H]
  \centering
  \includegraphics[width=\textwidth]{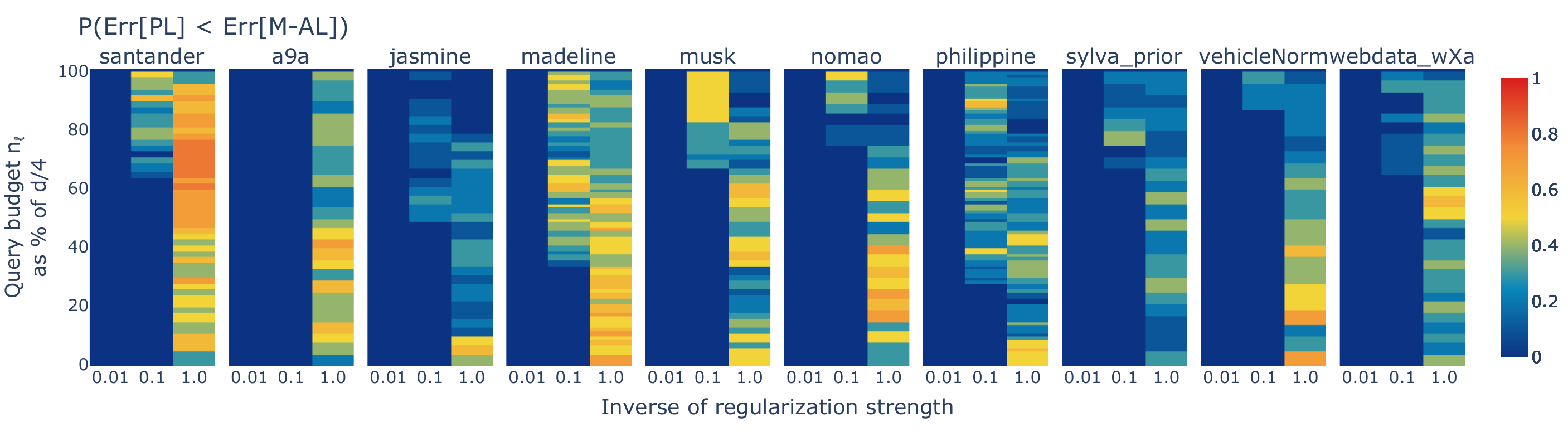}

  \vspace{-0.2cm}
  \caption{The probability that the test error is lower with PL than with M-AL,
    over $10$ draws of the seed set. We use an \textbf{$\ell_1$-regularized}
    classifier for both prediction and margin-based sampling. Note that smaller
    values along the x-axis correspond to stronger regularization. If we
  regularize too much (e.g.\ for a coefficient of $0.01$), the prediction error
is poor for both PL and M-AL, which explains the light-colored regions.}
\label{fig:l1}

  \vspace{-0.5cm}
\end{figure}

%
Figures~\ref{fig:l2} and~\ref{fig:l1} indicate that for strong enough
regularization, the gap between the test error of M-AL and PL vanishes. This
outcome is expected since stronger regularization leads to a poorer fit of the
data, and hence, classifiers trained on different data sets (e.g.\ one collected
with M-AL and another collected with PL) will tend to be similar. The downside
of increasing regularization is, of course, worse predictive performance. For
instance, for an $\ell_1$ penalty and a coefficient of $0.01$, the test error is
close to that of a random predictor (i.e.\ $50\%$) an all datasets for both
uniform and margin-based sampling. For moderate regularization, there continue to
exist broad ranges of query budgets for which M-AL underperforms
compared to passive learning.

  \vspace{-0.2cm}

  \vspace{-0.4cm}


\subsection{Experiments with different seed set sizes}
\label{sec:appendix_larger_seed}

\begin{figure}[H]
  \centering
  \includegraphics[width=0.8\textwidth]{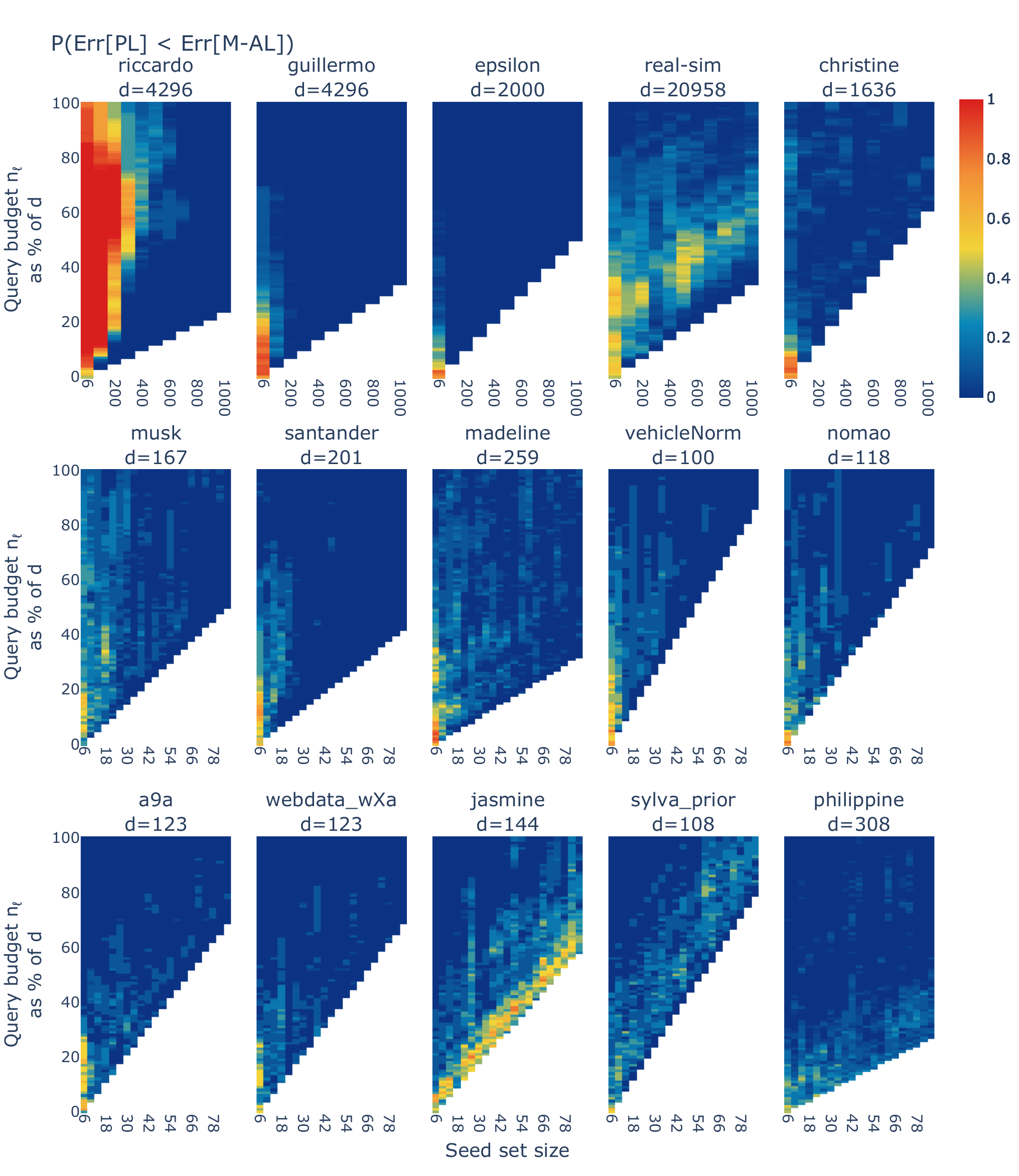}

  \caption{As predicted by our theory, increasing the seed size leads to
  improved performance when using M-AL to acquire new labeled
samples.}
  \label{fig:larger_seed}

\vspace{-0.8cm}
\end{figure}

\begin{figure}[H]
  \centering
  \includegraphics[width=0.8\textwidth]{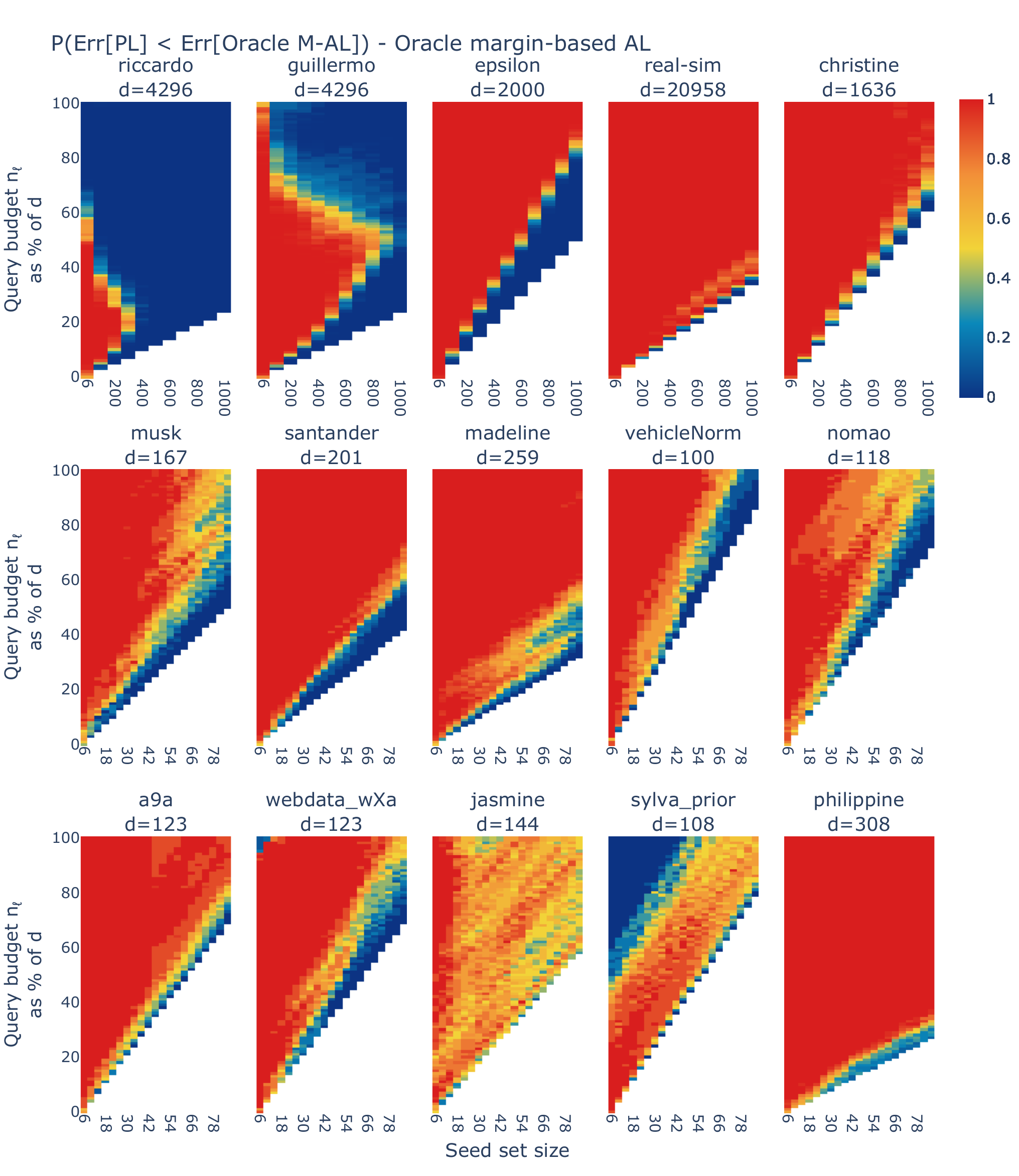}

  \caption{Surprisingly, increasing the seed set size also benefits oracle
  M-AL. This trend is not predicted by prior work and underlines
a novel failure case of M-AL for high-dimensional data.}
  \label{fig:larger_seed_oracle}

\vspace{-0.8cm}
\end{figure}

Our theory predicts that for a fixed labeling budget and an increasing see set
size, the gap between the error with margin-based sampling and uniform sampling
vanishes.\footnote{Note that if the seed set size matches the labeling budget
  M-AL is trivially equivalent to PL, since no
queries are issued.} We verify this insight experimentally in
Figures~\ref{fig:larger_seed} and~\ref{fig:larger_seed_oracle}. Our empirical
findings confirm the trend predicted by our theory: margin-based sampling leads
to better performance for large seed set sizes, but underperforms for small seed
sets.

Perhaps surprisingly, the same trend occurs even for oracle M-AL. This is
noteworthy, since prior work suggests that the failure of margin-based sampling
for small seed set sizes is due to the usage of a meaningless score: the
distance to a potentially very incorrect decision boundary obtained after
training a predictor on the small seed set.  Instead, our results show that
margin-based sampling fails even when using the Bayes optimal predictor, which
highlights a novel failure case of this sampling strategy.
%

\subsection{Combining informativeness and representativeness}
\label{sec:appendix_eps_greedy}

In this section we provide evidence that the shortcoming of margin-based
sampling that we identify in this paper also extends to other active learning
strategies that try to balance exploration and exploitation. In particular, we
focus on an $\eps$-greedy strategy which selects points using margin-based
sampling with probability $1-\eps$, and samples points uniformly at random with
probability $\eps$. Hence, this approach combines selecting informative samples
via M-AL with collecting a labeled set that is representative of
the training distribution. This strategy resembles the works of
\citet{brinker03, huang14, yang15, gissin19, shui20}.

First, we note that for oracle M-AL, the $\eps$-greedy strategy
is equivalent to simply selecting a larger uniform seed set, since the queries
are independent of each other when we use the oracle margin.
Therefore, for a fixed query budget $\nlab$, the $\eps$-greedy strategy with
oracle M-AL is identical to regular oracle M-AL sampling where
$\nseed = \eps \cdot \nlab$. We conclude that the results in
Section~\ref{sec:appendix_larger_seed}, and more specifically
Figure~\ref{fig:larger_seed_oracle}, show that the $\eps$-greedy strategy
performs worse than uniform sampling when using oracle M-AL.

\begin{figure}[H]
  \centering
  \includegraphics[width=0.75\textwidth]{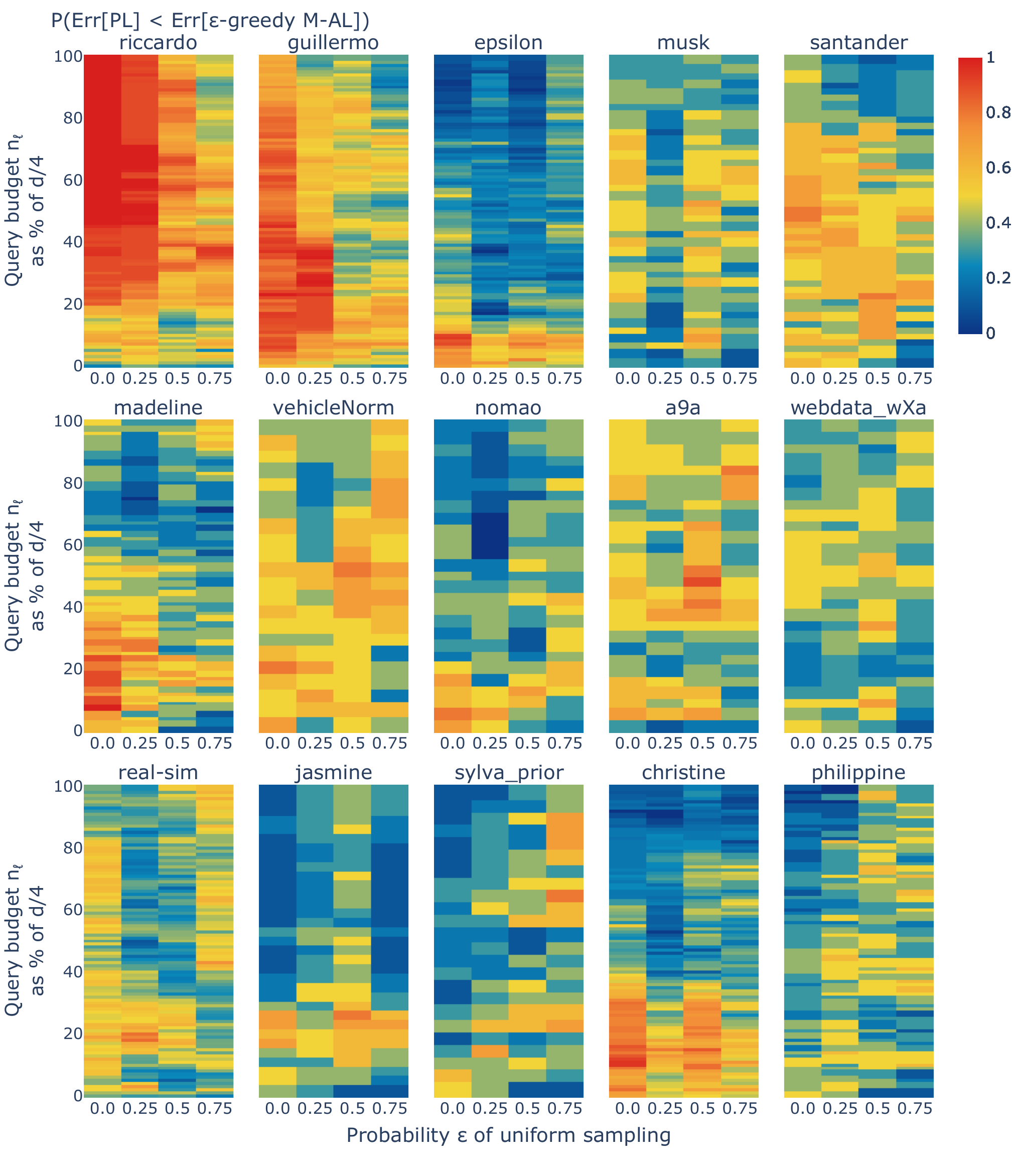}

  \vspace{-0.4cm}
  \caption{The probability that the test error is lower with uniform sampling
  than with an $\eps$-greedy sampling approach, over $10$ draws of the seed
set. The active learning strategy performs margin-based sampling, with probability
$1-\eps$ and samples uniformly at random with probability $\eps$.}
\label{fig:eps_greedy}

  \vspace{-0.2cm}
\end{figure}

Finally, we check whether the $\eps$-greedy strategy using the margin of
the empirical predictor $\thetahat$ is also detrimental compared to passive
learning.  We notice in Figure~\ref{fig:eps_greedy} that for different values of
$\eps$, active learning continues to perform worse than passive learning.
Varying $\eps$ between $0$ and $1$ effectively interpolates between vanilla
margin-based sampling and uniform sampling.

  \vspace{-0.2cm}


  \vspace{-0.4cm}
\subsection{Coreset-based active learning}
\label{sec:appendix_coreset}

In this section we investigate whether the coreset-based sampling strategy
proposed in \citet{sener18} can be a viable alternative to margin-based sampling
in low-sample regimes. We follow the same active learning methodology as
described in Section~\ref{sec:experiments}, but use the greedy algorithm from
\citet{sener18} to select queries. We use the Euclidean distance for our
experiments.

\begin{figure}[H]
  \centering
  \includegraphics[width=0.7\textwidth]{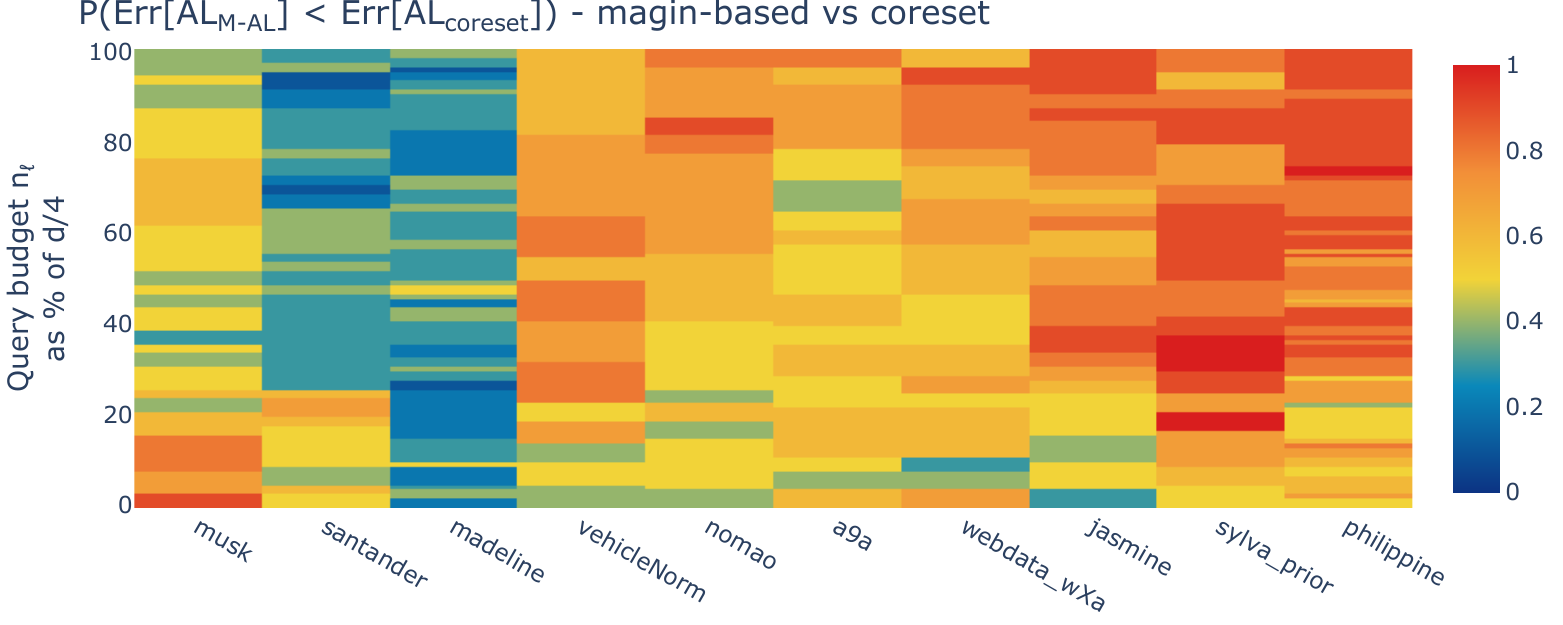}

  \vspace{-0.2cm}
  \caption{Coreset AL \citep{sener18} sometimes
  outperforms M-AL with high probability (blue regions).}
    \label{fig:coreset_vs_unc}
\end{figure}

\begin{figure}[H]
  \centering
  \includegraphics[width=0.7\textwidth]{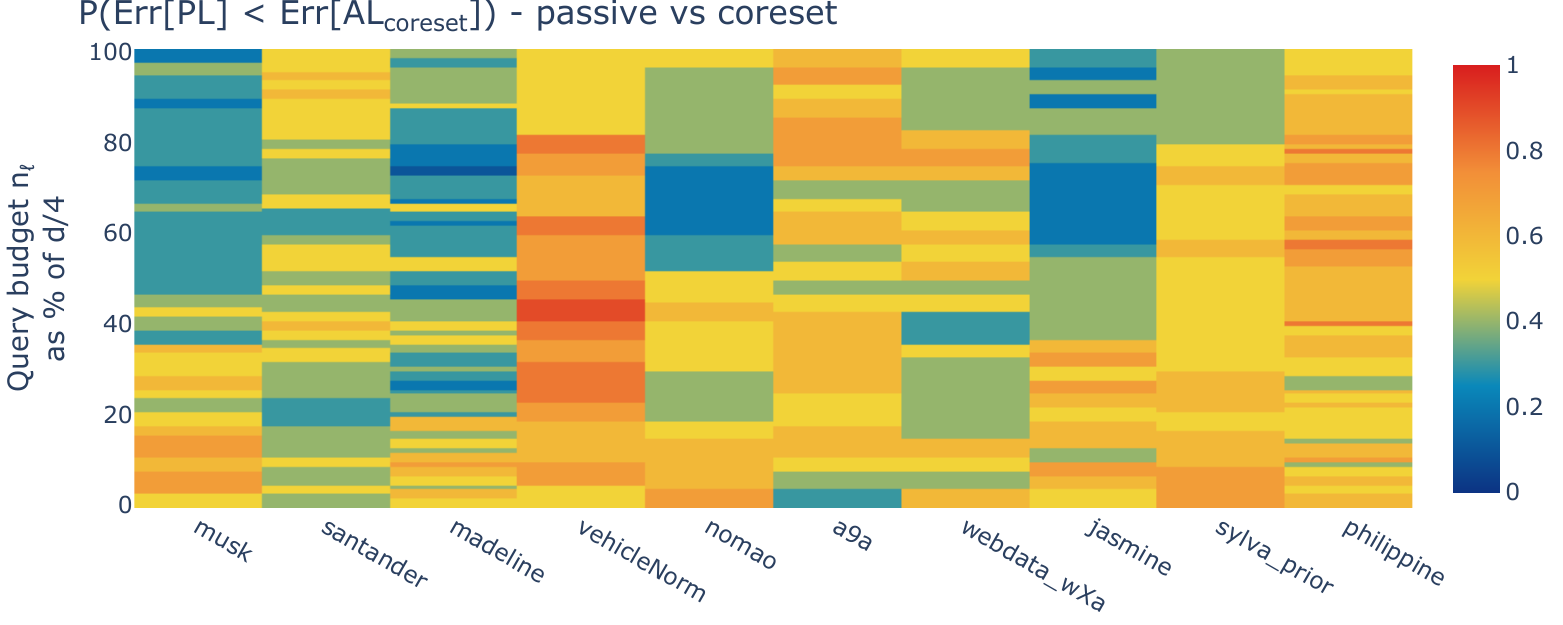}

  \vspace{-0.2cm}
  \caption{On some datasets (e.g.\ \emph{vehicleNorm}), PL
    outperforms coreset-based active learning in the low-sample regime.}
    \label{fig:coreset_vs_unif}
\end{figure}

Figure~\ref{fig:coreset_vs_unc} shows that for a large fraction of query
budgets, coreset-based active learning outperforms M-AL with
high probability (warm-colored areas). However, for some datasets (e.g.\
\emph{vehicleNorm, a9a, philippine}), coreset-based sampling can still lead to
larger error than passive learning, as illustrated in
Figure~\ref{fig:coreset_vs_unif}.
We hypothesize that this behavior is due to not constraining the queried points
to lie far from the Bayes optimal decision boundary. Hence, the high-dimensional
phenomenon that we describe in Section~\ref{sec:theory} still occurs.

  \vspace{-0.4cm}


  \vspace{-0.4cm}
\section{Experiments on image datasets}
\label{sec:appendix_image}

In this section we describe our experiments on image datasets in which we
explore the limitations of margin-based sampling for low query budgets.

\subsection{Experiment details}

\vspace{-0.2cm}
We consider three standard image datasets: CIFAR10 \citep{cifar}, CIFAR100
\citep{cifar}, SVHN \citep{svhn}. In addition to these, we also run experiments
on a binary classification task for medical images (PCAM \citep{Veeling2018})
and on a 10-class task on satellite images (EuroSAT \citep{helber2017}). For
prediction and for the sampling strategy we use ResNet18 networks \citep{He2015}
and start from weights pretrained on ImageNet. The sampling strategy consists in
selecting the unlabeled point on which the trained prediction model has lowest
confidence, as indicated by the softmax outputs. We note that this
uncertainty-based strategy is closely related to the margin-based strategy that
we use in binary classification problems.

To get a good estimate of the Bayes optimal classifier for Oracle M-AL, we train
on the entire labeled training set for each dataset until the training error
reaches $0$.  We consider batch active learning, as usual in the context of deep
learning, and set the batch size to $20$ (experiments with larger batch sizes
lead to similar results). For each dataset, we start from an initial seed set of
$100$ labeled examples and perform $50$ queries. Hence, the largest query budget
that we consider is of $1100$ labeled samples. After each query step, we
fine-tune the ResNet18 model for $20$ epochs, and achieve $0$ training error.
For fine-tuning we use SGD with a learning rate of $0.001$ and momentum
coefficient of $0.9$.

\vspace{-0.3cm}
\begin{figure}[h]
  \centering
  \begin{subfigure}[t]{0.7\textwidth}
    \centering
    \includegraphics[width=\textwidth]{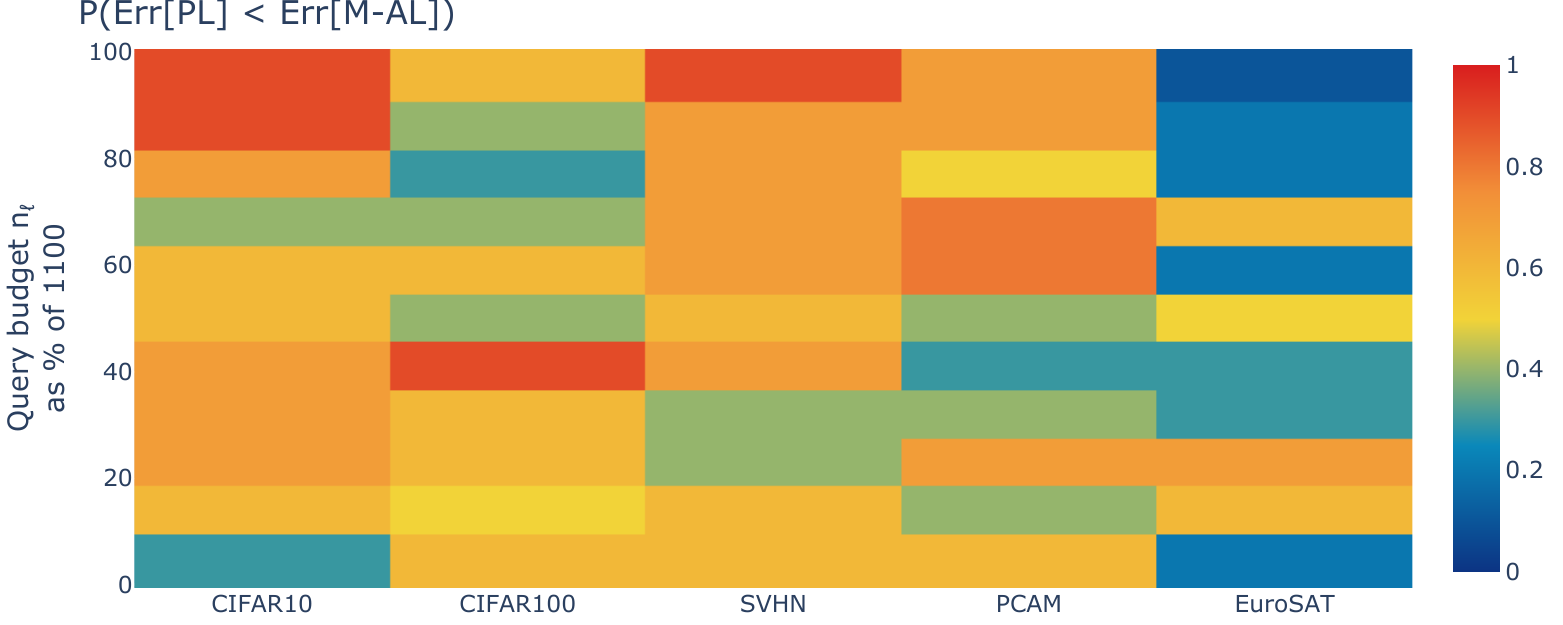}
  \end{subfigure}\\[-0.1cm]
  \begin{subfigure}[t]{0.7\textwidth}
    \centering
    \includegraphics[width=\textwidth]{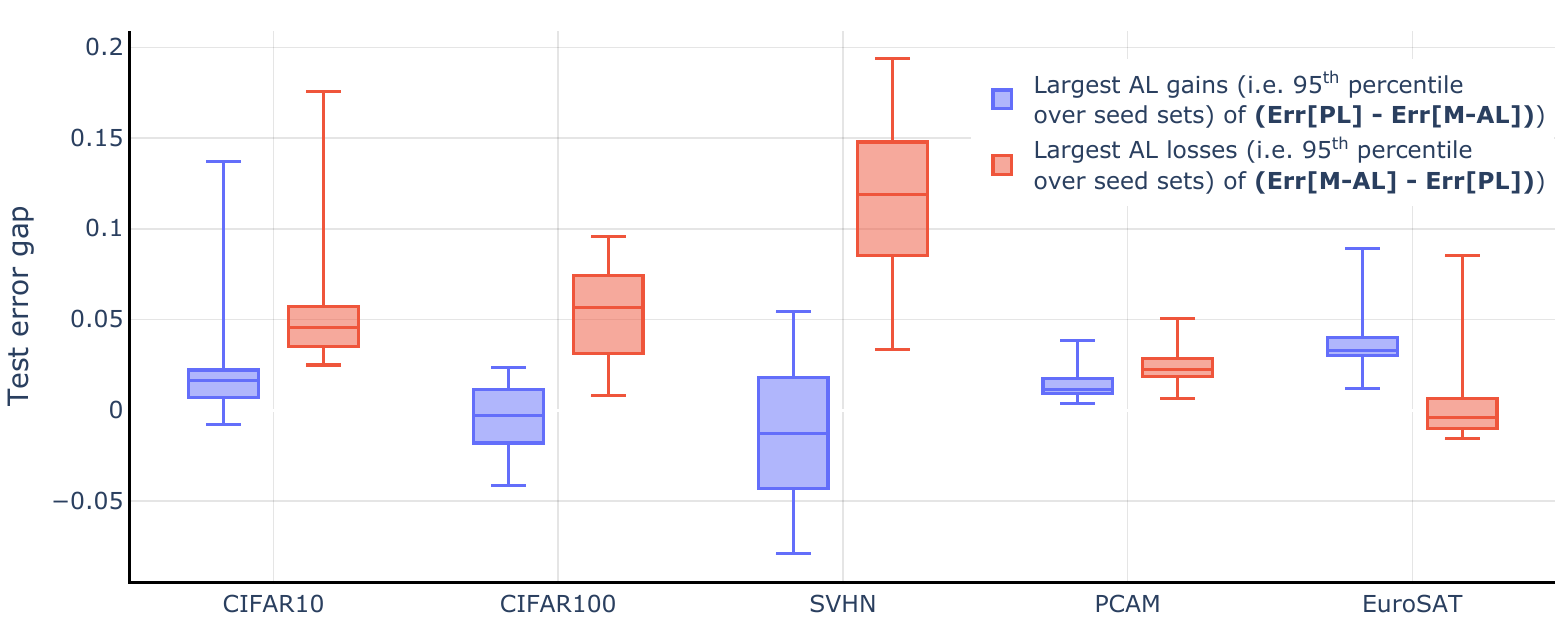}
  \end{subfigure}

\vspace{-0.1cm}
  \caption{\textbf{Top:} Probability that the test error is lower with PL
    versus M-AL, over 10 different random seeds. PL outperforms M-AL, for a
    significant fraction of the query budgets (i.e.\
    warm-colored regions). \textbf{Bottom:} The sporadic gains of M-AL over PL are
  generally similar or lower than the losses it can incur in terms of increased
test error (negative values indicate that PL is always better than AL). }
\label{fig:nn}

\vspace{-0.5cm}
\end{figure}

\begin{figure}[H]
  \centering
  \begin{subfigure}[t]{0.7\textwidth}
    \centering
    \includegraphics[width=\textwidth]{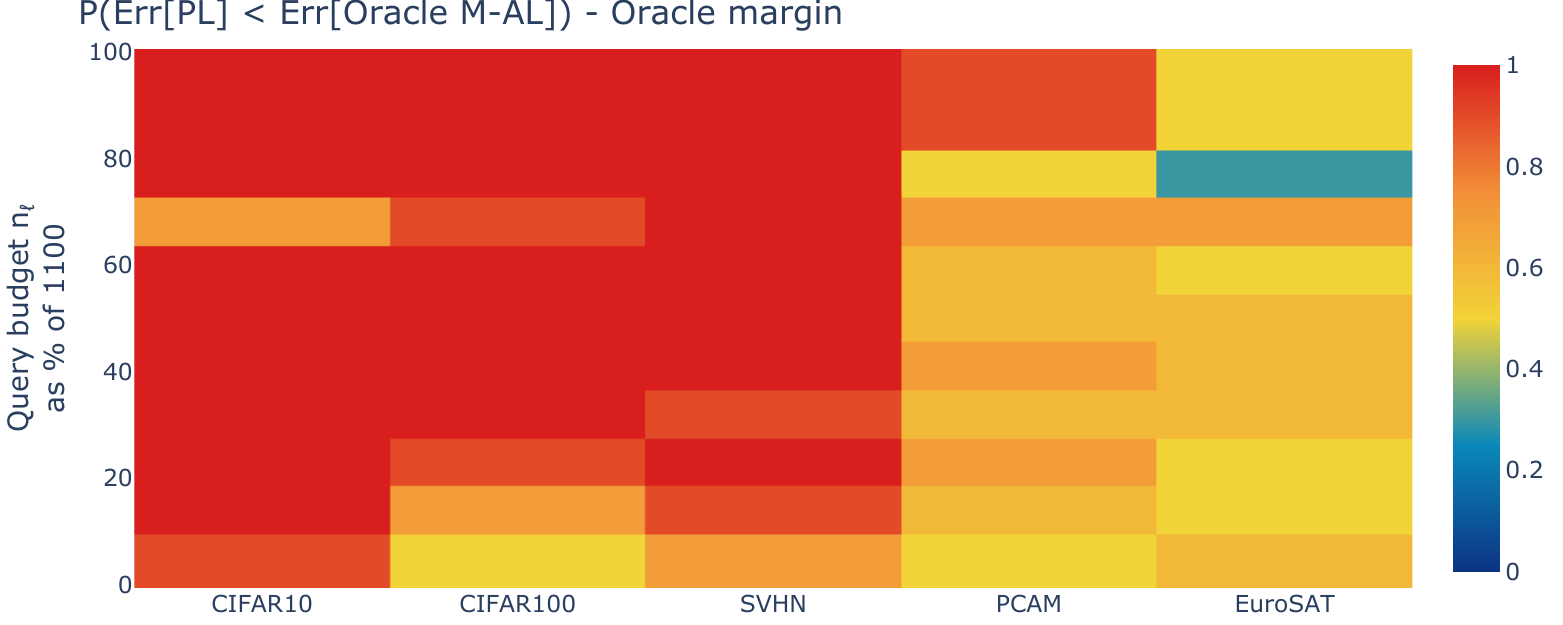}
  \end{subfigure}\\[-0.1cm]
  \begin{subfigure}[t]{0.7\textwidth}
    \centering
    \includegraphics[width=\textwidth]{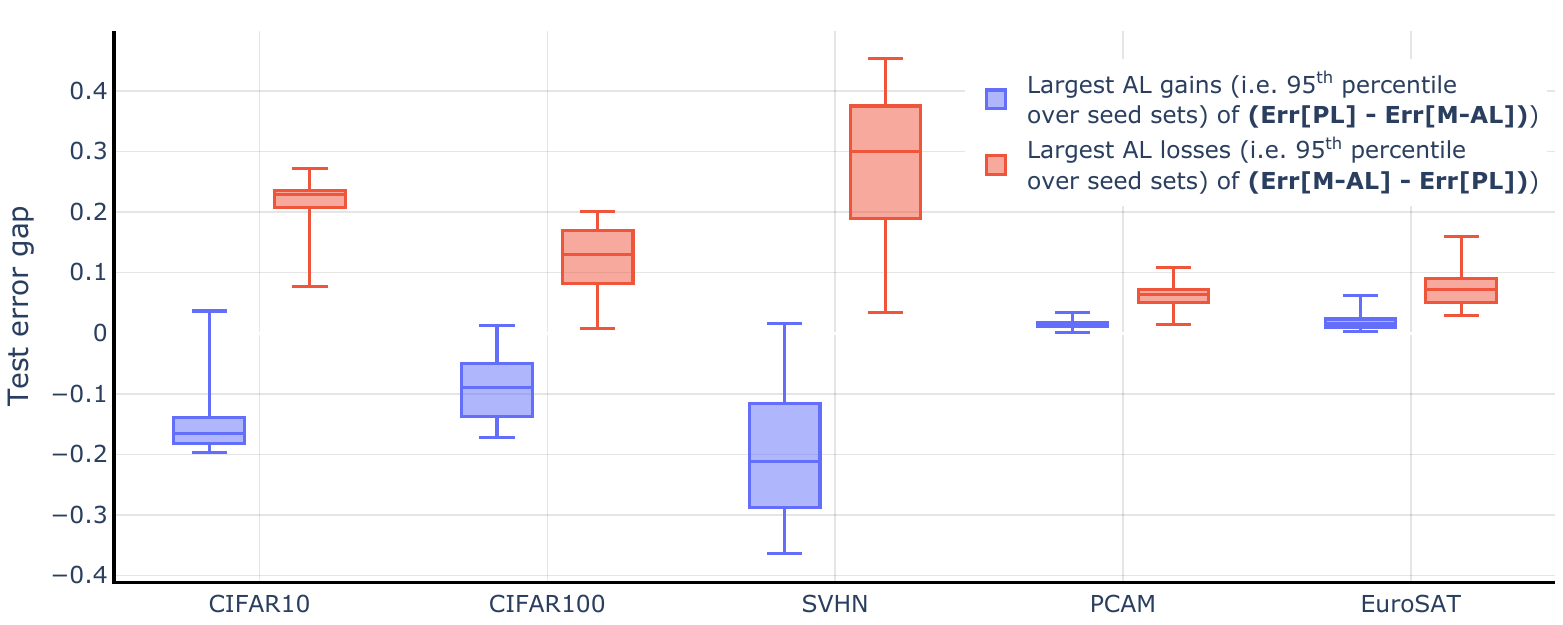}
  \end{subfigure}

\vspace{-0.1cm}
\caption{Same experiment as in Figure~\ref{fig:nn}, but this time
  using \textbf{oracle M-AL}. Similar to the logistic
regression experiments, M-AL leads to even worse error when
using the Bayes optimal classifier for sampling, as predicted by our theory.}
  \label{fig:nn_oracle}

  \vspace{-0.5cm}
\end{figure}

\vspace{-0.2cm}
\paragraph{Summary of results.} As illustrated in Figure~\ref{fig:nn}, AL leads
to significantly larger test error compared to PL.  This phenomenon persists
even when we use the Bayes optimal classifier (Figure~\ref{fig:nn_oracle}).
Moreover, the gains that the active learning strategy can produce, are often
dominated by the losses that it can incur. Note that for
Figure~\ref{fig:nn}-Bottom and Figure~\ref{fig:nn_oracle}-Bottom we take
$\ntrans= 1100$, namely the maximum query budget $\nlab$ that we consider.

\end{document}